\documentclass[11pt]{article}

\usepackage[final]{acl}

\usepackage{times}
\usepackage{latexsym}
\usepackage{todonotes}
\PassOptionsToPackage{table,xcdraw}{xcolor}

\usepackage{enumitem}
\usepackage{needspace}

\usepackage[T1]{fontenc}

\usepackage[utf8]{inputenc}

\usepackage{microtype}
\usepackage[table]{xcolor}
\usepackage{inconsolata}

\usepackage{graphicx}

\usepackage{multicol}
\usepackage{multirow}
\usepackage{amsfonts,amsmath,amssymb}
\usepackage{tcolorbox}
\usepackage{enumitem}
\usepackage{microtype}
\usepackage{array}
\usepackage{longtable}
\usepackage{booktabs}
\usepackage{url}
\definecolor{lightblue}{RGB}{220,235,255} 
\usepackage{tabularx}

\usepackage{fontawesome5}

\definecolor{headerblue}{RGB}{30,70,150}
\definecolor{pastelteal}{RGB}{100, 200, 190}

\usepackage{pgf}

\definecolor{DeltaNeg}{HTML}{D96C68} 
\definecolor{DeltaPos}{HTML}{6FBF95} 

\newcommand{\DeltaMax}{9.2}
\newcommand{\DeltaCap}{48}

\newcommand{\shadecell}[1]{%
  \pgfmathsetmacro{\magv}{min(\DeltaCap,abs(#1)/\DeltaMax*100)}%
  \ifdim #1pt<0pt
    \edef\nextcell{\noexpand\cellcolor{DeltaNeg!\magv!white}}%
  \else
    \edef\nextcell{\noexpand\cellcolor{DeltaPos!\magv!white}}%
  \fi
  \nextcell
}

\usepackage{pgf}

\definecolor{lightblue}{HTML}{E8F1FA}
\definecolor{DeltaNeg}{HTML}{D96C68}
\definecolor{DeltaPos}{HTML}{6FBF95}

\newcommand{\ChangeMax}{1.38}
\newcommand{\ChangeCap}{48}

\newcommand{\ChangeCell}[3]{%
  \pgfmathsetmacro{\diff}{#1-(#2)}%
  \pgfmathsetmacro{\magv}{min(\ChangeCap,abs(\diff)/\ChangeMax*100)}%
  \ifdim\diff pt>0pt
    \edef\nextcell{\noexpand\cellcolor{DeltaPos!\magv!white}}%
  \else
    \ifdim\diff pt<0pt
      \edef\nextcell{\noexpand\cellcolor{DeltaNeg!\magv!white}}%
    \else
      \def\nextcell{\cellcolor{gray!10}}%
    \fi
  \fi
  \nextcell #3{#1}%
}

\newcommand{\C}[2]{\ChangeCell{#1}{#2}{}}
\newcommand{\CB}[2]{\ChangeCell{#1}{#2}{\textbf}}

\usepackage{pgf}

\definecolor{MonoNeg}{HTML}{D96C68}
\definecolor{MonoPos}{HTML}{6FBF95}

\newcommand{\MonoMax}{0.51}
\newcommand{\MonoCap}{48}

\newcommand{\MonoCell}[2]{%
  \pgfmathsetmacro{\monodiff}{#1-(#2)}%
  \pgfmathsetmacro{\monomag}{min(\MonoCap,abs(\monodiff)/\MonoMax*100)}%
  \ifdim\monodiff pt>0pt
    \edef\monocolor{\noexpand\cellcolor{MonoPos!\monomag!white}}%
  \else
    \ifdim\monodiff pt<0pt
      \edef\monocolor{\noexpand\cellcolor{MonoNeg!\monomag!white}}%
    \else
      \def\monocolor{\cellcolor{gray!10}}%
    \fi
  \fi
  \monocolor #1%
}

\usepackage{xspace}

\newcommand{\model}[1]{\textsc{#1}\xspace}

\newcommand{\dataset}[1]{\textsc{#1}\xspace}

\newcommand{\ds}{\dataset{ArabPref}}

\title{Aligning Cultural Behaviors in Arabic Language Models via Preferences} 
\author{
\textbf{Junhong Liang\textsuperscript{1} \quad
Daniil Orel\textsuperscript{1} \quad
Saeed Almheiri\textsuperscript{1} \quad
Bilal Elbouardi\textsuperscript{1}}\\
\textbf{Younes Samih\textsuperscript{2} \quad
Abed Alhakim Freihat\textsuperscript{1} \quad
Zhuohan Xie\textsuperscript{1} \quad
Momina Ahsan\textsuperscript{1}}\\
\textbf{Muhammad Dehan Al Kautsar\textsuperscript{1} \quad
Sarfraz Ahmad\textsuperscript{1} \quad
Omar El Herraoui\textsuperscript{1}}\\
\textbf{Mohamed Anwar\textsuperscript{1} \quad
Preslav Nakov\textsuperscript{1} \quad
Fajri Koto\textsuperscript{1}}\\[4pt]
\textsuperscript{1}Mohamed bin Zayed University of Artificial Intelligence\\
\textsuperscript{2}IBM Research AI\\[2pt]
\texttt{\{firstname.lastname\}@mbzuai.ac.ae}\\
\faDatabase\ 
\href{https://huggingface.co/datasets/Skywalker-Harrison-mbz/ArabPref}
{\ds}
}

\begin{document}
\maketitle
\begin{abstract}

Arabic LLMs have improved rapidly, but aligning them with country-specific cultural behavioral preferences remains underexplored. Previous work has focused primarily on instruction-tuning for factual knowledge, overlooking nuanced cultural behaviors. We propose \ds, which to our knowledge is the first large-scale pan-Arab preference data set for the alignment of cultural behavior conditioned on the country. Our contribution includes two resources in English and Modern Standard Arabic: a culturally grounded preference dataset for training and evaluation, and a multiple-choice benchmark designed to assess culturally aligned behavior across nations. All test data are validated by native speakers to ensure authenticity. To optimize cultural alignment, we experiment with post-training methods across multilingual and Arabic-centric language models, evaluating performance on both generation tasks and our cultural multiple-choice benchmark. By releasing an evaluation benchmark, we provide a foundation for systematically studying and improving culturally aligned behavior in language models.

\end{abstract}

\section{Introduction}

\begin{figure*}[!t]
    \centering
    \includegraphics[width=\linewidth]{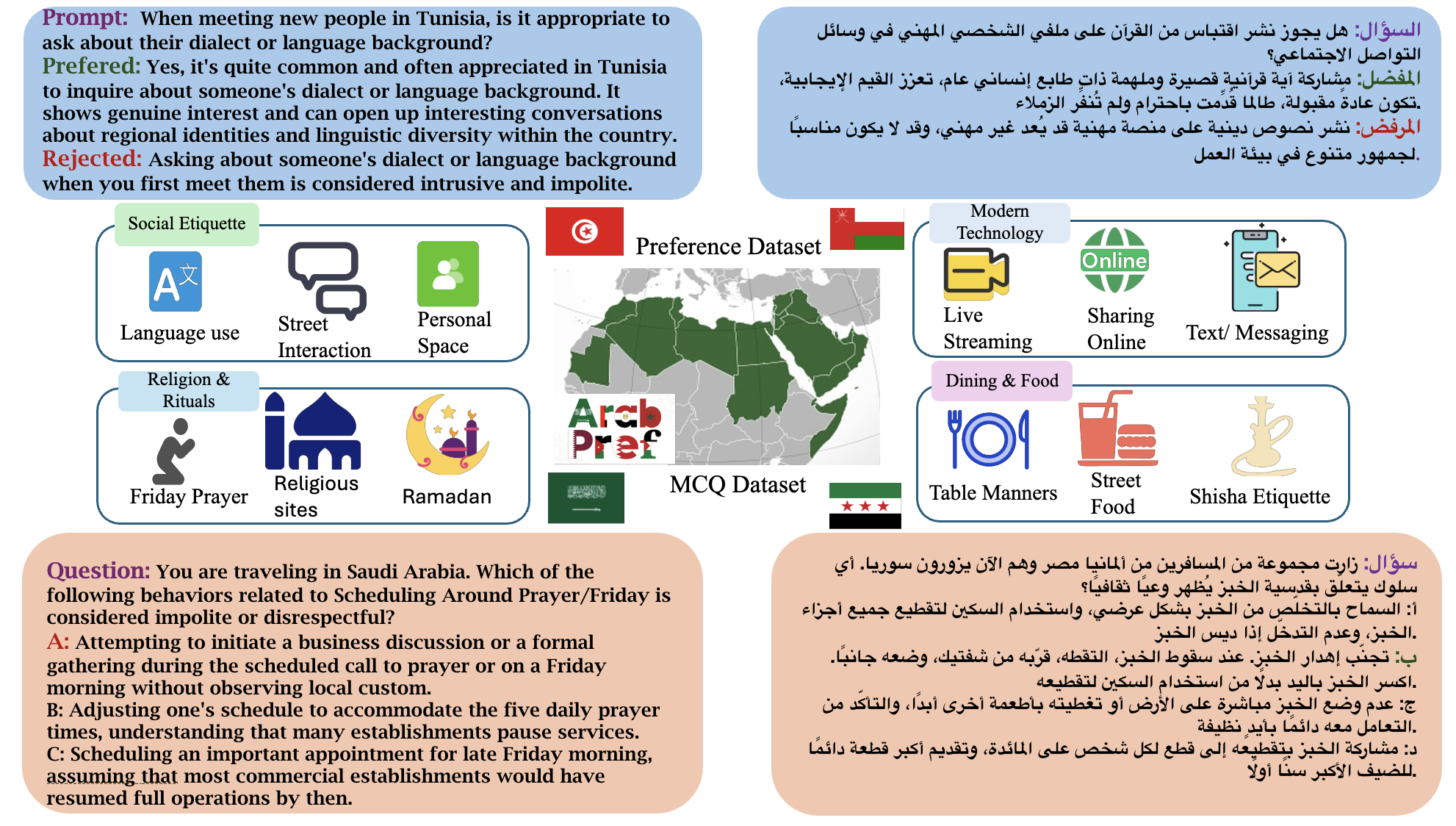}
    \caption{An Overview of our \textbf{\ds} dataset, covering preference and MCQ data across 22 Arab League countries, spanning multiple categories and evaluation aspects, and available in English and Arabic. The upper part demonstrates two preference dataset samples in English and Arabic from Tunisia and Oman, respectively, while the bottom part demonstrates two MCQs from Saudi Arabia and Syria.}
    \label{fig:data_overview}
\end{figure*}

People from different cultural backgrounds communicate and interact according
to different social norms. As large language models (LLMs) become part of
communication, supporting multiple languages is not enough; models
must also respond in ways that fit the cultural contexts of their users.
Culture shapes communication norms, meaning, and judgments of
appropriateness. Despite progress in multilingual modeling
\cite{ahuja-etal-2023-mega,yong-etal-2023-bloom}, LLMs still struggle to
recognize cultural cues and adapt to contexts
\cite{hershcovich-etal-2022-challenges}. As a result, factually correct
outputs may be inappropriate or insensitive, which can reduce trust and
reinforce harmful stereotypes \cite{Liu+2023+224+244}.

Recent work has documented several aspects of this problem. LLMs often reflect
Western value profiles, including in evaluations designed to capture
cross-cultural differences
\cite{cao-etal-2023-assessing,arora-etal-2023-probing,johnson2022ghost,abdulhai-etal-2024-moral,naous-etal-2024-beer}.
Studies of cultural commonsense and geographically diverse factual knowledge
further show that models have difficulty handling local practices, including
regional food traditions, socially grounded norms, and time expressions
\cite{nguyen2023extracting,yin-etal-2022-geomlama}. Other analyses examine how model
representations of cultural values compare with those found in human
societies and everyday dialogue
\cite{dokic2025mirroring,liu2026,khan2025,al-kautsar-etal-2026-cultural}.
Together, these studies point to two recurring gaps: model value profiles often
differ from those of the populations they are expected to represent, and model
decisions in social and economic settings only partly agree with human
preferences across regions. These gaps are particularly important in
non-English and non-Western settings, where Western-centered defaults may be
less appropriate.

We study country-conditioned behavioral appropriateness in culturally situated
social scenarios. This framing treats culture as context-dependent rather than
as a fixed identity or a universal set of values
\cite{naous-etal-2024-beer}. Arabic provides a useful setting for this study.
It is spoken by more than 400 million people across 22 countries that share
broad cultural ties but differ substantially in norms, traditions, and everyday
practices. Treating
Arabic-speaking populations as a single homogeneous group can obscure
country-specific behavioral expectations and make cultural alignment gains
difficult to interpret. This motivates our focus on country-conditioned
preferences and our evaluation of culturally appropriate behavior across both
English and Arabic settings. 

Our contributions are summarized as follows:

\begin{enumerate}
    \item We introduce \textbf{\ds}, the first large-scale behavioral preference dataset spanning all 22 Arab League countries. It covers 200 culturally grounded aspects across 16 categories, including social etiquette, religion, family and gender roles, food, work, and travel, and is released in both English and Modern Standard Arabic (MSA). \ds consists of 31,060 training and 6,264 test preference pairs, culturally grounded triples for alignment and generation, and a native-speaker-validated multiple-choice benchmark with 1,984 questions for evaluation.
   \item Using \textbf{\ds}, we assess how multilingual and Arabic-centric LLMs represent Arab cultural behavior and whether preference alignment improves cultural robustness. Across open-weight models, countries, languages, and cultural categories, we compare supervised fine-tuning (SFT), Direct Preference Optimization (DPO), and Kahneman-Tversky Optimization (KTO), finding that training on \ds improves culturally aligned behavior and downstream cultural reasoning over existing datasets.
\end{enumerate}

\section{Related Work}

\begin{table*}[!t]
\centering
\scriptsize
\setlength{\tabcolsep}{4pt}
\renewcommand{\arraystretch}{0.9}
\begin{tabular}{lccccccc}
\toprule
\textbf{Dataset} & \textbf{Arab-Specific} & \textbf{Topic Granularity} & \textbf{Prefs} & \textbf{Eval (MCQ)} & \textbf{Human Valid.} & \textbf{Langs} & \textbf{Scale} \\
\midrule
\dataset{CARE}         & Partial & --               & Yes & No  & Yes     & Multi (incl.\ Ar) & $\sim$3.5K Q / 32K prefs \\
\dataset{Cidar}        & Yes     & 17 topics        & No  & No  & Yes     & Ar (MSA)          & 10K instr. \\
\dataset{Palm}         & Yes     & 20 (coarse)      & No  & No  & Yes     & Ar (MSA + Dial.)  & $\sim$17.4K instr. \\
\dataset{ArabCulture}  & Yes     & 12 domains       & No  & Yes & Yes     & Ar (MSA)          & $\sim$3.5K MCQs \\
\textbf{\ds (Our)} & \textbf{Yes} & \textbf{200 (fine)} & \textbf{Yes} & \textbf{Yes} & \textbf{Yes} & \textbf{Ar (MSA), En} & \textbf{31K train / 6K test / 2K MCQ} \\
\bottomrule
\end{tabular}
\caption{Comparison of cultural datasets for analyzing and aligning LLM behavior. Existing resources focus on instruction tuning (\dataset{Cidar}, \dataset{Palm}), evaluation (\dataset{ArabCulture}), or limited preference annotation (\dataset{CARE}). \textsc{\ds} uniquely combines pan-Arab coverage, fine-grained topics, and explicit behavioral preferences, enabling controlled preference-based alignment and assessment. Unlike \dataset{Cidar} and \dataset{Palm}, \ds supports preference optimization; unlike \dataset{ArabCulture}, it supports both training and evaluation; unlike \dataset{CARE}, it provides comprehensive pan-Arab coverage with fine-grained country-level topics and a unified MCQ benchmark.}
\label{tab:dataset_compare}
\end{table*}

\paragraph{Cultural grounding and evaluation in Arabic NLP.}
Recent work in Arabic NLP has highlighted the need to go beyond linguistic competence toward culturally grounded modeling. While large Arabic and multilingual language models perform well on standard benchmarks, they often fail to reflect culturally appropriate behavior in socially sensitive domains such as etiquette and religion. To address this gap, \dataset{Cidar} \cite{alyafeai2024cidar} and \dataset{Palm} \cite{alwajih-etal-2025-palm} introduce culturally grounded Arabic instruction-following datasets curated by native speakers, and ArabCulture \cite{sadallah2025arabculture} proposes a multiple-choice benchmark for evaluating cultural commonsense reasoning. However, as summarized in Table~\ref{tab:dataset_compare}, these resources primarily focus on instruction compliance or answer selection and do not explicitly model \emph{normative behavioral preferences}.

\paragraph{Cultural alignment in LLMs.}
A body of work studies cultural alignment in large language models across regions and societies.
Studies show that LLMs often default to Western—frequently American—value profiles when evaluated across cultures \cite{cao-etal-2023-assessing, arora-etal-2023-probing, johnson2022ghost, abdulhai-etal-2024-moral}. 

Analyses grounded in sociological and economic data further demonstrate that model value representations diverge from those of target populations, and that decisions in culturally situated scenarios only partially align with human judgments across regions \cite{dokic2025mirroring, song-etal-2025-many, khan2025, naous-etal-2024-beer}. While these works provide important insights into cultural misalignment, they are largely diagnostic in nature and do not offer direct mechanisms for improving alignment.

\paragraph{Preference-based cultural alignment.}
Preference-based supervision offers a direct means of aligning language models with culturally situated behavioral expectations. Methods such as Reinforcement Learning from Human Feedback, DPO \cite{rafailov2023direct}, and Demonstration Iterated Task Optimization (DITTO) \cite{shaikh2025aligning} optimize models using human comparisons among candidate responses. Recent datasets have begun to extend preference learning beyond generic, English-centric interactions. \dataset{CARE} \cite{guo-etal-2025-care} provides multilingual, human-annotated preference judgments for culturally situated responses, including Arab contexts. IndoPref \cite{wiyono-etal-2025-indopref} introduces a native Indonesian, human-authored pairwise preference dataset across multiple domains, while PRISM \cite{Prism2024} connects fine-grained preference feedback with the demographic characteristics and stated values of participants from diverse countries. However, these resources differ in their alignment targets: IndoPref primarily focuses on response naturalness and quality, whereas PRISM emphasizes subjective and individualized preferences rather than explicit culture-specific behavioral norms. Moreover, \dataset{CARE} remains limited in scale and does not provide comprehensive pan-Arab coverage or a unified evaluation benchmark. 

Consequently, existing resources tend to support either culturally grounded instruction and evaluation or general and multicultural preference learning, leaving large-scale preference-based cultural alignment---particularly for Arabic and the wider Arab world---underexplored.

\section{\ds}

We design \ds around country--aspect pairs because cultural norms vary by locale and social context. Preference pairs capture contrastive behavioral judgments, while MCQs evaluate whether models can recognize culturally appropriate behavior among plausible alternatives. \ds has two components: a preference dataset and a multiple-choice question (MCQ) dataset. Figure~\ref{fig:data_overview} illustrates the samples. To construct datasets, we generate rudimentary preference data (Section~\ref{subsec:rudimentary_preference_data}). We refine the preference dataset through human annotation (Section~\ref{subsec:preference_generation}) and generate the MCQ dataset for evaluation (Section~\ref{subsec:mcq_generation}).

\begin{figure*}[!t]
    \centering
    \includegraphics[width=\linewidth]{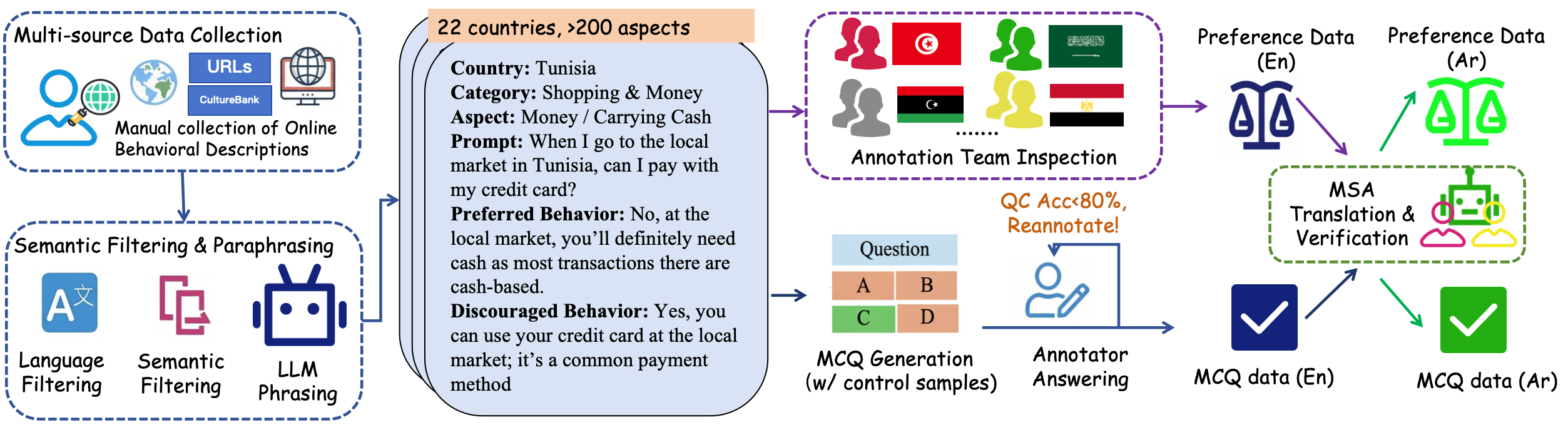}
    \caption{
    Overview of the data curation and validation pipeline. Our process begins with the manual collection of cultural behavioral descriptions from diverse online sources and URLs, supplemented by CultureBank. After automated semantic filtering and paraphrasing, the data undergoes a rigorous Human-in-the-loop (HITL) validation phase. Annotators perform an inspection to ensure cultural authenticity. For the MCQ generation, a Quality Control (QC) loop is implemented where samples with low accuracy are re-annotated. Finally, the data is translated into MSA and verified to produce the final bilingual Preference and MCQ datasets.}
    \label{fig:data_creation}
\end{figure*}

\subsection{Rudimentary Preference Data}
\label{subsec:rudimentary_preference_data}

\paragraph{Arab League Countries}
This stage aims to construct a behavioral preference dataset covering the 22 member states of the Arab League\footnote{\url{https://en.wikipedia.org/wiki/Arab\_League}}. Each country possesses a distinct cultural and historical background, which translates into nuanced behavioral patterns in real-world applications.

\paragraph{Cultural Categories} 
This stage establishes cultural categories before generating preference data. All extracted categories undergo manual review to ensure contextual appropriateness and cultural relevance. We obtain 16 cultural categories, including social etiquette, business and work, and religion and rituals, expanding them into 200 aspects. Humans verify all categories and aspects. A list of aspects for each category appears in Appendix~\ref{appx:categories}.

\paragraph{Preferred and Discouraged Behaviors} 
The dataset is constructed through a multi-stage pipeline, as illustrated in Figure~\ref{fig:data_creation}. First, we aggregate behavioral data from two primary sources: (i) manually filtered web-based
resources~\footnote{We use travel blogs and official travel recommendations from embassies; a full list with descriptions is given in Appendix~\ref{appx:urls_desc}} and (ii) entries from the CultureBank \citep{shi-etal-2024-culturebank}. All raw data are originally in English. We then process these sources to extract distinct country--aspect pairs and split these pairs into disjoint training and test sets before paraphrase generation.

For each country–aspect pair, we use Gemini-2.5-Flash to paraphrase pairs of preferred and discouraged behaviors using the cultural sources as context. We then apply automated filters: language validation, structural de-duplication via the \texttt{MinHash} algorithm~\cite{minhash}, and semantic overlap removal (see Appendix~\ref{apdx:semantic_filtering}). To ensure that these contrasts capture cultural nuances and correct potential model hallucinations, we implement a Human-in-the-Loop (HITL) annotation process.

\subsection{Preference Dataset Generation}
\label{subsec:preference_generation}

To refine the rudimentary preference data described in Section~\ref{subsec:rudimentary_preference_data}, we conducted a large-scale human annotation project with a professional linguistic service provider. This ensures the dataset captures authentic cultural nuances and maintains high linguistic quality in the 22 target countries. The preference generation follows a multi-stage HITL framework, including cultural preference validation and MSA verification. We describe the full quality-control procedure in Section~\ref{sec:quality_check}.

\subsection{MCQ Generation}
\label{subsec:mcq_generation}

Based on the validated preference aspects, we generate MCQs to evaluate a model's ability to distinguish culturally appropriate behaviors. Importantly, we split the data at the country--cultural-aspect level before generating preference paraphrases. Preference instances used for training are derived exclusively from the training country--aspect pairs, whereas MCQs are constructed only from held-out country--aspect pairs in the test split. Consequently, no training prompt, chosen response, or rejected response is reused as an MCQ stem or answer option.

\paragraph{Distractor Selection and Refinement}
For each \textit{country--aspect} pair, we create MCQs using predefined templates with scenario-specific details. To construct challenging MCQs, the target behavior serves as the \textit{correct answer}, while three \textit{distractors} are generated: one representing the opposite behavior (inverse polarity) within the same country, and two representing behaviors from the same aspect but associated with different countries. We then apply semantic filtering to ensure that the answer choices are sufficiently distinct. (Appendix \ref{apdx:semantic_filtering}) and a specific LLM prompt to ensure language neutrality (Appendix~\ref{apdx:llm_as_a_judge_prompts}).
This ensures that the model must distinguish within a specific cultural context and that its choices do not overlap with the training data.

\textbf{Human-in-the-Loop Validation}
The annotation team described in Section~\ref{subsec:preference_generation} performed a final validation of the MCQs. To ensure reliability, we randomly inserted quality control (QC) samples into the workflow. 

Annotators were required to maintain a minimum accuracy of 80\% on these QC items to be considered valid. Following the validation of the English MCQs, we utilized an LLM for initial MSA translation, followed by a final round of human post-editing to guarantee technical and cultural flawlessness. The whole annotation guideline is in Appendix~\ref{apdx:annotation_guidelines}. 

\begin{table}[t]
\centering
\scalebox{0.75}{
\begin{tabular}{lcccccc}
\toprule
\multirow{2}{*}{\textbf{Country}} 
& \multicolumn{2}{c}{\textbf{Pref Train}} 
& \multicolumn{2}{c}{\textbf{Pref Test}} 
& \multicolumn{2}{c}{\textbf{MCQ}} \\
\cmidrule(lr){2-3} \cmidrule(lr){4-5} \cmidrule(lr){6-7}
& \textbf{En} & \textbf{Ar} 
& \textbf{En} & \textbf{Ar} 
& \textbf{En} & \textbf{Ar} \\
\midrule
Egypt        & 1219 & 1219 & 329 & 329 & 62 & 62 \\
Morocco      & 1114 & 1114 & 226 & 159 & 59 & 59 \\
Saudi Arabia & 972  & 972  & 141 & 141 & 52 & 52 \\
Algeria      & 958  & 958  & 165 & 131 & 50 & 50 \\
Jordan       & 924  & 924  & 254 & 202 & 72 & 72 \\
Lebanon      & 907  & 907  & 199 & 198 & 39 & 39 \\
Tunisia      & 860  & 860  & 317 & 308 & 61 & 61 \\
Yemen        & 758  & 758  & 189 & 162 & 62 & 62 \\
Syria        & 757  & 757  & 128 & 117 & 35 & 35 \\
UAE          & 720  & 720  & 159 & 156 & 49 & 49 \\
Palestine    & 679  & 679  & 135 & 135 & 46 & 46 \\
Libya        & 628  & 628  & 114 & 79  & 27 & 27 \\
Iraq         & 598  & 598  & 135 & 127 & 28 & 28 \\
Oman         & 572  & 572  & 134 & 102 & 40 & 40 \\
Sudan        & 564  & 564  & 80  & 80  & 28 & 28 \\
Bahrain      & 545  & 545  & 147 & 134 & 41 & 41 \\
Qatar        & 517  & 517  & 104 & 86  & 49 & 49 \\
Somalia      & 488  & 488  & 32  & 31  & 43 & 43 \\
Kuwait       & 476  & 476  & 82  & 81  & 43 & 43 \\
Djibouti     & 436  & 436  & 51  & 41  & 41 & 41 \\
Comoros      & 421  & 421  & 82  & 82  & 31 & 31 \\
Mauritania   & 417  & 417  & 97  & 83  & 34 & 34 \\
\midrule
\textbf{Total} 
& \textbf{15530} & \textbf{15530} 
& \textbf{3300}  & \textbf{2964} 
& \textbf{992}   & \textbf{992} \\
\bottomrule
\end{tabular}
}
\caption{Statistics of the ArabPref dataset, including English and Arabic preference data and MCQ coverage across 22 Arab League countries. Where the test preferences and MCQs are verified by human annotators.}
\label{tab:ArabPref_count}
\end{table}

\subsection{Quality Check}
\label{sec:quality_check}

\paragraph{Validation safeguards}
To ensure that \ds reflects culturally grounded behavior rather than unverified model-generated content, we implement validation safeguards throughout the annotation pipeline for test preference data and MCQs. All test preference and MCQ data are reviewed by a team of 40 specialized annotators, with 2--3 experts assigned to each target country to provide country-specific cultural knowledge. We enforce a Native-In-Country policy whenever possible: 95\% of annotators reside in their native countries, while the remaining 5\% are native speakers who have lived in the target region for over 15 years. This composition helps ground validation decisions in linguistic competence and familiarity with local norms.

The preference data undergo a two-stage HITL protocol. In the first stage, \textit{Cultural Preference Validation}, annotators review English prompt-chosen-rejected triples and verify whether the chosen and rejected responses genuinely reflect preferred and discouraged behaviors in the target country's cultural context. In the second stage, \textit{MSA Verification}, native speakers verify the MSA translations for fluency, naturalness, and semantic alignment with the original English source, while filtering out unqualified translations. For MCQ validation, annotation quality is monitored through randomly inserted quality-control samples, and annotators are required to maintain at least 80\% accuracy on these items. Batches showing low engagement, such as indiscriminate acceptance, minimal edits, or lack of meaningful revisions, are flagged for manual review and, when necessary, escalated to senior linguistic experts. After that, as suggested by annotators, some MCQs are reduced to three answer options to avoid cases in which multiple answers could be considered correct. Finally, a 10\% subsample of the annotated data undergoes secondary expert verification.

For preference data, this process yields a raw agreement rate of 90.1\%, indicating strong inter-annotator concordance. For MCQ data, inter-annotator agreement yields a Cohen's $\kappa$ of 0.693, indicating substantial agreement \citep{PMID:843571}. In addition, 15.25\% of the originally collected English cultural preference candidates are rejected by human annotators, while a further suggestion that human validation substantially reshapes the automatically generated candidates rather than merely approving them. The final data distribution is summarized in Table~\ref{tab:ArabPref_count}.



\begin{table*}[!t]
\centering
\scalebox{0.75}{%
\begin{tabular}{lcccccccc}
\toprule
\multirow{2}{*}{Model \& Method}
& \multicolumn{4}{c}{English (En)}
& \multicolumn{4}{c}{Arabic (Ar)} \\
\cmidrule(lr){2-5}
\cmidrule(lr){6-9}
& Cultural & General & Literacy & Avg.
& Cultural & General & Literacy & Avg. \\
\midrule

\rowcolor{lightblue}
\textbf{Llama-3.1-8B-instruct} & & & & & & & & \\

\quad Vanilla
& 3.23 & 3.35 & 3.63 & 3.40
& 3.13 & 3.24 & 3.07 & 3.15 \\

\quad w/ \dataset{Palm} (SFT)
& \C{3.74}{3.23}* & \C{3.92}{3.35}* & \CB{4.08}{3.63}* & \C{3.91}{3.40}*
& \C{3.36}{3.13}* & \C{3.49}{3.24}* & \CB{3.88}{3.07}* & \C{3.58}{3.15}* \\

\quad w/ \dataset{CARE} (DPO)
& \C{3.85}{3.23}* & \C{4.08}{3.35}* & \C{3.83}{3.63}* & \C{3.92}{3.40}*
& \C{3.24}{3.13}* & \C{3.38}{3.24}* & \C{3.27}{3.07}* & \C{3.30}{3.15}* \\

\quad w/ \dataset{CARE} (KTO)
& \C{3.66}{3.23}* & \C{3.82}{3.35}* & \C{3.69}{3.63}* & \C{3.72}{3.40}*
& \C{3.17}{3.13} & \C{3.31}{3.24}* & \C{3.08}{3.07} & \C{3.19}{3.15} \\

\quad w/ \ds (DPO)
& \C{3.78}{3.23}* & \C{3.96}{3.35}* & \C{3.61}{3.63} & \C{3.78}{3.40}*
& \C{3.49}{3.13}* & \C{3.66}{3.24}* & \C{3.63}{3.07}* & \C{3.59}{3.15}* \\

\quad w/ \ds (KTO)
& \C{3.95}{3.23}* & \C{4.15}{3.35}* & \C{3.58}{3.63}* & \C{3.89}{3.40}*
& \C{3.50}{3.13}* & \C{3.68}{3.24}* & \C{3.08}{3.07} & \C{3.42}{3.15}* \\

\quad w/ \ds+\dataset{CARE} (DPO)
& \C{3.78}{3.23}* & \C{3.97}{3.35}* & \C{3.97}{3.63}* & \C{3.91}{3.40}*
& \CB{3.76}{3.13}* & \CB{3.94}{3.24}* & \C{3.36}{3.07}* & \CB{3.69}{3.15}* \\

\quad w/ \ds+\dataset{CARE} (KTO)
& \C{3.85}{3.23}* & \C{4.05}{3.35}* & \C{3.76}{3.63}* & \C{3.89}{3.40}*
& \C{3.37}{3.13}* & \C{3.54}{3.24}* & \C{3.15}{3.07}* & \C{3.35}{3.15}* \\

\quad w/ \ds+\dataset{CARE} (SFT)
& \CB{4.12}{3.23}* & \CB{4.32}{3.35}* & \C{3.67}{3.63} & \CB{4.04}{3.40}*
& \C{3.36}{3.13}* & \C{3.52}{3.24}* & \C{2.89}{3.07}* & \C{3.26}{3.15}* \\

\midrule

\rowcolor{lightblue}
\textbf{Qwen-2.5-7B-Instruct} & & & & & & & & \\

\quad Vanilla
& 3.41 & 3.58 & 4.07 & 3.69
& 3.89 & 4.08 & 3.98 & 3.98 \\

\quad w/ \dataset{Palm} (SFT)
& \C{4.02}{3.41}* & \C{4.28}{3.58}* & \C{4.57}{4.07}* & \C{4.29}{3.69}*
& \C{3.36}{3.89}* & \C{3.53}{4.08}* & \C{3.68}{3.98}* & \C{3.52}{3.98}* \\

\quad w/ \dataset{CARE} (DPO)
& \CB{4.28}{3.41}* & \C{4.55}{3.58}* & \C{4.06}{4.07} & \C{4.30}{3.69}*
& \C{4.21}{3.89}* & \C{4.46}{4.08}* & \C{3.66}{3.98}* & \C{4.11}{3.98}* \\

\quad w/ \dataset{CARE} (KTO)
& \C{4.11}{3.41}* & \C{4.37}{3.58}* & \C{4.00}{4.07}* & \C{4.16}{3.69}*
& \C{4.19}{3.89}* & \C{4.44}{4.08}* & \C{3.57}{3.98}* & \C{4.07}{3.98}* \\

\quad w/ \ds (DPO)
& \C{4.26}{3.41}* & \C{4.53}{3.58}* & \C{4.43}{4.07}* & \C{4.41}{3.69}*
& \C{4.01}{3.89}* & \C{4.25}{4.08}* & \C{3.81}{3.98}* & \C{4.02}{3.98} \\

\quad w/ \ds (KTO)
& \C{4.33}{3.41}* & \C{4.60}{3.58}* & \C{4.71}{4.07}* & \C{4.55}{3.69}*
& \CB{4.33}{3.89}* & \CB{4.59}{4.08}* & \C{4.22}{3.98}* & \C{4.38}{3.98}* \\

\quad w/ \ds+\dataset{CARE} (DPO)
& \CB{4.38}{3.41}* & \C{4.31}{3.58}* & \C{3.86}{4.07}* & \C{4.18}{3.69}*
& \C{4.27}{3.89}* & \C{4.53}{4.08}* & \C{3.94}{3.98}* & \C{4.25}{3.98}* \\

\quad w/ \ds+\dataset{CARE} (KTO)
& \C{4.19}{3.41}* & \CB{4.69}{3.58}* & \C{3.53}{4.07}* & \C{4.14}{3.69}*
& \C{4.32}{3.89}* & \C{4.58}{4.08}* & \CB{4.51}{3.98}* & \CB{4.47}{3.98}* \\

\quad w/ \ds+\dataset{CARE} (SFT)
& \C{4.36}{3.41}* & \C{4.67}{3.58}* & \CB{4.72}{4.07}* & \CB{4.58}{3.69}*
& \C{3.94}{3.89}* & \C{4.18}{4.08}* & \C{2.60}{3.98}* & \C{3.57}{3.98}* \\

\midrule

\rowcolor{lightblue}
\textbf{Jais-adapted-7b-chat} & & & & & & & & \\

\quad Vanilla
& 3.41 & 3.58 & 4.22 & 3.74
& 3.40 & 3.57 & 3.64 & 3.54 \\

\quad w/ \dataset{Palm} (SFT)
& \C{3.18}{3.41}* & \C{3.34}{3.58}* & \C{3.95}{4.22}* & \C{3.49}{3.74}*
& \C{3.26}{3.40}* & \C{3.42}{3.57}* & \C{3.74}{3.64}* & \C{3.47}{3.54}* \\

\quad w/ \dataset{CARE} (DPO)
& \C{3.99}{3.41}* & \C{4.23}{3.58}* & \C{4.31}{4.22}* & \C{4.18}{3.74}*
& \C{3.36}{3.40}* & \C{3.56}{3.57} & \C{4.20}{3.64}* & \C{3.71}{3.54}* \\

\quad w/ \dataset{CARE} (KTO)
& \C{3.28}{3.41}* & \C{3.44}{3.58}* & \C{4.20}{4.22} & \C{3.64}{3.74}*
& \C{3.37}{3.40} & \C{3.54}{3.57} & \C{3.84}{3.64}* & \C{3.58}{3.54} \\

\quad w/ \ds (DPO)
& \C{4.33}{3.41}* & \C{4.59}{3.58}* & \C{4.68}{4.22}* & \C{4.53}{3.74}*
& \C{4.24}{3.40}* & \C{4.49}{3.57}* & \CB{4.21}{3.64}* & \C{4.31}{3.54}* \\

\quad w/ \ds (KTO)
& \C{3.42}{3.41} & \C{3.59}{3.58} & \C{4.06}{4.22}* & \C{3.69}{3.74}*
& \C{3.62}{3.40} & \C{3.80}{3.57}* & \C{3.89}{3.64}* & \C{3.77}{3.54}* \\

\quad w/ \ds+\dataset{CARE} (DPO)
& \C{4.39}{3.41}* & \C{4.65}{3.58}* & \CB{4.77}{4.22}* & \C{4.60}{3.74}*
& \C{4.28}{3.40}* & \C{4.54}{3.57}* & \C{4.20}{3.64}* & \C{4.34}{3.54}* \\

\quad w/ \ds+\dataset{CARE} (KTO)
& \C{3.52}{3.41}* & \C{3.70}{3.58}* & \C{4.37}{4.22}* & \C{3.86}{3.74}*
& \C{3.86}{3.40}* & \C{4.09}{3.57}* & \C{3.81}{3.64}* & \C{3.92}{3.54}* \\

\quad w/ \ds+\dataset{CARE} (SFT)
& \CB{4.50}{3.41}* & \CB{4.77}{3.58}* & \C{4.58}{4.22}* & \CB{4.62}{3.74}*
& \CB{4.42}{3.40}* & \CB{4.68}{3.57}* & \C{4.20}{3.64}* & \CB{4.43}{3.54}* \\

\midrule

\rowcolor{lightblue}
\multicolumn{9}{l}{\textbf{Proprietary models (zero-shot)}} \\

\model{DeepSeek-V4-Flash}
& 4.80 & 4.56 & 4.78 &  4.71
& 4.76 & 4.74 & 4.80 & 4.77   \\

\model{Qwen3.6-Plus}
& 4.75 & 4.83 & 4.71 & 4.76
& 4.74 & 4.63 & 4.61 & 4.66 \\

\model{Claude-Sonnet-4}
& 4.91 & 4.85 & 4.99 & 4.92
& 4.87 & 4.36 & 4.50 & 4.58 \\

\bottomrule
\end{tabular}%
}
\caption{Preference-alignment results in English and Arabic, evaluated using
an LLM-as-a-judge framework along three dimensions: Cultural Appropriateness,
General Quality, and Literacy. All values are scores on a 1--5 Likert scale.
Cell colors indicate changes relative to the corresponding Vanilla model:
\textcolor{MonoPos}{green} denotes improvement and \textcolor{MonoNeg}{red} denotes degradation, with darker shades
representing larger absolute changes. The best result for each model is
highlighted in \textbf{bold}. Asterisks on the changes
indicate statistically significant differences from the corresponding vanilla
model ($p<0.05$) through two-sided paired permutation tests. Proprietary models are evaluated in a zero-shot
setting.}
\label{tab:model_results_main}
\end{table*}
\subsection{Cultural Differences} 
\ds is explicitly designed to capture the substantial cultural heterogeneity within the Arab world. For \textit{Alcohol}, preferred responses range from strict prohibition (\textit{Saudi Arabia}, \textit{Libya}) through venue-restricted permissibility (\textit{UAE}, \textit{Jordan}) to comparatively liberal norms in \textit{Lebanon}. Similarly, \textit{Business Dress Codes} range from traditional attire, such as the thobe and turban in \textit{Yemen} or the abaya in \textit{Saudi Arabia}, to conservative Western-style clothing in \textit{Algeria} and \textit{Egypt}. \textit{Public Affection} follows a similar gradient: while \textit{Egypt} and \textit{Syria} advise against hand-holding, \textit{Lebanon} and \textit{Tunisia} (specifically in urban areas) mark it as acceptable. Details are available in Appendix \ref{apdx:intra_arab}.

\section{Arabic Cultural Behavior Alignment}
\subsection{Alignment Methods}
We employ three post-training alignment methods: Supervised Fine-Tuning (SFT),
Direct Preference Optimization (DPO)~\cite{rafailov2023direct}, and
Kahneman--Tversky Optimization (KTO)~\cite{ethayarajh2024kto}. These methods
cover complementary supervision signals, including direct response imitation,
pairwise preference learning, and binary preference optimization. In addition,
to investigate sampling efficiency, we include Demonstration-Driven Iterative
Training with Preference Optimization (DITTO)~\cite{shaikh2025aligning}.
This setup allows us to compare different alignment objectives and examine
their effects on culturally appropriate behavior. Detailed method formulas are
provided in Appendix~\ref{apdx:alignment_methods}, while the sampling-efficiency
experiments are reported in Appendix~\ref{apdx:sample_efficiency}.



\subsection{Experiment Setup}


\paragraph{Baselines}  We evaluate a diverse range of LLMs categorized into three distinct groups: (1) \textbf{Multilingual Models}, including Llama-3.1-8B-Instruct~\cite{grattafiori2024llama3herdmodels} and Qwen-2.5-7B-Instruct~\cite{qwen2025qwen25technicalreport}; (2) \textbf{Arabic-centric Models}, represented by Jais-Adapted-7B-Chat~\cite{sengupta2023jais,jaisfamilymodelcard}; (3) \textbf{Proprietary Models}, including DeepSeek-V4-Flash~\cite{deepseekai2026deepseekv4highlyefficientmilliontoken}, Qwen3.6-Plus~\cite{yang2025qwen3technicalreport}, GPT-5.4~\cite{gpt5} and Claude-Sonnet-4~\cite{anthropic2025system}.

\paragraph{Alignment Data}  
We use the preference training part of our \ds dataset for cultural alignment. We also use \dataset{CARE}~\cite{guo-etal-2025-care}, which is a multilingual preference dataset spanning Chinese, Japanese, and Arabic cultures. We select the Arabic portion of \dataset{CARE} for reinforcement learning. Furthermore, we use \dataset{Palm}~\cite{alwajih-etal-2025-palm}, an Arabic instruction dataset, from which we choose the MSA portion for fine-tuning.

\paragraph{Evaluation Metrics}  
To quantify the cultural alignment and linguistic quality of model generations, we employ an \textit{LLM-as-a-judge} framework. We utilize GPT-4o~\cite{openai2024gpt4technicalreport} as a backbone evaluator, guided by a multi-dimensional rubric specifically designed to capture the nuances of cultural behavior following \citet{guo-etal-2025-care}. Each response is scored on a Likert scale (1--5) across three primary criteria, cultural appropriatenesss, general quality, and literacy. A detailed description of the metrics and scoring can be found in Appendix~\ref{apdx:evaluation_criterion}. We also ask humans to annotate the responses; the Spearman's correlation between human scoring and machine scoring is available in Appendix~\ref{apdx:correlation_coefficiency}, showing a strong Spearman's correlation exceeding 0.60 between human and machine annotations. A stability analysis of LLM-as-a-judge evaluation is available in Appendix~\ref{apdx:stability_of_llm_as_a_judge}.

\begin{table}[t]
\centering
\small
\setlength{\tabcolsep}{5pt}
\renewcommand{\arraystretch}{1}
\scalebox{0.87}{
\begin{tabular}{lcccc}
\toprule
\multirow{2}{*}{\textbf{Model \& Method}}
& \multicolumn{2}{c}{\textbf{English (En)}}
& \multicolumn{2}{c}{\textbf{Arabic (Ar)}} \\
\cmidrule(lr){2-3}
\cmidrule(lr){4-5}
& \textbf{Acc.} & $\boldsymbol{\Delta}$
& \textbf{Acc.} & $\boldsymbol{\Delta}$ \\
\midrule

\rowcolor{gray!15}
\multicolumn{5}{l}{\textbf{Jais-adapted-7B-chat}} \\

\quad Vanilla
& 49.6 & -- & 39.4 & -- \\

\quad w/ \dataset{Palm} (SFT)
& 40.4 & $\downarrow 9.2$
& 37.1 & $\downarrow 2.3$ \\

\quad w/ \dataset{CARE} (DPO)
& 51.0 & $\uparrow 1.4$
& \textbf{45.6} & $\mathbf{\uparrow 6.2}$ \\

\quad w/ \dataset{CARE} (KTO)
& 50.0 & $\uparrow 0.4$
& 40.1 & $\uparrow 0.7$ \\

\quad w/ \ds (DPO)
& \underline{52.6} & $\underline{\uparrow 3.0}$
& 40.2 & $\uparrow 0.8$ \\

\quad w/ \ds (KTO)
& 51.7 & $\uparrow 2.1$
& 41.8 & $\uparrow 2.4$ \\

\quad w/ \ds+\dataset{CARE} (DPO)
& \textbf{53.7} & $\mathbf{\uparrow 4.1}$
& \underline{43.1} & $\underline{\uparrow 3.7}$ \\

\quad w/ \ds+\dataset{CARE} (KTO)
& 51.3 & $\uparrow 1.7$
& 42.4 & $\uparrow 3.0$ \\

\midrule

\rowcolor{gray!15}
\multicolumn{5}{l}{\textbf{Qwen-2.5-7B-Instruct}} \\

\quad Vanilla
& 63.3 & -- & 57.6 & -- \\

\quad w/ \dataset{Palm} (SFT)
& 63.0 & $\downarrow 0.3$
& 56.9 & $\downarrow 0.7$ \\

\quad w/ \dataset{CARE} (DPO)
& 67.5 & $\uparrow 4.2$
& 57.6 & $\uparrow 0.0$ \\

\quad w/ \dataset{CARE} (KTO)
& 66.2 & $\uparrow 2.9$
& 57.8 & $\uparrow 0.2$ \\

\quad w/ \ds (DPO)
& 67.0 & $\uparrow 3.7$
& 58.0 & $\uparrow 0.4$ \\

\quad w/ \ds (KTO)
& 66.4 & $\uparrow 3.1$
& 57.5 & $\downarrow 0.1$ \\

\quad w/ \ds+\dataset{CARE} (DPO)
& \textbf{68.6} & $\mathbf{\uparrow 5.3}$
& \textbf{58.7} & $\mathbf{\uparrow 1.1}$ \\

\quad w/ \ds+\dataset{CARE} (KTO)
& \underline{67.1} & $\underline{\uparrow 3.8}$
& \underline{58.3} & $\underline{\uparrow 0.7}$ \\

\midrule

\rowcolor{gray!15}
\multicolumn{5}{l}{\textbf{Proprietary models (zero-shot)}} \\

\quad \model{DeepSeek-V4-Flash}
& 75.3 & -- & 71.6 & -- \\

\quad \model{Qwen3.6-Plus}
& \textbf{80.5} & -- & \textbf{76.5} & -- \\

\quad \model{GPT-5.4}
& \underline{79.2} & -- & \underline{75.9} & -- \\

\quad \model{Claude-Sonnet-4}
& 74.8 & -- & 73.7 & -- \\

\bottomrule
\end{tabular}
}

\caption{MCQ accuracy (\%) on the English and Arabic datasets.
$\Delta$ denotes the absolute change relative to the corresponding Vanilla
model. For each open-source model, the best result is shown in bold and the
second-best result is underlined. Proprietary models are evaluated in the
zero-shot setting.}
\label{tab:mcq_accuracy}
\vspace{-1.5em}
\end{table}

\subsection{Preference Generation}

Table~\ref{tab:model_results_main} shows the preference-generation results
from the LLM-as-a-judge evaluation. Preference alignment improves
open-weight models, but the gains vary across models, methods, and
languages. Jais shows the most consistent improvement. With \ds+\dataset{CARE} SFT, its
English average score increases from 3.74 to 4.62, and its Arabic average
score increases from 3.54 to 4.43. In contrast, Qwen achieves its best
English average score with \ds+\dataset{CARE} SFT (4.58), but its Arabic average
score drops from 3.98 to 3.57, mainly because its Literacy score decreases
from 3.98 to 2.60. \ds+\dataset{CARE} KTO instead gives Qwen the highest Arabic
average score of 4.47. Llama also improves mainly in English, while its best
Arabic average score is 3.69 with \ds+\dataset{CARE} DPO. Overall, no single alignment method consistently benefits all models and languages. 

The zero-shot proprietary models provide a reference for the open-weight models. Claude Sonnet 4 achieves the highest English average (4.92), followed by Qwen3.6-Plus (4.76) and DeepSeek-V4-Flash (4.71). In Arabic, DeepSeek-V4-Flash ranks first at 4.77, ahead of Qwen3.6-Plus (4.66) and Claude Sonnet 4 (4.58). The best open-weight averages are 4.62 for Jais in English and 4.47 for Qwen in Arabic. Although the proprietary models score higher on average, the differences are modest for DeepSeek-V4-Flash in English and Claude Sonnet 4 in Arabic. Results vary by dimension. Jais matches or exceeds DeepSeek-V4-Flash on English General Quality (4.77 vs.\ 4.56), and Qwen outperforms Qwen3.6-Plus and Claude Sonnet 4 on Arabic General Quality (4.68 vs.\ 4.63 and 4.36). Proprietary models obtain the highest Cultural Appropriateness scores in both languages. These zero-shot results are reference points rather than direct comparisons of alignment methods.

\subsection{MCQ Answering}

Results in Table~\ref{tab:mcq_accuracy} show that the effect of alignment
varies across models and languages. SFT provides limited benefits and can
even reduce accuracy, whereas DPO and KTO generally perform better on the
structured MCQ task.

\ds+\dataset{CARE} DPO yields the largest gains for Qwen,
increasing accuracy by $+5.2$ points in English and $+1.1$ points in Arabic,
and improves Jais by $+3.7$ points in Arabic. Jais also shows a larger Arabic
gain under \dataset{CARE} DPO ($+6.2$) than Qwen, despite starting from a lower baseline.
Combining \dataset{CARE} with \ds gives the best overall performance, reaching
$68.6\%$ English accuracy for Qwen. These results suggest that general
preference data and culturally focused data provide complementary benefits,
but the gains are not consistent across models and languages.

\begin{figure}[t]
    \centering
    \includegraphics[width=0.85\linewidth]{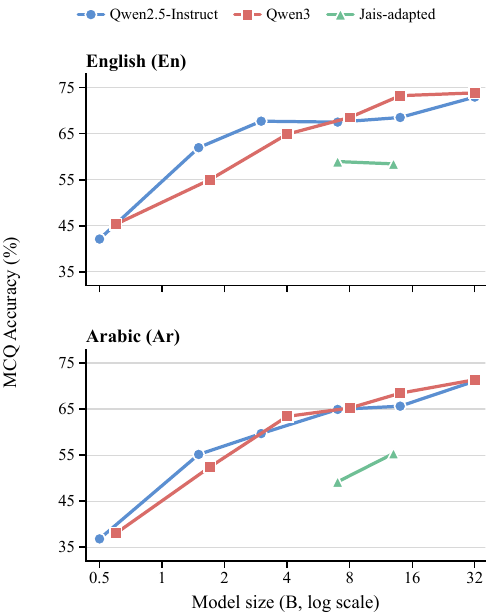}
    \caption{Scaling behavior on the \ds MCQ benchmark. The upper and lower panels report accuracy on the English (En) and Arabic (Ar) test sets, respectively. Results are shown for Qwen2.5-Instruct, Qwen3, and Jais-adapted models across different model sizes. Each language contains 992 questions.}
    \label{fig:mcq_scaling}
\end{figure}

\begin{figure*}[!t]
    \centering
    \includegraphics[width=\linewidth]{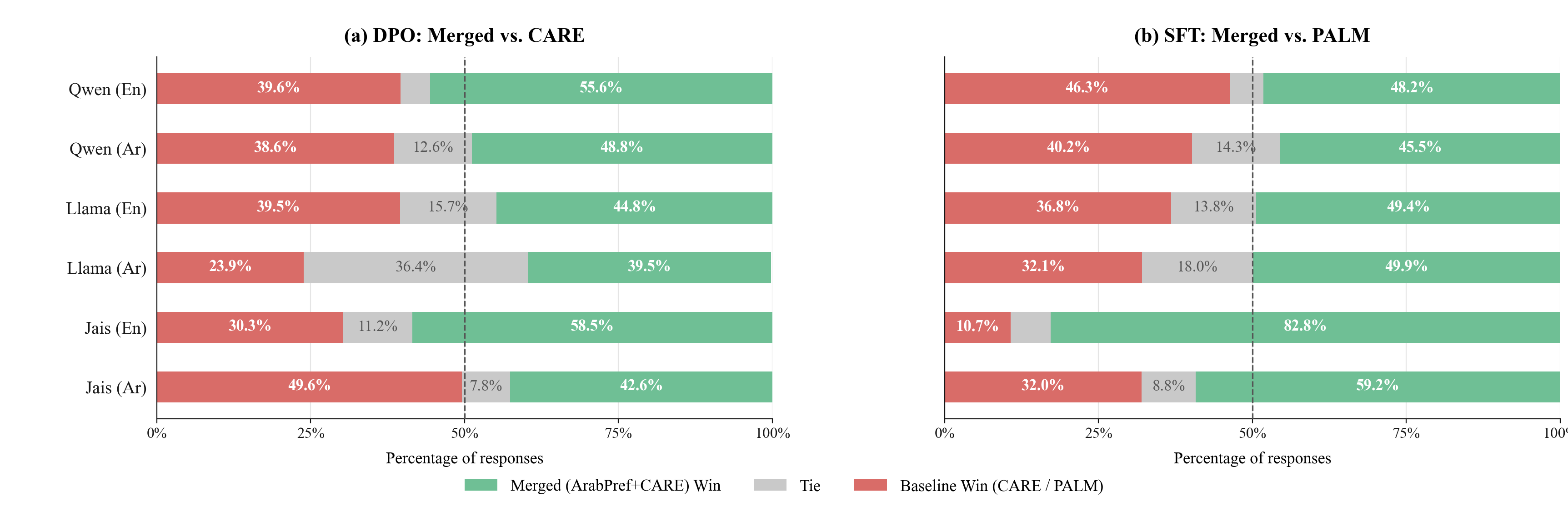}
    \caption{Head-to-head win rate comparisons evaluated by GPT-4o as judge. Each bar shows the win percentage of the respective system relative to the 50\% midpoint. (a) DPO training: our merged dataset (\ds+\dataset{CARE}) vs. \dataset{CARE}-only across three base models and two languages (Ar/En). (b) SFT training: merged dataset vs. \dataset{Palm}. Results show that the merged dataset consistently achieves competitive or superior win rates, demonstrating that the observed gains generalise beyond vanilla baselines.}
    \label{fig:win_rate_compare}
    \vspace{-1em}
\end{figure*}

\section{Discussion}

\textbf{Scaling}
We examine scaling across the Qwen2.5, Qwen3~\cite{yang2025qwen3technicalreport}, and Jais-adapted model families. Figure~\ref{fig:mcq_scaling} shows that MCQ accuracy generally increases with model size, although the pattern varies by model family and language. Both Qwen2.5-Instruct and Qwen3 improve rapidly at smaller scales, with gains becoming more gradual beyond the 3--8B range. Qwen3 performs best at 32B, reaching 73.9\% in English and 71.4\% in Arabic, slightly above Qwen2.5-Instruct on the same scale. Jais-adapted shows a different pattern: scaling from 7B to 13B increases accuracy for Arabic from 49.2\% to 55.3\%, but not for English. 

English accuracy is consistently higher than Arabic accuracy across all three model families. This gap may reflect the stronger representation of English in pretraining data. At 32B, both Qwen model series approach the performance of the proprietary reference models. Overall, scaling improves culturally grounded MCQ performance, but model architecture and language coverage remain important factors.

\paragraph{Head-to-head win rates.}
Figure~\ref{fig:win_rate_compare} compares models trained on \ds+\dataset{CARE} with their \dataset{CARE}-only (DPO) or \dataset{Palm} (SFT) counterparts, using GPT-4o for pairwise evaluation. Under DPO, the \ds+\dataset{CARE} models receive more wins in five of the six model-language comparisons. The largest difference is observed for Jais in English, with 58.5\% wins for \ds+\dataset{CARE} and 30.3\% for \dataset{CARE}-only. Jais in Arabic is the exception: \dataset{CARE}-only receives more wins than \ds+\dataset{CARE} (49.6\% vs.\ 42.6\%). Under SFT, \ds+\dataset{CARE} is preferred in all six comparisons, although the difference for Qwen in English is small (48.2\% vs.\ 46.3\%). The largest SFT gains are again observed for Jais, which receives 82.8\% wins in English and 59.2\% in Arabic. These results indicate that \ds can complement \dataset{CARE} and \dataset{Palm}, but its effect varies across training objectives, models, and languages.

\begin{table}[t]
\centering
\scalebox{0.85}{%
\begin{tabular}{lccccc}
\toprule
Model & Language & \multicolumn{3}{c}{Training Data} \\
\cmidrule(lr){3-5}
& & English & Arabic & Whole \\
\midrule

\multirow{2}{*}{Llama}
& En
& \MonoCell{3.53}{3.78}
& \MonoCell{3.47}{3.78}
& \cellcolor{gray!10}3.78 \\

& Ar
& \MonoCell{3.17}{3.59}
& \MonoCell{3.08}{3.59}
& \cellcolor{gray!10}3.59 \\

\multirow{2}{*}{Qwen}
& En
& \MonoCell{4.56}{4.41}
& \MonoCell{4.63}{4.41}
& \cellcolor{gray!10}4.41 \\

& Ar
& \MonoCell{4.10}{4.02}
& \MonoCell{3.96}{4.02}
& \cellcolor{gray!10}4.02 \\

\multirow{2}{*}{Jais}
& En
& \MonoCell{4.34}{4.53}
& \MonoCell{4.30}{4.53}
& \cellcolor{gray!10}4.53 \\

& Ar
& \MonoCell{4.20}{4.31}
& \MonoCell{4.14}{4.31}
& \cellcolor{gray!10}4.31 \\

\bottomrule
\end{tabular}%
}
\caption{Model performance comparison with monolingual versus whole-dataset
training in preference generation. Cell colors compare English and Arabic
training against the corresponding Whole-training score. \textcolor{MonoPos}{Green} indicates an
improvement, \textcolor{MonoNeg}{red} indicates a decrease, and darker shades indicate larger
absolute differences.}
\label{tab:monolingual_vs_multilingual}
\end{table}

\paragraph{Monolingual vs. Multilingual Training}

The results demonstrate that the impact of training data composition is highly architecture-dependent. For Llama and Jais, the whole dataset consistently yields peak performance across both English and Arabic, suggesting these architectures benefit from cross-lingual transfer and broader cultural context. 

In contrast, the Qwen model exhibits a specialized sensitivity to data language: it achieves its higher score for English and Arabic via English-only training, with the combined dataset leading to marginal performance declines across both English and Arabic. This indicates that while models like Llama and Jais thrive on diverse multilingual exposure, Qwen is more susceptible to interference between language-specific specialization and the general-purpose distribution of the full corpus. Additional discussion on sampling efficiency is provided in Appendix~\ref{apdx:sample_efficiency}.

\section{Conclusion and Future Work}

We introduced \ds, a culturally grounded Arabic-centric preference dataset covering a total of 22 Arab countries, 16 categories, and 200 aspects. Our extensive post-training experiments demonstrated that models trained on \ds achieve stronger cultural alignment compared to the original models. By supporting both training and evaluation, \ds advances culturally-aware Arabic language modeling and helps models reflect regional values and norms.

In future work, we plan to expand coverage of regional dialects, code-switching, underrepresented subcultures, and within-country variation. We further plan to track how preferences evolve over time, and to evaluate multi-turn interactions shaped by speaker relationships, social settings, and conversational history. Broader annotator coverage and longitudinal data could further distinguish shared norms from context-specific preferences without homogenizing national cultures.

\section*{Limitations}
While \ds provides comprehensive coverage of cultural preferences across 22 Arab countries, it may not capture the full cultural diversity within each nation, especially at the regional or local level. The focus on behavioral preferences may overlook other important cultural dimensions, such as emotional expression, non-verbal communication, or specific context-dependent behaviors. Moreover, although the MCQ-based evaluation framework offers useful insights, it may not fully capture the complexity of cultural alignment in more dynamic, open-ended, or real-world scenarios where models need to respond flexibly to diverse and evolving cultural contexts. As such, further research is needed to develop more nuanced evaluation methodologies and expand the dataset to address these gaps.

\section*{Ethics and Broader Impact}

The development and use of \ds aim to enhance cultural alignment in Arabic language models, promoting more respectful and accurate interactions across diverse Arab contexts. However, it is important to acknowledge that cultural behaviors are dynamic and diverse, and any dataset, including \ds, can only represent a portion of the full spectrum of cultural practices. Ethical considerations must be taken into account when deploying these models, especially to avoid reinforcing stereotypes or biases. Furthermore, ongoing efforts are needed to ensure the dataset reflects evolving cultural norms and values, with an emphasis on inclusive and sensitive data collection.

\section*{Acknowledgments}
We thank the anonymous reviewers for their constructive feedback during the review process. This research was partially supported by the Dubai Research, Development and Innovation (RDI) Grant\footnote{\url{https://dubairdi.ae/}} and the MBZUAI supercomputing cluster.








\bibliography{custom}

@inproceedings{shi-etal-2024-culturebank,
    title = "{C}ulture{B}ank: An Online Community-Driven Knowledge Base Towards Culturally Aware Language Technologies",
    author = "Shi, Weiyan  and
      Li, Ryan  and
      Zhang, Yutong  and
      Ziems, Caleb  and
      Yu, Sunny  and
      Horesh, Raya  and
      Paula, Rog{\'e}rio Abreu De  and
      Yang, Diyi",
    editor = "Al-Onaizan, Yaser  and
      Bansal, Mohit  and
      Chen, Yun-Nung",
    booktitle = "Findings of the Association for Computational Linguistics: EMNLP 2024",
    month = nov,
    year = "2024",
    address = "Miami, Florida, USA",
    publisher = "Association for Computational Linguistics",
    url = "https://aclanthology.org/2024.findings-emnlp.288/",
    doi = "10.18653/v1/2024.findings-emnlp.288",
    pages = "4996--5025"
}

@misc{embedding_gemma_2025,
      title={{EmbeddingGemma}: Powerful and Lightweight Text Representations}, 
      author={Henrique Schechter Vera and Sahil Dua and Biao Zhang and Daniel Salz and Ryan Mullins and Sindhu Raghuram Panyam and Sara Smoot and Iftekhar Naim and Joe Zou and Feiyang Chen and Daniel Cer and Alice Lisak and Min Choi and Lucas Gonzalez and Omar Sanseviero and Glenn Cameron and Ian Ballantyne and Kat Black and Kaifeng Chen and Weiyi Wang and Zhe Li and Gus Martins and Jinhyuk Lee and Mark Sherwood and Juyeong Ji and Renjie Wu and Jingxiao Zheng and Jyotinder Singh and Abheesht Sharma and Divyashree Sreepathihalli and Aashi Jain and Adham Elarabawy and AJ Co and Andreas Doumanoglou and Babak Samari and Ben Hora and Brian Potetz and Dahun Kim and Enrique Alfonseca and Fedor Moiseev and Feng Han and Frank Palma Gomez and Gustavo Hernández Ábrego and Hesen Zhang and Hui Hui and Jay Han and Karan Gill and Ke Chen and Koert Chen and Madhuri Shanbhogue and Michael Boratko and Paul Suganthan and Sai Meher Karthik Duddu and Sandeep Mariserla and Setareh Ariafar and Shanfeng Zhang and Shijie Zhang and Simon Baumgartner and Sonam Goenka and Steve Qiu and Tanmaya Dabral and Trevor Walker and Vikram Rao and Waleed Khawaja and Wenlei Zhou and Xiaoqi Ren and Ye Xia and Yichang Chen and Yi-Ting Chen and Zhe Dong and Zhongli Ding and Francesco Visin and Gaël Liu and Jiageng Zhang and Kathleen Kenealy and Michelle Casbon and Ravin Kumar and Thomas Mesnard and Zach Gleicher and Cormac Brick and Olivier Lacombe and Adam Roberts and Qin Yin and Yunhsuan Sung and Raphael Hoffmann and Tris Warkentin and Armand Joulin and Tom Duerig and Mojtaba Seyedhosseini},
      year={2025},
      eprint={2509.20354},
      archivePrefix={arXiv},
      primaryClass={cs.CL},
      url={https://arxiv.org/abs/2509.20354}, 
}

@inproceedings{minhash,
  title = {On the resemblance and containment of documents},
  author = {Andrei Z. Broder},
  year = {1997},
  doi = {10.1109/SEQUEN.1997.666900},
  url = {https://doi.org/10.1109/SEQUEN.1997.666900},
  researchr = {https://researchr.org/publication/Broder97},
  cites = {0},
  citedby = {0},
  pages = {21-29},
  booktitle = {Compression and Complexity of SEQUENCES 1997, Positano, Amalfitan Coast, Salerno, Italy, June 11-13, 1997, Proceedings},
  editor = {Bruno Carpentieri and Alfredo De Santis and Ugo Vaccaro and James A. Storer},
  publisher = {IEEE},
  isbn = {0-8186-8133-0},
}

@inproceedings{naous-etal-2024-beer,
    title = "Having Beer after Prayer? {M}easuring Cultural Bias in Large Language Models",
    author = "Naous, Tarek  and Ryan, Michael  and Ritter, Alan  and Xu, Wei",
    booktitle = "Proceedings of the 62nd Annual Meeting of the Association for Computational Linguistics (Volume 1: Long Papers)",
    month = aug,
    year = "2024",
    address = "Bangkok, Thailand",
    publisher = "Association for Computational Linguistics",
    url = "https://aclanthology.org/2024.acl-long.862",
    pages = "16366--16393",
}

@article{rafailov2023direct,
  title={Direct preference optimization: Your language model is secretly a reward model},
  author={Rafailov, Rafael and Sharma, Archit and Mitchell, Eric and Manning, Christopher D and Ermon, Stefano and Finn, Chelsea},
  journal={Advances in Neural Information Processing Systems},
  series = {NuerIPS'23},
  address = {New Orleans, Louisiana, USA},
  volume={36},
  pages={53728--53741},
  year={2023}
}

@misc{shaikh2025aligning,
      title={Aligning Language Models with Demonstrated Feedback}, 
      author={Omar Shaikh and Michelle S. Lam and Joey Hejna and Yijia Shao and Hyundong Cho and Michael S. Bernstein and Diyi Yang},
      year={2025},
      eprint={2406.00888},
      archivePrefix={arXiv},
      primaryClass={cs.CL},
      url={https://arxiv.org/abs/2406.00888}, 
}

@article{Liu+2023+224+244,
url = {https://doi.org/10.1515/jtc-2023-0019},
title = {Cultural Bias in Large Language Models: {A} Comprehensive Analysis and Mitigation Strategies},
title = {},
author = {Zhaoming Liu},
pages = {224--244},
volume = {3},
number = {2},
journal = {Journal of Transcultural Communication},
doi = {doi:10.1515/jtc-2023-0019},
year = {2023},
lastchecked = {2026-08-29}
}

@inproceedings{al-kautsar-etal-2026-cultural,
    title = "Cultural Benchmarking of {LLM}s in Standard and Dialectal {A}rabic Dialogues",
    author = "Al Kautsar, Muhammad Dehan  and
      Almheiri, Saeed  and
      Ahsan, Momina  and
      Elbouardi, Bilal  and
      Samih, Younes  and
      Ahmad, Sarfraz  and
      Keleg, Amr  and
      El Herraoui, Omar  and
      Elzeky, Kareem  and
      Freihat, Abed Alhakim  and
      Anwar, Mohamed  and
      Xie, Zhuohan  and
      Liang, Junhong  and
      Al Nasar, Mohammad Rustom  and
      Nakov, Preslav  and
      Koto, Fajri",
    editor = "Liakata, Maria  and
      Moreira, Viviane P.  and
      Zhang, Jiajun  and
      Jurgens, David",
    booktitle = "Proceedings of the 64th Annual Meeting of the {A}ssociation for {C}omputational {L}inguistics (Volume 1: Long Papers)",
    month = jul,
    year = "2026",
    address = "San Diego, California, United States",
    publisher = "Association for Computational Linguistics",
    url = "https://aclanthology.org/2026.acl-long.963/",
    doi = "10.18653/v1/2026.acl-long.963",
    pages = "21016--21039",
    ISBN = "979-8-89176-390-6"
}

@inproceedings{khan2025,
author = {Khan, Ariba and Casper, Stephen and Hadfield-Menell, Dylan},
title = {Randomness, Not Representation: The Unreliability of Evaluating Cultural Alignment in {LLMs}},
year = {2025},
isbn = {9798400714825},
publisher = {Association for Computing Machinery},
address = {New York, NY, USA},
url = {https://doi.org/10.1145/3715275.3732147},
doi = {10.1145/3715275.3732147},
booktitle = {Proceedings of the 2025 ACM Conference on Fairness, Accountability, and Transparency},
pages = {2151–2165},
numpages = {15},
location = {
},
series = {FAccT '25}
}

@article{Liu2026, 
author = {Jianing Liu and Bing Song and Vinayak Dixit and Chenyang Wu and Sisi Jian},
title = {Can large language models capture human risk preferences? {A} cross-cultural study},
year = {2026},
journal = {Communications in Transportation Research},
volume = {6},
number = {2},
pages = {9640025},
url = {https://www.sciopen.com/article/10.26599/COMMTR.2026.9640025},
doi = {10.26599/COMMTR.2026.9640025}
}

@inproceedings{ahuja-etal-2023-mega,
    title = "{MEGA}: Multilingual Evaluation of Generative {AI}",
    author = "Ahuja, Kabir  and
      Diddee, Harshita  and
      Hada, Rishav  and
      Ochieng, Millicent  and
      Ramesh, Krithika  and
      Jain, Prachi  and
      Nambi, Akshay  and
      Ganu, Tanuja  and
      Segal, Sameer  and
      Ahmed, Mohamed  and
      Bali, Kalika  and
      Sitaram, Sunayana",
    editor = "Bouamor, Houda  and
      Pino, Juan  and
      Bali, Kalika",
    booktitle = "Proceedings of the 2023 Conference on Empirical Methods in Natural Language Processing",
    month = dec,
    year = "2023",
    address = "Singapore",
    publisher = "Association for Computational Linguistics",
    url = "https://aclanthology.org/2023.emnlp-main.258/",
    doi = "10.18653/v1/2023.emnlp-main.258",
    pages = "4232--4267"
}

@inproceedings{yong-etal-2023-bloom,
    title = "{BLOOM}+1: Adding Language Support to {BLOOM} for Zero-Shot Prompting",
    author = "Yong, Zheng Xin  and
      Schoelkopf, Hailey  and
      Muennighoff, Niklas  and
      Aji, Alham Fikri  and
      Adelani, David Ifeoluwa  and
      Almubarak, Khalid  and
      Bari, M Saiful  and
      Sutawika, Lintang  and
      Kasai, Jungo  and
      Baruwa, Ahmed  and
      Winata, Genta  and
      Biderman, Stella  and
      Raff, Edward  and
      Radev, Dragomir  and
      Nikoulina, Vassilina",
    editor = "Rogers, Anna  and
      Boyd-Graber, Jordan  and
      Okazaki, Naoaki",
    booktitle = "Proceedings of the 61st Annual Meeting of the Association for Computational Linguistics (Volume 1: Long Papers)",
    month = jul,
    year = "2023",
    address = "Toronto, Canada",
    publisher = "Association for Computational Linguistics",
    url = "https://aclanthology.org/2023.acl-long.653/",
    doi = "10.18653/v1/2023.acl-long.653",
    pages = "11682--11703"
}

@inproceedings{hershcovich-etal-2022-challenges,
    title = "Challenges and Strategies in Cross-Cultural {NLP}",
    author = "Hershcovich, Daniel  and
      Frank, Stella  and
      Lent, Heather  and
      de Lhoneux, Miryam  and
      Abdou, Mostafa  and
      Brandl, Stephanie  and
      Bugliarello, Emanuele  and
      Cabello Piqueras, Laura  and
      Chalkidis, Ilias  and
      Cui, Ruixiang  and
      Fierro, Constanza  and
      Margatina, Katerina  and
      Rust, Phillip  and
      S{\o}gaard, Anders",
    editor = "Muresan, Smaranda  and
      Nakov, Preslav  and
      Villavicencio, Aline",
    booktitle = "Proceedings of the 60th Annual Meeting of the Association for Computational Linguistics (Volume 1: Long Papers)",
    month = may,
    year = "2022",
    address = "Dublin, Ireland",
    publisher = "Association for Computational Linguistics",
    url = "https://aclanthology.org/2022.acl-long.482/",
    doi = "10.18653/v1/2022.acl-long.482",
    pages = "6997--7013"
}

@inproceedings{arora-etal-2023-probing,
    title = "Probing Pre-Trained Language Models for Cross-Cultural Differences in Values",
    author = "Arora, Arnav  and
      Kaffee, Lucie-aim{\'e}e  and
      Augenstein, Isabelle",
    editor = "Dev, Sunipa  and
      Prabhakaran, Vinodkumar  and
      Adelani, David Ifeoluwa  and
      Hovy, Dirk  and
      Benotti, Luciana",
    booktitle = "Proceedings of the First Workshop on Cross-Cultural Considerations in NLP (C3NLP)",
    month = may,
    year = "2023",
    address = "Dubrovnik, Croatia",
    publisher = "Association for Computational Linguistics",
    url = "https://aclanthology.org/2023.c3nlp-1.12/",
    doi = "10.18653/v1/2023.c3nlp-1.12",
    pages = "114--130"
}

@misc{openai2024gpt4technicalreport,
      title={{GPT-4} Technical Report}, 
      author={OpenAI and Josh Achiam and Steven Adler and Sandhini Agarwal and Lama Ahmad and Ilge Akkaya and Florencia Leoni Aleman and Diogo Almeida and Janko Altenschmidt and Sam Altman and Shyamal Anadkat and Red Avila and Igor Babuschkin and Suchir Balaji and Valerie Balcom and Paul Baltescu and Haiming Bao and Mohammad Bavarian and Jeff Belgum and Irwan Bello and Jake Berdine and Gabriel Bernadett-Shapiro and Christopher Berner and Lenny Bogdonoff and Oleg Boiko and Madelaine Boyd and Anna-Luisa Brakman and Greg Brockman and Tim Brooks and Miles Brundage and Kevin Button and Trevor Cai and Rosie Campbell and Andrew Cann and Brittany Carey and Chelsea Carlson and Rory Carmichael and Brooke Chan and Che Chang and Fotis Chantzis and Derek Chen and Sully Chen and Ruby Chen and Jason Chen and Mark Chen and Ben Chess and Chester Cho and Casey Chu and Hyung Won Chung and Dave Cummings and Jeremiah Currier and Yunxing Dai and Cory Decareaux and Thomas Degry and Noah Deutsch and Damien Deville and Arka Dhar and David Dohan and Steve Dowling and Sheila Dunning and Adrien Ecoffet and Atty Eleti and Tyna Eloundou and David Farhi and Liam Fedus and Niko Felix and Simón Posada Fishman and Juston Forte and Isabella Fulford and Leo Gao and Elie Georges and Christian Gibson and Vik Goel and Tarun Gogineni and Gabriel Goh and Rapha Gontijo-Lopes and Jonathan Gordon and Morgan Grafstein and Scott Gray and Ryan Greene and Joshua Gross and Shixiang Shane Gu and Yufei Guo and Chris Hallacy and Jesse Han and Jeff Harris and Yuchen He and Mike Heaton and Johannes Heidecke and Chris Hesse and Alan Hickey and Wade Hickey and Peter Hoeschele and Brandon Houghton and Kenny Hsu and Shengli Hu and Xin Hu and Joost Huizinga and Shantanu Jain and Shawn Jain and Joanne Jang and Angela Jiang and Roger Jiang and Haozhun Jin and Denny Jin and Shino Jomoto and Billie Jonn and Heewoo Jun and Tomer Kaftan and Łukasz Kaiser and Ali Kamali and Ingmar Kanitscheider and Nitish Shirish Keskar and Tabarak Khan and Logan Kilpatrick and Jong Wook Kim and Christina Kim and Yongjik Kim and Jan Hendrik Kirchner and Jamie Kiros and Matt Knight and Daniel Kokotajlo and Łukasz Kondraciuk and Andrew Kondrich and Aris Konstantinidis and Kyle Kosic and Gretchen Krueger and Vishal Kuo and Michael Lampe and Ikai Lan and Teddy Lee and Jan Leike and Jade Leung and Daniel Levy and Chak Ming Li and Rachel Lim and Molly Lin and Stephanie Lin and Mateusz Litwin and Theresa Lopez and Ryan Lowe and Patricia Lue and Anna Makanju and Kim Malfacini and Sam Manning and Todor Markov and Yaniv Markovski and Bianca Martin and Katie Mayer and Andrew Mayne and Bob McGrew and Scott Mayer McKinney and Christine McLeavey and Paul McMillan and Jake McNeil and David Medina and Aalok Mehta and Jacob Menick and Luke Metz and Andrey Mishchenko and Pamela Mishkin and Vinnie Monaco and Evan Morikawa and Daniel Mossing and Tong Mu and Mira Murati and Oleg Murk and David Mély and Ashvin Nair and Reiichiro Nakano and Rajeev Nayak and Arvind Neelakantan and Richard Ngo and Hyeonwoo Noh and Long Ouyang and Cullen O'Keefe and Jakub Pachocki and Alex Paino and Joe Palermo and Ashley Pantuliano and Giambattista Parascandolo and Joel Parish and Emy Parparita and Alex Passos and Mikhail Pavlov and Andrew Peng and Adam Perelman and Filipe de Avila Belbute Peres and Michael Petrov and Henrique Ponde de Oliveira Pinto and Michael and Pokorny and Michelle Pokrass and Vitchyr H. Pong and Tolly Powell and Alethea Power and Boris Power and Elizabeth Proehl and Raul Puri and Alec Radford and Jack Rae and Aditya Ramesh and Cameron Raymond and Francis Real and Kendra Rimbach and Carl Ross and Bob Rotsted and Henri Roussez and Nick Ryder and Mario Saltarelli and Ted Sanders and Shibani Santurkar and Girish Sastry and Heather Schmidt and David Schnurr and John Schulman and Daniel Selsam and Kyla Sheppard and Toki Sherbakov and Jessica Shieh and Sarah Shoker and Pranav Shyam and Szymon Sidor and Eric Sigler and Maddie Simens and Jordan Sitkin and Katarina Slama and Ian Sohl and Benjamin Sokolowsky and Yang Song and Natalie Staudacher and Felipe Petroski Such and Natalie Summers and Ilya Sutskever and Jie Tang and Nikolas Tezak and Madeleine B. Thompson and Phil Tillet and Amin Tootoonchian and Elizabeth Tseng and Preston Tuggle and Nick Turley and Jerry Tworek and Juan Felipe Cerón Uribe and Andrea Vallone and Arun Vijayvergiya and Chelsea Voss and Carroll Wainwright and Justin Jay Wang and Alvin Wang and Ben Wang and Jonathan Ward and Jason Wei and CJ Weinmann and Akila Welihinda and Peter Welinder and Jiayi Weng and Lilian Weng and Matt Wiethoff and Dave Willner and Clemens Winter and Samuel Wolrich and Hannah Wong and Lauren Workman and Sherwin Wu and Jeff Wu and Michael Wu and Kai Xiao and Tao Xu and Sarah Yoo and Kevin Yu and Qiming Yuan and Wojciech Zaremba and Rowan Zellers and Chong Zhang and Marvin Zhang and Shengjia Zhao and Tianhao Zheng and Juntang Zhuang and William Zhuk and Barret Zoph},
      year={2024},
      eprint={2303.08774},
      archivePrefix={arXiv},
      primaryClass={cs.CL},
      url={https://arxiv.org/abs/2303.08774}, 
}

@article{johnson2022ghost,
  title={The Ghost in the Machine has an {A}merican accent: value conflict in {GPT-3}},
  author={Johnson, Rebecca L and Pistilli, Giada and Men{\'e}dez-Gonz{\'a}lez, Natalia and Duran, Leslye Denisse Dias and Panai, Enrico and Kalpokiene, Julija and Bertulfo, Donald Jay},
  journal={arXiv preprint arXiv:2203.07785},
  year={2022}
}

@inproceedings{abdulhai-etal-2024-moral,
    title = "Moral Foundations of Large Language Models",
    author = "Abdulhai, Marwa  and
      Serapio-Garc{\'i}a, Gregory  and
      Crepy, Clement  and
      Valter, Daria  and
      Canny, John  and
      Jaques, Natasha",
    editor = "Al-Onaizan, Yaser  and
      Bansal, Mohit  and
      Chen, Yun-Nung",
    booktitle = "Proceedings of the 2024 Conference on Empirical Methods in Natural Language Processing",
    month = nov,
    year = "2024",
    address = "Miami, Florida, USA",
    publisher = "Association for Computational Linguistics",
    url = "https://aclanthology.org/2024.emnlp-main.982/",
    doi = "10.18653/v1/2024.emnlp-main.982",
    pages = "17737--17752"
}

@inproceedings{nguyen2023extracting,
author = {Nguyen, Tuan-Phong and Razniewski, Simon and Varde, Aparna and Weikum, Gerhard},
title = {Extracting Cultural Commonsense Knowledge at Scale},
year = {2023},
isbn = {9781450394161},
publisher = {Association for Computing Machinery},
NOaddress = {New York, NY, USA},
url = {https://doi.org/10.1145/3543507.3583535},
doi = {10.1145/3543507.3583535},
booktitle = {Proceedings of the ACM Web Conference 2023},
pages = {1907–1917},
numpages = {11},
address = {Austin, TX, USA},
series = {WWW '23}
}

@inproceedings{ethayarajh2024kto,
author = {Ethayarajh, Kawin and Xu, Winnie and Muennighoff, Niklas and Jurafsky, Dan and Kiela, Douwe},
title = {Model alignment as prospect theoretic optimization},
year = {2024},
publisher = {JMLR.org},
booktitle = {Proceedings of the 41st International Conference on Machine Learning},
articleno = {504},
numpages = {18},
address = {Vienna, Austria},
series = {ICML'24}
}

@inproceedings{guo-etal-2025-care,
    title = "{CARE}: Multilingual Human Preference Learning for Cultural Awareness",
    author = "Guo, Geyang  and
      Naous, Tarek  and
      Wakaki, Hiromi  and
      Nishimura, Yukiko  and
      Mitsufuji, Yuki  and
      Ritter, Alan  and
      Xu, Wei",
    editor = "Christodoulopoulos, Christos  and
      Chakraborty, Tanmoy  and
      Rose, Carolyn  and
      Peng, Violet",
    booktitle = "Proceedings of the 2025 Conference on Empirical Methods in Natural Language Processing",
    month = nov,
    year = "2025",
    address = "Suzhou, China",
    publisher = "Association for Computational Linguistics",
    url = "https://aclanthology.org/2025.emnlp-main.1669/",
    doi = "10.18653/v1/2025.emnlp-main.1669",
    pages = "32866--32895",
    ISBN = "979-8-89176-332-6"
}

@article{sengupta2023jais,
  author  = {Sengupta, Neha and Sahu, Sunil Kumar and Jia, Bokang and
             Katipomu, Satheesh and Li, Haonan and Koto, Fajri and
             Marshall, William and Gosal, Gurpreet and Liu, Cynthia and
             Chen, Zhiming and Afzal, Osama Mohammed and Kamboj, Samta and
             Pandit, Onkar and Pal, Rahul and Pradhan, Lalit and
             Mujahid, Zain Muhammad and Baali, Massa and Han, Xudong and
             Mahmoud Bsharat, Sondos and Aji, Alham Fikri and Shen, Zhiqiang and
             Liu, Zhengzhong and Vassilieva, Natalia and Hestness, Joel and
             Hock, Andy and Feldman, Andrew and Lee, Jonathan and Jackson, Andrew and
             Ren, Hector Xuguang and Nakov, Preslav and Baldwin, Timothy and
             Xing, Eric},
  title   = {{Jais} and {Jais-chat}: {Arabic}-Centric Foundation and
             Instruction-Tuned Open Generative Large Language Models},
  journal = {arXiv preprint arXiv:2308.16149},
  year    = {2023},
  url     = {https://arxiv.org/abs/2308.16149}
}

@article{jaisfamilymodelcard,
    title={Jais Family Model Card},
    author={Inception},
    year={2024},
    url = {https://huggingface.co/inceptionai/jais-family-30b-16k-chat/blob/main/README.md}
}

@misc{qwen2025qwen25technicalreport,
      title={{Qwen2.5} Technical Report}, 
      author={Qwen and : and An Yang and Baosong Yang and Beichen Zhang and Binyuan Hui and Bo Zheng and Bowen Yu and Chengyuan Li and Dayiheng Liu and Fei Huang and Haoran Wei and Huan Lin and Jian Yang and Jianhong Tu and Jianwei Zhang and Jianxin Yang and Jiaxi Yang and Jingren Zhou and Junyang Lin and Kai Dang and Keming Lu and Keqin Bao and Kexin Yang and Le Yu and Mei Li and Mingfeng Xue and Pei Zhang and Qin Zhu and Rui Men and Runji Lin and Tianhao Li and Tianyi Tang and Tingyu Xia and Xingzhang Ren and Xuancheng Ren and Yang Fan and Yang Su and Yichang Zhang and Yu Wan and Yuqiong Liu and Zeyu Cui and Zhenru Zhang and Zihan Qiu},
      year={2025},
      eprint={2412.15115},
      archivePrefix={arXiv},
      primaryClass={cs.CL},
      url={https://arxiv.org/abs/2412.15115}, 
}

@misc{grattafiori2024llama3herdmodels,
      title={The {Llama 3} Herd of Models}, 
      author={Aaron Grattafiori and Abhimanyu Dubey and Abhinav Jauhri and Abhinav Pandey and Abhishek Kadian and Ahmad Al-Dahle and Aiesha Letman and Akhil Mathur and Alan Schelten and Alex Vaughan and Amy Yang and Angela Fan and Anirudh Goyal and Anthony Hartshorn and Aobo Yang and Archi Mitra and Archie Sravankumar and Artem Korenev and Arthur Hinsvark and Arun Rao and Aston Zhang and Aurelien Rodriguez and Austen Gregerson and Ava Spataru and Baptiste Roziere and Bethany Biron and Binh Tang and Bobbie Chern and Charlotte Caucheteux and Chaya Nayak and Chloe Bi and Chris Marra and Chris McConnell and Christian Keller and Christophe Touret and Chunyang Wu and Corinne Wong and Cristian Canton Ferrer and Cyrus Nikolaidis and Damien Allonsius and Daniel Song and Danielle Pintz and Danny Livshits and Danny Wyatt and David Esiobu and Dhruv Choudhary and Dhruv Mahajan and Diego Garcia-Olano and Diego Perino and Dieuwke Hupkes and Egor Lakomkin and Ehab AlBadawy and Elina Lobanova and Emily Dinan and Eric Michael Smith and Filip Radenovic and Francisco Guzmán and Frank Zhang and Gabriel Synnaeve and Gabrielle Lee and Georgia Lewis Anderson and Govind Thattai and Graeme Nail and Gregoire Mialon and Guan Pang and Guillem Cucurell and Hailey Nguyen and Hannah Korevaar and Hu Xu and Hugo Touvron and Iliyan Zarov and Imanol Arrieta Ibarra and Isabel Kloumann and Ishan Misra and Ivan Evtimov and Jack Zhang and Jade Copet and Jaewon Lee and Jan Geffert and Jana Vranes and Jason Park and Jay Mahadeokar and Jeet Shah and Jelmer van der Linde and Jennifer Billock and Jenny Hong and Jenya Lee and Jeremy Fu and Jianfeng Chi and Jianyu Huang and Jiawen Liu and Jie Wang and Jiecao Yu and Joanna Bitton and Joe Spisak and Jongsoo Park and Joseph Rocca and Joshua Johnstun and Joshua Saxe and Junteng Jia and Kalyan Vasuden Alwala and Karthik Prasad and Kartikeya Upasani and Kate Plawiak and Ke Li and Kenneth Heafield and Kevin Stone and Khalid El-Arini and Krithika Iyer and Kshitiz Malik and Kuenley Chiu and Kunal Bhalla and Kushal Lakhotia and Lauren Rantala-Yeary and Laurens van der Maaten and Lawrence Chen and Liang Tan and Liz Jenkins and Louis Martin and Lovish Madaan and Lubo Malo and Lukas Blecher and Lukas Landzaat and Luke de Oliveira and Madeline Muzzi and Mahesh Pasupuleti and Mannat Singh and Manohar Paluri and Marcin Kardas and Maria Tsimpoukelli and Mathew Oldham and Mathieu Rita and Maya Pavlova and Melanie Kambadur and Mike Lewis and Min Si and Mitesh Kumar Singh and Mona Hassan and Naman Goyal and Narjes Torabi and Nikolay Bashlykov and Nikolay Bogoychev and Niladri Chatterji and Ning Zhang and Olivier Duchenne and Onur Çelebi and Patrick Alrassy and Pengchuan Zhang and Pengwei Li and Petar Vasic and Peter Weng and Prajjwal Bhargava and Pratik Dubal and Praveen Krishnan and Punit Singh Koura and Puxin Xu and Qing He and Qingxiao Dong and Ragavan Srinivasan and Raj Ganapathy and Ramon Calderer and Ricardo Silveira Cabral and Robert Stojnic and Roberta Raileanu and Rohan Maheswari and Rohit Girdhar and Rohit Patel and Romain Sauvestre and Ronnie Polidoro and Roshan Sumbaly and Ross Taylor and Ruan Silva and Rui Hou and Rui Wang and Saghar Hosseini and Sahana Chennabasappa and Sanjay Singh and Sean Bell and Seohyun Sonia Kim and Sergey Edunov and Shaoliang Nie and Sharan Narang and Sharath Raparthy and Sheng Shen and Shengye Wan and Shruti Bhosale and Shun Zhang and Simon Vandenhende and Soumya Batra and Spencer Whitman and Sten Sootla and Stephane Collot and Suchin Gururangan and Sydney Borodinsky and Tamar Herman and Tara Fowler and Tarek Sheasha and Thomas Georgiou and Thomas Scialom and Tobias Speckbacher and Todor Mihaylov and Tong Xiao and Ujjwal Karn and Vedanuj Goswami and Vibhor Gupta and Vignesh Ramanathan and Viktor Kerkez and Vincent Gonguet and Virginie Do and Vish Vogeti and Vítor Albiero and Vladan Petrovic and Weiwei Chu and Wenhan Xiong and Wenyin Fu and Whitney Meers and Xavier Martinet and Xiaodong Wang and Xiaofang Wang and Xiaoqing Ellen Tan and Xide Xia and Xinfeng Xie and Xuchao Jia and Xuewei Wang and Yaelle Goldschlag and Yashesh Gaur and Yasmine Babaei and Yi Wen and Yiwen Song and Yuchen Zhang and Yue Li and Yuning Mao and Zacharie Delpierre Coudert and Zheng Yan and Zhengxing Chen and Zoe Papakipos and Aaditya Singh and Aayushi Srivastava and Abha Jain and Adam Kelsey and Adam Shajnfeld and Adithya Gangidi and Adolfo Victoria and Ahuva Goldstand and Ajay Menon and Ajay Sharma and Alex Boesenberg and Alexei Baevski and Allie Feinstein and Amanda Kallet and Amit Sangani and Amos Teo and Anam Yunus and Andrei Lupu and Andres Alvarado and Andrew Caples and Andrew Gu and Andrew Ho and Andrew Poulton and Andrew Ryan and Ankit Ramchandani and Annie Dong and Annie Franco and Anuj Goyal and Aparajita Saraf and Arkabandhu Chowdhury and Ashley Gabriel and Ashwin Bharambe and Assaf Eisenman and Azadeh Yazdan and Beau James and Ben Maurer and Benjamin Leonhardi and Bernie Huang and Beth Loyd and Beto De Paola and Bhargavi Paranjape and Bing Liu and Bo Wu and Boyu Ni and Braden Hancock and Bram Wasti and Brandon Spence and Brani Stojkovic and Brian Gamido and Britt Montalvo and Carl Parker and Carly Burton and Catalina Mejia and Ce Liu and Changhan Wang and Changkyu Kim and Chao Zhou and Chester Hu and Ching-Hsiang Chu and Chris Cai and Chris Tindal and Christoph Feichtenhofer and Cynthia Gao and Damon Civin and Dana Beaty and Daniel Kreymer and Daniel Li and David Adkins and David Xu and Davide Testuggine and Delia David and Devi Parikh and Diana Liskovich and Didem Foss and Dingkang Wang and Duc Le and Dustin Holland and Edward Dowling and Eissa Jamil and Elaine Montgomery and Eleonora Presani and Emily Hahn and Emily Wood and Eric-Tuan Le and Erik Brinkman and Esteban Arcaute and Evan Dunbar and Evan Smothers and Fei Sun and Felix Kreuk and Feng Tian and Filippos Kokkinos and Firat Ozgenel and Francesco Caggioni and Frank Kanayet and Frank Seide and Gabriela Medina Florez and Gabriella Schwarz and Gada Badeer and Georgia Swee and Gil Halpern and Grant Herman and Grigory Sizov and Guangyi and Zhang and Guna Lakshminarayanan and Hakan Inan and Hamid Shojanazeri and Han Zou and Hannah Wang and Hanwen Zha and Haroun Habeeb and Harrison Rudolph and Helen Suk and Henry Aspegren and Hunter Goldman and Hongyuan Zhan and Ibrahim Damlaj and Igor Molybog and Igor Tufanov and Ilias Leontiadis and Irina-Elena Veliche and Itai Gat and Jake Weissman and James Geboski and James Kohli and Janice Lam and Japhet Asher and Jean-Baptiste Gaya and Jeff Marcus and Jeff Tang and Jennifer Chan and Jenny Zhen and Jeremy Reizenstein and Jeremy Teboul and Jessica Zhong and Jian Jin and Jingyi Yang and Joe Cummings and Jon Carvill and Jon Shepard and Jonathan McPhie and Jonathan Torres and Josh Ginsburg and Junjie Wang and Kai Wu and Kam Hou U and Karan Saxena and Kartikay Khandelwal and Katayoun Zand and Kathy Matosich and Kaushik Veeraraghavan and Kelly Michelena and Keqian Li and Kiran Jagadeesh and Kun Huang and Kunal Chawla and Kyle Huang and Lailin Chen and Lakshya Garg and Lavender A and Leandro Silva and Lee Bell and Lei Zhang and Liangpeng Guo and Licheng Yu and Liron Moshkovich and Luca Wehrstedt and Madian Khabsa and Manav Avalani and Manish Bhatt and Martynas Mankus and Matan Hasson and Matthew Lennie and Matthias Reso and Maxim Groshev and Maxim Naumov and Maya Lathi and Meghan Keneally and Miao Liu and Michael L. Seltzer and Michal Valko and Michelle Restrepo and Mihir Patel and Mik Vyatskov and Mikayel Samvelyan and Mike Clark and Mike Macey and Mike Wang and Miquel Jubert Hermoso and Mo Metanat and Mohammad Rastegari and Munish Bansal and Nandhini Santhanam and Natascha Parks and Natasha White and Navyata Bawa and Nayan Singhal and Nick Egebo and Nicolas Usunier and Nikhil Mehta and Nikolay Pavlovich Laptev and Ning Dong and Norman Cheng and Oleg Chernoguz and Olivia Hart and Omkar Salpekar and Ozlem Kalinli and Parkin Kent and Parth Parekh and Paul Saab and Pavan Balaji and Pedro Rittner and Philip Bontrager and Pierre Roux and Piotr Dollar and Polina Zvyagina and Prashant Ratanchandani and Pritish Yuvraj and Qian Liang and Rachad Alao and Rachel Rodriguez and Rafi Ayub and Raghotham Murthy and Raghu Nayani and Rahul Mitra and Rangaprabhu Parthasarathy and Raymond Li and Rebekkah Hogan and Robin Battey and Rocky Wang and Russ Howes and Ruty Rinott and Sachin Mehta and Sachin Siby and Sai Jayesh Bondu and Samyak Datta and Sara Chugh and Sara Hunt and Sargun Dhillon and Sasha Sidorov and Satadru Pan and Saurabh Mahajan and Saurabh Verma and Seiji Yamamoto and Sharadh Ramaswamy and Shaun Lindsay and Shaun Lindsay and Sheng Feng and Shenghao Lin and Shengxin Cindy Zha and Shishir Patil and Shiva Shankar and Shuqiang Zhang and Shuqiang Zhang and Sinong Wang and Sneha Agarwal and Soji Sajuyigbe and Soumith Chintala and Stephanie Max and Stephen Chen and Steve Kehoe and Steve Satterfield and Sudarshan Govindaprasad and Sumit Gupta and Summer Deng and Sungmin Cho and Sunny Virk and Suraj Subramanian and Sy Choudhury and Sydney Goldman and Tal Remez and Tamar Glaser and Tamara Best and Thilo Koehler and Thomas Robinson and Tianhe Li and Tianjun Zhang and Tim Matthews and Timothy Chou and Tzook Shaked and Varun Vontimitta and Victoria Ajayi and Victoria Montanez and Vijai Mohan and Vinay Satish Kumar and Vishal Mangla and Vlad Ionescu and Vlad Poenaru and Vlad Tiberiu Mihailescu and Vladimir Ivanov and Wei Li and Wenchen Wang and Wenwen Jiang and Wes Bouaziz and Will Constable and Xiaocheng Tang and Xiaojian Wu and Xiaolan Wang and Xilun Wu and Xinbo Gao and Yaniv Kleinman and Yanjun Chen and Ye Hu and Ye Jia and Ye Qi and Yenda Li and Yilin Zhang and Ying Zhang and Yossi Adi and Youngjin Nam and Yu and Wang and Yu Zhao and Yuchen Hao and Yundi Qian and Yunlu Li and Yuzi He and Zach Rait and Zachary DeVito and Zef Rosnbrick and Zhaoduo Wen and Zhenyu Yang and Zhiwei Zhao and Zhiyu Ma},
      year={2024},
      eprint={2407.21783},
      archivePrefix={arXiv},
      primaryClass={cs.AI},
      url={https://arxiv.org/abs/2407.21783}, 
}

@inproceedings{alwajih-etal-2025-palm,
    title = "{Palm}: A Culturally Inclusive and Linguistically Diverse Dataset for {A}rabic {LLM}s",
    author = "Alwajih, Fakhraddin and El Mekki, Abdellah and Magdy, Samar Mohamed and
              Elmadany, AbdelRahim A. and Nacar, Omer and Nagoudi, El Moatez Billah and
              Abdel-Salam, Reem and Atwany, Hanin and Nafea, Youssef and
              Yahya, Abdulfattah Mohammed and Alhamouri, Rahaf and Alsayadi, Hamzah A. and
              Zayed, Hiba and Shatnawi, Sara and Sibaee, Serry and Ech-chammakhy, Yasir and
              Al-Dhabyani, Walid and Ali, Marwa Mohamed and Jarraya, Imen and
              El-Shangiti, Ahmed Oumar and Alraeesi, Aisha and AL-Ghrawi, Mohammed Anwar and
              Al-Batati, Abdulrahman S. and Mohamed, Elgizouli and Elgindi, Noha Taha and
              Saeed, Muhammed and Atou, Houdaifa and Yahia, Issam Ait and Bouayad, Abdelhak and
              Machrouh, Mohammed and Makouar, Amal and Alkawi, Dania and Mohamed, Mukhtar and
              Abdelfadil, Safaa Taher and Ounnoughene, Amine Ziad and Rouabhia, Anfel and
              Assi, Rwaa and Sorkatti, Ahmed and Tourad, Mohamedou Cheikh and Koubaa, Anis and
              Berrada, Ismail and Jarrar, Mustafa and Shehata, Shady and Abdul-Mageed, Muhammad",
    booktitle = "Proceedings of the 63rd Annual Meeting of the Association for Computational Linguistics (Volume 1: Long Papers)",
    month = jul,
    year = "2025",
    address = "Vienna, Austria",
    publisher = "Association for Computational Linguistics",
    url = "https://aclanthology.org/2025.acl-long.1579/",
    pages = "32871--32894",
    ISBN = "979-8-89176-251-0"
}

@inproceedings{alyafeai2024cidar,
    title = {{CIDAR}: Culturally Relevant Instruction Dataset For {A}rabic},
    author = {Alyafeai, Zaid and Almubarak, Khalid and Ashraf, Ahmed and Alnuhait, Deema and Alshahrani, Saied and Abdulrahman, Gubran A. Q. and Ahmed, Gamil and Gawah, Qais and Saleh, Zead and Ghaleb, Mustafa and Ali, Yousef and Al-Shaibani, Maged S.},
    booktitle = {Findings of the Association for Computational Linguistics: ACL 2024},
    year = {2024},
    publisher = {Association for Computational Linguistics},
    address = {Bangkok, Thailand},
    pages = {12878--12901},
    url = {https://aclanthology.org/2024.findings-acl.764/}
}

@inproceedings{sadallah2025arabculture,
    title = {Commonsense Reasoning in {A}rab Culture},
    author = {Sadallah, Abdelrahman and Tonga, Junior Cedric and Almubarak, Khalid and Almheiri, Saeed and Atif, Farah and Qwaider, Chatrine and Kadaoui, Karima and Shatnawi, Sara and Alesh, Yaser and Koto, Fajri},
    booktitle = {Proceedings of the 63rd Annual Meeting of the Association for Computational Linguistics (Volume 1: Long Papers)},
    year = {2025},
    publisher = {Association for Computational Linguistics},
    address = {Vienna, Austria},
    pages = {7695--7710},
    url = {https://aclanthology.org/2025.acl-long.380/}
}

@inproceedings{cao-etal-2023-assessing,
    title = "Assessing Cross-Cultural Alignment between {C}hat{GPT} and Human Societies: An Empirical Study",
    author = "Cao, Yong  and
      Zhou, Li  and
      Lee, Seolhwa  and
      Cabello, Laura  and
      Chen, Min  and
      Hershcovich, Daniel",
    editor = "Dev, Sunipa  and
      Prabhakaran, Vinodkumar  and
      Adelani, David Ifeoluwa  and
      Hovy, Dirk  and
      Benotti, Luciana",
    booktitle = "Proceedings of the First Workshop on Cross-Cultural Considerations in NLP (C3NLP)",
    month = may,
    year = "2023",
    address = "Dubrovnik, Croatia",
    publisher = "Association for Computational Linguistics",
    url = "https://aclanthology.org/2023.c3nlp-1.7/",
    doi = "10.18653/v1/2023.c3nlp-1.7",
    pages = "53--67"
}

@article{dokic2025mirroring,
    title = {Mirroring Cultural Dominance: Disclosing Large Language Models Social Values, Attitudes and Stereotypes},
    author = {Dokic, Kristian and Pisker, Barbara and Radisic, Bojan},
    journal = {Societies},
    volume = {15},
    number = {5},
    article-number = {142},
    year = {2025},
    publisher = {MDPI},
    doi = {10.3390/soc15050142},
    url = {https://www.mdpi.com/2075-4698/15/5/142}
}

@inproceedings{song-etal-2025-many,
    title = "Can Many-Shot {In-Context Learning} Help {LLM}s as Evaluators? {A} Preliminary Empirical Study",
    author = "Song, Mingyang  and
      Zheng, Mao  and
      Luo, Xuan",
    editor = "Rambow, Owen  and
      Wanner, Leo  and
      Apidianaki, Marianna  and
      Al-Khalifa, Hend  and
      Eugenio, Barbara Di  and
      Schockaert, Steven",
    booktitle = "Proceedings of the 31st International Conference on Computational Linguistics",
    month = jan,
    year = "2025",
    address = "Abu Dhabi, UAE",
    publisher = "Association for Computational Linguistics",
    url = "https://aclanthology.org/2025.coling-main.548/",
    pages = "8232--8241"
}

@inproceedings{yin-etal-2022-geomlama,
    title = "{G}eo{MLAMA}: Geo-Diverse Commonsense Probing on Multilingual Pre-Trained Language Models",
    author = "Yin, Da  and
      Bansal, Hritik  and
      Monajatipoor, Masoud  and
      Li, Liunian Harold  and
      Chang, Kai-Wei",
    editor = "Goldberg, Yoav  and
      Kozareva, Zornitsa  and
      Zhang, Yue",
    booktitle = "Proceedings of the 2022 Conference on Empirical Methods in Natural Language Processing",
    month = dec,
    year = "2022",
    address = "Abu Dhabi, United Arab Emirates",
    publisher = "Association for Computational Linguistics",
    url = "https://aclanthology.org/2022.emnlp-main.132/",
    doi = "10.18653/v1/2022.emnlp-main.132",
    pages = "2039--2055"
}

@article {PMID:843571,
	Title = {The measurement of observer agreement for categorical data},
	Author = {Landis, JR and Koch, GG},
	DOI = {10.2307/2529310},
	Number = {1},
	Volume = {33},
	Month = {March},
	Year = {1977},
	Journal = {Biometrics},
	ISSN = {0006-341X},
	Pages = {159—174},
	URL = {https://repositorio.unal.edu.co/bitstream/unal/80119/2/1010175682.2021.pdf},
}

@misc{yang2025qwen3technicalreport,
      title={Qwen3 Technical Report}, 
      author={An Yang and Anfeng Li and Baosong Yang and Beichen Zhang and Binyuan Hui and Bo Zheng and Bowen Yu and Chang Gao and Chengen Huang and Chenxu Lv and Chujie Zheng and Dayiheng Liu and Fan Zhou and Fei Huang and Feng Hu and Hao Ge and Haoran Wei and Huan Lin and Jialong Tang and Jian Yang and Jianhong Tu and Jianwei Zhang and Jianxin Yang and Jiaxi Yang and Jing Zhou and Jingren Zhou and Junyang Lin and Kai Dang and Keqin Bao and Kexin Yang and Le Yu and Lianghao Deng and Mei Li and Mingfeng Xue and Mingze Li and Pei Zhang and Peng Wang and Qin Zhu and Rui Men and Ruize Gao and Shixuan Liu and Shuang Luo and Tianhao Li and Tianyi Tang and Wenbiao Yin and Xingzhang Ren and Xinyu Wang and Xinyu Zhang and Xuancheng Ren and Yang Fan and Yang Su and Yichang Zhang and Yinger Zhang and Yu Wan and Yuqiong Liu and Zekun Wang and Zeyu Cui and Zhenru Zhang and Zhipeng Zhou and Zihan Qiu},
      year={2025},
      eprint={2505.09388},
      archivePrefix={arXiv},
      primaryClass={cs.CL},
      url={https://arxiv.org/abs/2505.09388}, 
}

@misc{deepseekai2026deepseekv4highlyefficientmilliontoken,
      title={{DeepSeek-V4}: Towards Highly Efficient Million-Token Context Intelligence}, 
      author={DeepSeek-AI and Anyi Xu and Bangcai Lin and Bing Xue and Bingxuan Wang and Bingzheng Xu and Bochao Wu and Bowei Zhang and Chaofan Lin and Chen Dong and Chenchen Ling and Chengda Lu and Chenggang Zhao and Chengqi Deng and Chengyu Hou and Chenhao Xu and Chenze Shao and Chong Ruan and Conner Sun and Damai Dai and Daya Guo and Dejian Yang and Deli Chen and Donghao Li and Dongjie Ji and Erhang Li and Fang Wei and Fangyun Lin and Fangzhou Yuan and Feiyu Xia and Fucong Dai and Guangbo Hao and Guanting Chen and Guoai Cao and Guolai Meng and Guowei Li and Han Yu and Han Zhang and Hanwei Xu and Hao Li and Haofen Liang and Haoling Zhang and Haoming Luo and Haoran Wei and Haotian Yuan and Haowei Zhang and Haowen Luo and Haoyu Chen and Haozhe Ji and Hengqing Zhang and Honghui Ding and Hongxuan Tang and Huanqi Cao and Huazuo Gao and Hui Qu and Hui Zeng and J Yang and JQ Zhu and Jia Luo and Jia Song and Jia Yu and Jialiang Huang and Jialu Cai and Jian Liang and Jiangting Zhou and Jiasheng Ye and Jiashi Li and Jiaxin Xu and Jiewen Hu and Jieyu Yang and Jin Chen and Jin Yan and Jingchang Chen and Jingli Zhou and Jingting Xiang and Jingyang Yuan and Jingyuan Cheng and Jingzi Zhou and Jinhua Zhu and Jiping Yu and Joseph Sun and Jun Ran and Junguang Jiang and Junjie Qiu and Junlong Li and Junmin Zheng and Junxiao Song and Kai Dong and Kaige Gao and Kang Guan and Kexing Zhou and Kezhao Huang and Kuai Yu and Lean Wang and Lecong Zhang and Lei Wang and Leyi Xia and Li Zhang and Liang Zhao and Lihua Guo and Lingxiao Luo and Linwang Ma and Linyan Zhu and Litong Wang and Liyu Cai and Liyue Zhang and Longhao Chen and MS Di and MY Xu and Max Mei and Miaojun Wang and Mingchuan Zhang and Minghua Zhang and Minghui Tang and Mingming Li and Mingxu Zhou and Minmin Han and Ning Wang and Panpan Huang and Panpan Wang and Peixin Cong and Peiyi Wang and Peng Zhang and Qiancheng Wang and Qihao Zhu and Qingyang Li and Qinyu Chen and Qiushi Du and Qiwei Jiang and Rui Tian and Ruifan Xu and Ruijie Lu and Ruiling Xu and Ruiqi Ge and Ruisong Zhang and Ruizhe Pan and Runji Wang and Runqian Chen and Runqiu Yin and Runxin Xu and Ruomeng Shen and Ruoyu Zhang and Ruyi Chen and SH Liu and Shanghao Lu and Shangmian Sun and Shangyan Zhou and Shanhuang Chen and Shaofei Cai and Shaoheng Nie and Shaoqing Wu and Shaoyuan Chen and Shengding Hu and Shengyu Liu and Shiqiang Hu and Shirong Ma and Shiyu Wang and Shuiping Yu and Shunfeng Zhou and Shuting Pan and Shuying Yu and Songyang Zhou and Tao Ni and Tao Yun and Tian Jin and Tian Pei and Tian Ye and Tianle Lin and Tianran Ji and Tianyi Cui and Tianyuan Yue and Tingting Yu and Tun Wang and W Zhang and WL Xiao and Wangding Zeng and Wei An and Weilin Zhao and Wen Liu and Wenfeng Liang and Wenjie Pang and Wenjing Luo and Wenjing Yao and Wenjun Gao and Wenkai Yang and Wenlve Huang and Wenqing Hou and Wentao Zhang and Wenting Ma and Xi Gao and Xiang He and Xiangwen Wang and Xianzu Wang and Xiao Bi and Xiaodong Liu and Xiaohan Wang and Xiaokang Chen and Xiaokang Zhang and Xiaotao Nie and Xiaowen Sun and Xiaoxiang Wang and Xin Cheng and Xin Liu and Xin Xie and Xingchao Liu and Xingchen Liu and Xingkai Yu and Xingyou Li and Xinyu Yang and Xinyu Zhang and Xu Chen and Xuanyu Wang and Xuecheng Su and Xueyin Chen and Xuheng Lin and Xuwei Fu and YC Yan and YQ Wang and YW Ma and Yanfeng Luo and Yang Zhang and Yanhong Xu and Yanru Ma and Yanwen Huang and Yao Li and Yao Li and Yao Xu and Yao Zhao and Yaofeng Sun and Yaohui Wang and Yi Qian and Yi Shao and Yi Yu and Yichao Zhang and Yifan Ding and Yifan Shi and Yijia Wu and Yiliang Xiong and Yiling Ma and Ying He and Ying Tang and Ying Zhou and Yingjia Luo and Yinmin Zhong and Yishi Piao and Yisong Wang and Yixiang Zhang and Yixiao Chen and Yixuan Tan and Yixuan Wei and Yiyang Ma and Yiyuan Liu and Yonglun Yang and Yongqiang Guo and Yongtong Wu and Yu Wu and YuKun Li and Yuan Cheng and Yuan Ou and Yuanfan Xu and Yuanhao Li and Yuduan Wang and Yuehan Yang and Yuer Xu and Yuhan Wu and Yuhao Meng and Yuheng Zou and Yukun Zha and Yunfan Xiong and Yupeng Chen and Yuping Lin and Yuqian Cao and Yuqian Wang and Yushun Zhang and Yuting Yan and Yutong Lin and Yuxian Gu and Yuxiang Luo and Yuxiang You and Yuxuan Liu and Yuxuan Zhou and Yuyang Zhou and Yuzhen Huang and ZF Wu and Zehao Wang and Zehua Zhao and Zehui Ren and Zekai Zhang and Zhangli Sha and Zhe Fu and Zhe Ju and Zhean Xu and Zhenda Xie and Zhengyan Zhang and Zheren Gao and Zhewen Hao and Zhibin Gou and Zhicheng Ma and Zhigang Yan and Zhihong Shao and Zhixian Huang and Zhixuan Chen and Zhiyu Wu and Zhizhou Ren and Zhongyu Wu and Zhuoshu Li and Zhuping Zhang and Zian Xu and Zihao Wang and Zihua Qu and Zihui Gu and Zijia Zhu and Zilin Li and Zipeng Zhang and Ziwei Xie and Ziyi Gao and Ziyi Wan and Zizheng Pan and Zongqing Yao},
      year={2026},
      eprint={2606.19348},
      archivePrefix={arXiv},
      primaryClass={cs.CL},
      url={https://arxiv.org/abs/2606.19348}, 
}

@misc{gpt5,
      title={{OpenAI} {GPT-5} System Card}, 
      author={Aaditya Singh and Adam Fry and Adam Perelman and Adam Tart and Adi Ganesh and Ahmed El-Kishky and Aidan McLaughlin and Aiden Low and AJ Ostrow and Akhila Ananthram and Akshay Nathan and Alan Luo and Alec Helyar and Aleksander Madry and Aleksandr Efremov and Aleksandra Spyra and Alex Baker-Whitcomb and Alex Beutel and Alex Karpenko and Alex Makelov and Alex Neitz and Alex Wei and Alexandra Barr and Alexandre Kirchmeyer and Alexey Ivanov and Alexi Christakis and Alistair Gillespie and Allison Tam and Ally Bennett and Alvin Wan and Alyssa Huang and Amy McDonald Sandjideh and Amy Yang and Ananya Kumar and Andre Saraiva and Andrea Vallone and Andrei Gheorghe and Andres Garcia Garcia and Andrew Braunstein and Andrew Liu and Andrew Schmidt and Andrey Mereskin and Andrey Mishchenko and Andy Applebaum and Andy Rogerson and Ann Rajan and Annie Wei and Anoop Kotha and Anubha Srivastava and Anushree Agrawal and Arun Vijayvergiya and Ashley Tyra and Ashvin Nair and Avi Nayak and Ben Eggers and Bessie Ji and Beth Hoover and Bill Chen and Blair Chen and Boaz Barak and Borys Minaiev and Botao Hao and Bowen Baker and Brad Lightcap and Brandon McKinzie and Brandon Wang and Brendan Quinn and Brian Fioca and Brian Hsu and Brian Yang and Brian Yu and Brian Zhang and Brittany Brenner and Callie Riggins Zetino and Cameron Raymond and Camillo Lugaresi and Carolina Paz and Cary Hudson and Cedric Whitney and Chak Li and Charles Chen and Charlotte Cole and Chelsea Voss and Chen Ding and Chen Shen and Chengdu Huang and Chris Colby and Chris Hallacy and Chris Koch and Chris Lu and Christina Kaplan and Christina Kim and CJ Minott-Henriques and Cliff Frey and Cody Yu and Coley Czarnecki and Colin Reid and Colin Wei and Cory Decareaux and Cristina Scheau and Cyril Zhang and Cyrus Forbes and Da Tang and Dakota Goldberg and Dan Roberts and Dana Palmie and Daniel Kappler and Daniel Levine and Daniel Wright and Dave Leo and David Lin and David Robinson and Declan Grabb and Derek Chen and Derek Lim and Derek Salama and Dibya Bhattacharjee and Dimitris Tsipras and Dinghua Li and Dingli Yu and DJ Strouse and Drew Williams and Dylan Hunn and Ed Bayes and Edwin Arbus and Ekin Akyurek and Elaine Ya Le and Elana Widmann and Eli Yani and Elizabeth Proehl and Enis Sert and Enoch Cheung and Eri Schwartz and Eric Han and Eric Jiang and Eric Mitchell and Eric Sigler and Eric Wallace and Erik Ritter and Erin Kavanaugh and Evan Mays and Evgenii Nikishin and Fangyuan Li and Felipe Petroski Such and Filipe de Avila Belbute Peres and Filippo Raso and Florent Bekerman and Foivos Tsimpourlas and Fotis Chantzis and Francis Song and Francis Zhang and Gaby Raila and Garrett McGrath and Gary Briggs and Gary Yang and Giambattista Parascandolo and Gildas Chabot and Grace Kim and Grace Zhao and Gregory Valiant and Guillaume Leclerc and Hadi Salman and Hanson Wang and Hao Sheng and Haoming Jiang and Haoyu Wang and Haozhun Jin and Harshit Sikchi and Heather Schmidt and Henry Aspegren and Honglin Chen and Huida Qiu and Hunter Lightman and Ian Covert and Ian Kivlichan and Ian Silber and Ian Sohl and Ibrahim Hammoud and Ignasi Clavera and Ikai Lan and Ilge Akkaya and Ilya Kostrikov and Irina Kofman and Isak Etinger and Ishaan Singal and Jackie Hehir and Jacob Huh and Jacqueline Pan and Jake Wilczynski and Jakub Pachocki and James Lee and James Quinn and Jamie Kiros and Janvi Kalra and Jasmyn Samaroo and Jason Wang and Jason Wolfe and Jay Chen and Jay Wang and Jean Harb and Jeffrey Han and Jeffrey Wang and Jennifer Zhao and Jeremy Chen and Jerene Yang and Jerry Tworek and Jesse Chand and Jessica Landon and Jessica Liang and Ji Lin and Jiancheng Liu and Jianfeng Wang and Jie Tang and Jihan Yin and Joanne Jang and Joel Morris and Joey Flynn and Johannes Ferstad and Johannes Heidecke and John Fishbein and John Hallman and Jonah Grant and Jonathan Chien and Jonathan Gordon and Jongsoo Park and Jordan Liss and Jos Kraaijeveld and Joseph Guay and Joseph Mo and Josh Lawson and Josh McGrath and Joshua Vendrow and Joy Jiao and Julian Lee and Julie Steele and Julie Wang and Junhua Mao and Kai Chen and Kai Hayashi and Kai Xiao and Kamyar Salahi and Kan Wu and Karan Sekhri and Karan Sharma and Karan Singhal and Karen Li and Kenny Nguyen and Keren Gu-Lemberg and Kevin King and Kevin Liu and Kevin Stone and Kevin Yu and Kristen Ying and Kristian Georgiev and Kristie Lim and Kushal Tirumala and Kyle Miller and Lama Ahmad and Larry Lv and Laura Clare and Laurance Fauconnet and Lauren Itow and Lauren Yang and Laurentia Romaniuk and Leah Anise and Lee Byron and Leher Pathak and Leon Maksin and Leyan Lo and Leyton Ho and Li Jing and Liang Wu and Liang Xiong and Lien Mamitsuka and Lin Yang and Lindsay McCallum and Lindsey Held and Liz Bourgeois and Logan Engstrom and Lorenz Kuhn and Louis Feuvrier and Lu Zhang and Lucas Switzer and Lukas Kondraciuk and Lukasz Kaiser and Manas Joglekar and Mandeep Singh and Mandip Shah and Manuka Stratta and Marcus Williams and Mark Chen and Mark Sun and Marselus Cayton and Martin Li and Marvin Zhang and Marwan Aljubeh and Matt Nichols and Matthew Haines and Max Schwarzer and Mayank Gupta and Meghan Shah and Melody Y. Guan and Melody Huang and Meng Dong and Mengqing Wang and Mia Glaese and Micah Carroll and Michael Lampe and Michael Malek and Michael Sharman and Michael Zhang and Michele Wang and Michelle Pokrass and Mihai Florian and Mikhail Pavlov and Miles Wang and Ming Chen and Mingxuan Wang and Minnia Feng and Mo Bavarian and Molly Lin and Moose Abdool and Mostafa Rohaninejad and Nacho Soto and Natalie Staudacher and Natan LaFontaine and Nathan Marwell and Nelson Liu and Nick Preston and Nick Turley and Nicklas Ansman and Nicole Blades and Nikil Pancha and Nikita Mikhaylin and Niko Felix and Nikunj Handa and Nishant Rai and Nitish Keskar and Noam Brown and Ofir Nachum and Oleg Boiko and Oleg Murk and Olivia Watkins and Oona Gleeson and Pamela Mishkin and Patryk Lesiewicz and Paul Baltescu and Pavel Belov and Peter Zhokhov and Philip Pronin and Phillip Guo and Phoebe Thacker and Qi Liu and Qiming Yuan and Qinghua Liu and Rachel Dias and Rachel Puckett and Rahul Arora and Ravi Teja Mullapudi and Raz Gaon and Reah Miyara and Rennie Song and Rishabh Aggarwal and RJ Marsan and Robel Yemiru and Robert Xiong and Rohan Kshirsagar and Rohan Nuttall and Roman Tsiupa and Ronen Eldan and Rose Wang and Roshan James and Roy Ziv and Rui Shu and Ruslan Nigmatullin and Saachi Jain and Saam Talaie and Sam Altman and Sam Arnesen and Sam Toizer and Sam Toyer and Samuel Miserendino and Sandhini Agarwal and Sarah Yoo and Savannah Heon and Scott Ethersmith and Sean Grove and Sean Taylor and Sebastien Bubeck and Sever Banesiu and Shaokyi Amdo and Shengjia Zhao and Sherwin Wu and Shibani Santurkar and Shiyu Zhao and Shraman Ray Chaudhuri and Shreyas Krishnaswamy and Shuaiqi and Xia and Shuyang Cheng and Shyamal Anadkat and Simón Posada Fishman and Simon Tobin and Siyuan Fu and Somay Jain and Song Mei and Sonya Egoian and Spencer Kim and Spug Golden and SQ Mah and Steph Lin and Stephen Imm and Steve Sharpe and Steve Yadlowsky and Sulman Choudhry and Sungwon Eum and Suvansh Sanjeev and Tabarak Khan and Tal Stramer and Tao Wang and Tao Xin and Tarun Gogineni and Taya Christianson and Ted Sanders and Tejal Patwardhan and Thomas Degry and Thomas Shadwell and Tianfu Fu and Tianshi Gao and Timur Garipov and Tina Sriskandarajah and Toki Sherbakov and Tomek Korbak and Tomer Kaftan and Tomo Hiratsuka and Tongzhou Wang and Tony Song and Tony Zhao and Troy Peterson and Val Kharitonov and Victoria Chernova and Vineet Kosaraju and Vishal Kuo and Vitchyr Pong and Vivek Verma and Vlad Petrov and Wanning Jiang and Weixing Zhang and Wenda Zhou and Wenlei Xie and Wenting Zhan and Wes McCabe and Will DePue and Will Ellsworth and Wulfie Bain and Wyatt Thompson and Xiangning Chen and Xiangyu Qi and Xin Xiang and Xinwei Shi and Yann Dubois and Yaodong Yu and Yara Khakbaz and Yifan Wu and Yilei Qian and Yin Tat Lee and Yinbo Chen and Yizhen Zhang and Yizhong Xiong and Yonglong Tian and Young Cha and Yu Bai and Yu Yang and Yuan Yuan and Yuanzhi Li and Yufeng Zhang and Yuguang Yang and Yujia Jin and Yun Jiang and Yunyun Wang and Yushi Wang and Yutian Liu and Zach Stubenvoll and Zehao Dou and Zheng Wu and Zhigang Wang},
      year={2026},
      eprint={2601.03267},
      archivePrefix={arXiv},
      primaryClass={cs.CL},
      url={https://arxiv.org/abs/2601.03267}, 
}

@inproceedings{10.1145/3600006.3613165,
author = {Kwon, Woosuk and Li, Zhuohan and Zhuang, Siyuan and Sheng, Ying and Zheng, Lianmin and Yu, Cody Hao and Gonzalez, Joseph and Zhang, Hao and Stoica, Ion},
title = {Efficient Memory Management for Large Language Model Serving with {PagedAttention}},
year = {2023},
isbn = {9798400702297},
publisher = {Association for Computing Machinery},
address = {New York, NY, USA},
url = {https://doi.org/10.1145/3600006.3613165},
doi = {10.1145/3600006.3613165},
booktitle = {Proceedings of the 29th Symposium on Operating Systems Principles},
pages = {611–626},
numpages = {16},
address = {Koblenz, Germany},
series = {SOSP '23}
}

@misc{eval-harness,
  author       = {Gao, Leo and Tow, Jonathan and Abbasi, Baber and Biderman, Stella and Black, Sid and DiPofi, Anthony and Foster, Charles and Golding, Laurence and Hsu, Jeffrey and Le Noac'h, Alain and Li, Haonan and McDonell, Kyle and Muennighoff, Niklas and Ociepa, Chris and Phang, Jason and Reynolds, Laria and Schoelkopf, Hailey and Skowron, Aviya and Sutawika, Lintang and Tang, Eric and Thite, Anish and Wang, Ben and Wang, Kevin and Zou, Andy},
  title        = {The Language Model Evaluation Harness},
  month        = 07,
  year         = 2024,
  publisher    = {Zenodo},
  version      = {v0.4.3},
  doi          = {10.5281/zenodo.12608602},
  url          = {https://zenodo.org/records/12608602}
}

@article{hu2021lora,
  title={{LoRA}: Low-rank adaptation of large language models},
  author={Hu, Edward J and Shen, Yelong and Wallis, Phillip and Allen-Zhu, Zeyuan and Li, Yuanzhi and Wang, Shean and Wang, Lu and Chen, Weizhu},
  journal={arXiv preprint arXiv:2106.09685},
  year={2021}
}

@misc{loshchilov2019decoupledweightdecayregularization,
      title={Decoupled Weight Decay Regularization}, 
      author={Ilya Loshchilov and Frank Hutter},
      year={2019},
      eprint={1711.05101},
      archivePrefix={arXiv},
      primaryClass={cs.LG},
      url={https://arxiv.org/abs/1711.05101}, 
}

@inproceedings{wiyono-etal-2025-indopref,
    title = "{I}ndo{P}ref: A Multi-Domain Pairwise Preference Dataset for {I}ndonesian",
    author = "Wiyono, Vanessa Rebecca  and
      Anugraha, David  and
      Purwarianti, Ayu  and
      Winata, Genta Indra",
    editor = "Inui, Kentaro  and
      Sakti, Sakriani  and
      Wang, Haofen  and
      Wong, Derek F.  and
      Bhattacharyya, Pushpak  and
      Banerjee, Biplab  and
      Ekbal, Asif  and
      Chakraborty, Tanmoy  and
      Singh, Dhirendra Pratap",
    booktitle = "Proceedings of the 14th International Joint Conference on Natural Language Processing and the 4th Conference of the Asia-Pacific Chapter of the Association for Computational Linguistics",
    month = dec,
    year = "2025",
    address = "Mumbai, India",
    publisher = "The Asian Federation of Natural Language Processing and The Association for Computational Linguistics",
    url = "https://aclanthology.org/2025.ijcnlp-short.12/",
    doi = "10.18653/v1/2025.ijcnlp-short.12",
    pages = "128--138",
    ISBN = "979-8-89176-299-2"
}

@inproceedings{Prism2024,
author = {Kirk, Hannah Rose and Whitefield, Alexander and R{\"o}ttger, Paul and Bean, Andrew and Margatina, Katerina and Ciro, Juan and Mosquera, Rafael and Bartolo, Max and Williams, Adina and He, He and Vidgen, Bertie and Hale, Scott A.},
title = {The {PRISM} alignment dataset: {W}hat participatory, representative and individualised human feedback reveals about the subjective and multicultural alignment of large language models},
year = {2024},
isbn = {9798331314385},
publisher = {Curran Associates Inc.},
address = {Red Hook, NY, USA},
booktitle = {Proceedings of the 38th International Conference on Neural Information Processing Systems},
articleno = {3342},
numpages = {109},
address = {Vancouver, BC, Canada},
series = {NIPS '24}
}

@article{anthropic2025system,
  title={System card: {Claude} {Opus} 4 \& {Claude} {Sonnet} 4},
  author={Anthropic},
  journal={Claude-4 Model Card},
  year={2025}
}

\appendix
\section{Data Generation}
\label{apdx:data_generation}

\subsection{Semantic Filtering}
\label{apdx:semantic_filtering}
\paragraph{Automated Filtering Steps in Preference Generation}

We apply three sequential filtering stages to the generated preference pairs. First, Language Validation ensures all entries are strictly in English. Second, Near-Duplicate Removal utilizes the \texttt{MinHash} algorithm~\cite{minhash} to eliminate structural redundancies. Finally, we perform Semantic Deduplication by encoding behaviors with \textit{GemmaEmbedding}~\cite{embedding_gemma_2025} and calculating cosine similarity. After evaluating thresholds in the range $\{0.70, \dots, 0.95\}$ via manual inspection, we select 0.90 for the preference dataset to balance diversity and preservation, while a stricter threshold of 0.80 is applied to the MCQ data to ensure options remain clearly distinguishable. This pipeline yields a high-quality corpus of 15,530 training and 3,894 raw test samples in English.

\paragraph{Semantic Similarity Filtering in MCQ Generation:} To ensure that the four choices are not semantically similar, we use GemmaEmbedding and set a similarity threshold of 0.80. If the similarity between any two choices exceeds the threshold, we replace one of the distractor countries and select a new choice. This process continues until no pair of choices exceeds the threshold. The test splits of preference data are deliberately constructed to mirror MCQ's (country, category, aspect) combinations while the training preference data has much broader, non-overlapping coverage.

\subsection{Used Web-Resources}
\label{appx:urls_desc}
We compile cultural etiquette guidance from publicly available online sources covering 13 countries (Saudi Arabia, United Arab Emirates, Yemen, Jordan, Lebanon, Syria, Palestine, Egypt, Sudan, Morocco, Algeria, Tunisia, and Libya), plus a small set of region-wide (Arab World) references. The sources include travel blogs and culture/etiquette guides, complemented by official travel recommendations and advisories from government and embassy-affiliated services (e.g., the U.S. Department of State, the UK FCDO, the Government of Canada). URLs are provided in Table~\ref{tab:urls_desc}.

\begin{table*}[!h]
\centering
\begin{tabular}{p{4cm} p{9cm}}
\hline
\textbf{Country} & \textbf{Sources} \\
\hline

Saudi Arabia &
\href{https://culturalatlas.sbs.com.au/saudi-arabian-culture/saudi-arabian-culture-do-s-and-don-ts}{S1},
\href{https://www.traveloka.com/en-en/explore/destination/pl-dos-donts-saudi-arabia/271991}{S2},
\href{https://www.wanderlustmagazine.com/inspiration/dos-and-donts-for-travel-in-saudi/}{S3},
\href{https://www.smartraveller.gov.au/destinations/middle-east/saudi-arabia}{S4},
\href{https://travel.gc.ca/destinations/saudi-arabia}{S5} \\

United Arab Emirates &
\href{https://www.holidaykeepers.com/blogs/etiquette-in-dubai/}{UAE1},
\href{https://www.roughguides.com/united-arab-emirates/customs-etiquette/}{UAE2},
\href{https://www.theinvisibletourist.com/visiting-abu-dhabi-for-tourists/}{UAE3},
\href{https://northpennnow.com/news/2025/apr/10/cultural-etiquette-in-the-uae-dos-and-donts-for-tourists/}{UAE4},
\href{https://www.omartourism.com/cultural-etiquette-in-the-uae-what-every-tourist-should-know/}{UAE5},
\href{https://www.uae-embassy.org/discover-uae/traveling-in-the-uae/uae-culture}{UAE6},
\href{https://asianahospitality.com/dubai-travel-etiquette-guide/}{UAE7} \\

Yemen &
\href{https://travel.state.gov/content/travel/en/traveladvisories/traveladvisories/yemen-travel-advisory.html}{Y1},
\href{https://www.gaylecotton.com/cultural-clues-communication-guidelines-for-yemen/}{Y2},
\href{https://guide.culturecrossing.net/basics_business_student.php?id=228}{Y3},
\href{https://www.etiquettescholar.com/dining_etiquette/table-etiquette/mideast_dinner_etiquette/yemen.html}{Y4},
\href{https://www.travpa.com/asia/yemen/local-customs-and-etiquette-in-yemen/}{Y5},
\href{https://cso-yemen.org/yemen-dress-code/}{Y6} \\

Jordan &
\href{https://traveler.marriott.com/jordan/jordan-travel-tips-etiquette/}{J1},
\href{https://www.zamantours.com/blog/cultural-etiquette-in-jordan-dos-and-donts-for-travelers/}{J2},
\href{https://flowtraveljo.com/list-of-dos-and-donts-in-jordan/}{J3},
\href{https://www.engagingcultures.com/21-cultural-tips-for-traveling-in-jordan/}{J4},
\href{https://alittleadrift.com/jordan-culture-norms/}{J5},
\href{https://www.roughguides.com/jordan/customs-etiquette/}{J6} \\

Lebanon &
\href{https://culturalatlas.sbs.com.au/lebanese-culture/lebanese-culture-etiquette}{L1},
\href{https://culturalatlas.sbs.com.au/lebanese-culture/lebanese-culture-do-s-and-don-ts}{L2},
\href{https://rachelsruminations.com/lebanon-travel-tips/}{L3},
\href{https://racheloffduty.com/blog/lebanon-travel-tips/}{L4},
\href{https://www.agatetravel.com/lebanon/tips.html}{L5},
\href{https://www.malikahkelly.com/blog/2023/6/29/how-to-explore-lebanon-safety-tips-you-need-before-you-plan-your-trip}{L6},
\href{https://www.reddit.com/r/lebanon/comments/5u0dxc/what_are_some_dos_and_donts_in_lebanon/}{L7} \\

Syria &
\href{https://culturalatlas.sbs.com.au/syrian-culture/syrian-culture-do-s-and-don-ts}{SY1},
\href{https://culturalatlas.sbs.com.au/syrian-culture/syrian-culture-etiquette}{SY2},
\href{https://guide.culturecrossing.net/basics_business_student.php?id=199}{SY3},
\href{https://www.ccinform.co.uk/practice-guidance/syrian-culture-and-customs-quick-guide/}{SY4},
\href{https://www.anothertravel.com/middle-east/syria/culture/}{SY5},
\href{https://www.thejjadventures.com/2024/11/syria-guide/}{SY6},
\href{https://www.reddit.com/r/Syria/comments/10gqxdk/visiting_a_syrian_family_what_are_the_usual/}{SY7} \\

Palestine &
\href{https://culturalatlas.sbs.com.au/palestinian-culture/palestinian-culture-do-s-and-don-ts}{P1},
\href{https://ecpalestine.org/orientation-guide-for-visitors-to-palestine/}{P2},
\href{https://excellencenter.org/etiquette-tips-for-your-stay-in-palestine/}{P3},
\href{https://www.walkpalestine.com/en/information}{P4},
\href{https://gopalestine.org/things-you-should-know-before-visiting-palestine/}{P5},
\href{https://ecpalestine.org/8-tips-for-travelling-to-palestine-in-2025/}{P6},
\href{https://www.reddit.com/r/Palestine/comments/n40a3w/what_are_the_dos_and_donts_when_visiting_palestine/}{P7},
\href{https://expatpanda.com/planning-a-trip-to-palestine-a-detailed-guide/}{P8} \\

Egypt &
\href{https://www.traveltailors.com/en/client/site/m37260-egypt-do-s-and-don-ts.htm}{E1},
\href{https://travel.state.gov/content/travel/en/international-travel/International-Travel-Country-Information-Pages/Egypt.html}{E2},
\href{https://www.viator.com/blog/Egypt-Dos-and-Donts-Unwritten-Rules-That-Every-Egyptian-Knows/l98078}{E3},
\href{https://www.smartraveller.gov.au/destinations/africa/egypt}{E4},
\href{https://www.tripsinegypt.com/blog/egypt-travel-guide/do-s-and-do-not-s-in-egypt/}{E5},
\href{https://travel.gc.ca/destinations/egypt}{E6},
\href{https://www.roughguides.com/egypt/customs-etiquette/}{E7},
\href{https://www.gov.uk/foreign-travel-advice/egypt/safety-and-security}{E8} \\

Sudan &
\href{https://www.smartraveller.gov.au/destinations/africa/sudan}{SU1},
\href{https://tours.solofemaletravelers.club/sudan-travel-tips/}{SU2},
\href{https://culturalatlas.sbs.com.au/north-sudanese-culture/north-sudanese-culture-do-s-and-don-ts}{SU3},
\href{https://againstthecompass.com/en/sudan-travel-guide-backpacking/}{SU4},
\href{https://travel.gc.ca/destinations/sudan}{SU5},
\href{https://againstthecompass.com/en/visit-sudan-female-traveler/}{SU6} \\

Morocco &
\href{https://www.morocco-adventure-holidays.com/morocco/cultural-etiquette/}{M1},
\href{https://isleblue.co/categories/morocco-guide/articles/dos-and-dont-in-morocco}{M2},
\href{https://www.jetsetandforget.com/moroccotips/}{M3},
\href{https://blog.tatonka.com/en/dos-and-donts-morocco/}{M4},
\href{https://www.lonelyplanet.com/articles/things-to-know-before-traveling-to-morocco}{M5},
\href{https://pinatravels.org/what-not-to-do-in-morocco/}{M6},
\href{https://www.skyhookadventure.com/blog/morocco-travel-tips}{M7},
\href{https://janicerohrssen.com/blog/18-things-you-must-know-before-visiting-morocco-travel-tips}{M8},
\href{https://www.trafalgar.com/real-word/what-know-moroccan-culture-etiquette/}{M9},
\href{https://touristjourney.com/culture-guide-to-morocco/}{M10},
\href{https://www.journeybeyondtravel.com/morocco/culture-etiquette}{M11},
\href{https://www.rivieratravel.co.uk/blog/the-dos-and-donts-moroccan-culture}{M12} \\

Algeria &
\href{https://www.mosaicnorthafrica.com/cultural-etiquette-while-traveling-in-algeria/}{A1},
\href{https://guide.culturecrossing.net/basics_business_student.php?id=3}{A2},
\href{https://timetravelbee.com/en/places/traveling-to-algeria-survival-guide/}{A3},
\href{https://toltips.com/dos-and-donts-in-algeria/}{A4},
\href{https://timetravelbee.com/en/places/safety-in-algeria/}{A5},
\href{https://travel.gc.ca/destinations/algeria}{A6},
\href{https://www.smartraveller.gov.au/destinations/africa/algeria}{A7},
\href{https://www.fancyalgeria.com/knowbeforeyougo}{A8} \\

Tunisia &
\href{https://www.traveletiquette.co.uk/EtiquetteTunisia.html}{T1},
\href{https://itinerarytrip.com/travel-guides/Tunisia/guide/en/Etiquette-Essentials-for-amazing-Tunisia%3A-Do%27s-and-Don%27ts}{T2},
\href{https://nickipoststravelstuff.com/know-before-visiting-tunisia/}{T3},
\href{https://www.reddit.com/r/Tunisia/comments/1bsyji5/tips_for_travelers/}{T4},
\href{https://www.sarahdegheselle.com/en/10-tunisia-travel-tips/}{T5},
\href{https://www.engagingcultures.com/what-to-wear-when-traveling-in-tunisia/}{T6},
\href{https://www.gov.uk/foreign-travel-advice/tunisia}{T7},
\href{https://travel.gc.ca/destinations/tunisia}{T8} \\

Libya &
\href{https://www.smartraveller.gov.au/destinations/africa/libya}{LI1},
\href{https://travel.state.gov/content/travel/en/traveladvisories/traveladvisories/libya-travel-advisory.html}{LI2},
\href{https://libyaadventures.com/some-useful-travel-tips-when-visiting-libya/}{LI3},
\href{https://www.commisceo-global.com/resources/country-guides/libya-guide}{LI4},
\href{https://www.wewillnomad.com/destination/libya}{LI5},
\href{https://againstthecompass.com/en/travel-libya/}{LI6},
\href{https://en.wikivoyage.org/wiki/Libya}{LI7},
\href{https://www.worldpackers.com/articles/travel-etiquette-north-africa}{LI8},
\href{https://globalgaz.com/visiting-libya-what-you-need-to-know/}{LI9} \\

Arab World (Generic) &
\href{https://www.simplexitytravel.com/inspirations/the-dos-and-donts-of-travelling-in-the-middle-east/}{AR1},
\href{https://24hourkandura.com/blog/understanding-arabic-customs-and-etiquette-a-guide-for-travelers/}{AR2},
\href{https://www.arabacademy.com/arabic-customs-traditions/}{AR3},
\href{https://www.hostelworld.com/blog/arab-culture/}{AR4},
\href{https://blog.windstarcruises.com/everything-you-need-to-know-about-customs-in-the-middle-east-travel/}{AR5} \\

\hline
\end{tabular}%
\caption{List of travel etiquette and travel-advisory sources used in our work.}
\label{tab:urls_desc}
\end{table*}

\subsection{Categories: Description and Statistics}
\label{appx:categories}

This appendix clarifies each culture-related category used in our annotations. Each category captures culturally grounded norms, expectations, or constraints that may vary across Arab countries.

\paragraph{Social Etiquette}
Every day interpersonal norms in public and social interactions, such as forms of greeting, appropriate topics of conversation, respectful address, queuing behavior, and when it is acceptable to refuse or accept an offer.

\paragraph{Hospitality \& Home Life}
Practices and expectations around visiting, hosting, and household life, including guest treatment, entering homes, seating arrangements, gift-giving, and norms around privacy and family space.

\paragraph{Religion \& Rituals}
Religious practices and observances that shape daily life, such as prayer times, fasting periods, religious holidays, visiting places of worship, and behaviors considered respectful or disrespectful in religious contexts.

\paragraph{Dining \& Food}
Customs and etiquette related to eating and shared meals (how to eat rather than what to eat), such as waiting for the host to start, eating with the right hand, using bread to scoop food, and expectations around sharing dishes.

\paragraph{Business \& Work}
Workplace and professional norms, including meeting etiquette, negotiation style, hierarchy and titles, punctuality expectations, communication formality, and norms around work hours or availability.

\paragraph{Modern \& Technology}
Cultural norms around photography, social media, privacy, and device use in public or formal settings.

\paragraph{Gender \& Travel}
Gender-specific norms and rules relevant to travelers, such as clothing expectations for women, gender segregation in certain venues, and guidance on mixed-gender interactions in public or private settings.

\paragraph{Family \& Children}
Norms centered on family structure, roles, and interactions with children, including respect for elders, family decision-making, acceptable behavior around children, and expectations for family-oriented settings.

\paragraph{Health \& Environment}
Cultural practices tied to health, hygiene, and environmental context, such as attitudes toward heat and hydration, norms around cleanliness, public health expectations, and locally relevant environmental considerations.

\paragraph{Politics \& Law}
Norms and constraints shaped by legal or political context, including sensitive topics to avoid, restrictions on speech or behavior, and locally specific rules that visitors should be aware of.

\paragraph{Shopping \& Money}
Norms related to commerce and payments, such as bargaining customs, tipping expectations, appropriate conduct in markets, and payment preferences (cash vs.\ card) where culturally relevant.

\paragraph{Charity \& Society}
Social responsibility norms, including charitable giving practices, community support expectations, volunteering, and culturally grounded concepts of generosity and mutual aid.

\paragraph{Personal Appearance}
Expectations around dress and grooming in different contexts (e.g., modesty norms, acceptable swimwear in certain settings, or formality levels), distinct from explicitly gender-specific travel rules.

\paragraph{Transport}
Norms and rules specific to getting around, such as women-only metro cars, seating expectations, etiquette in taxis or ride-hailing, and behavior on public transport.

\paragraph{Pets \& Environment}
Cultural attitudes and practical norms involving animals in daily life, such as expectations about dogs in public spaces, interactions with stray animals, and where pets are typically welcomed or not.

\paragraph{Safety}
Safety guidance, including norms that reduce risk (e.g., situational awareness expectations), locally sensitive areas or behaviors, and advice shaped by context rather than general travel tips.

\begin{figure}[!h]
    \centering
    \includegraphics[width=0.93\linewidth]{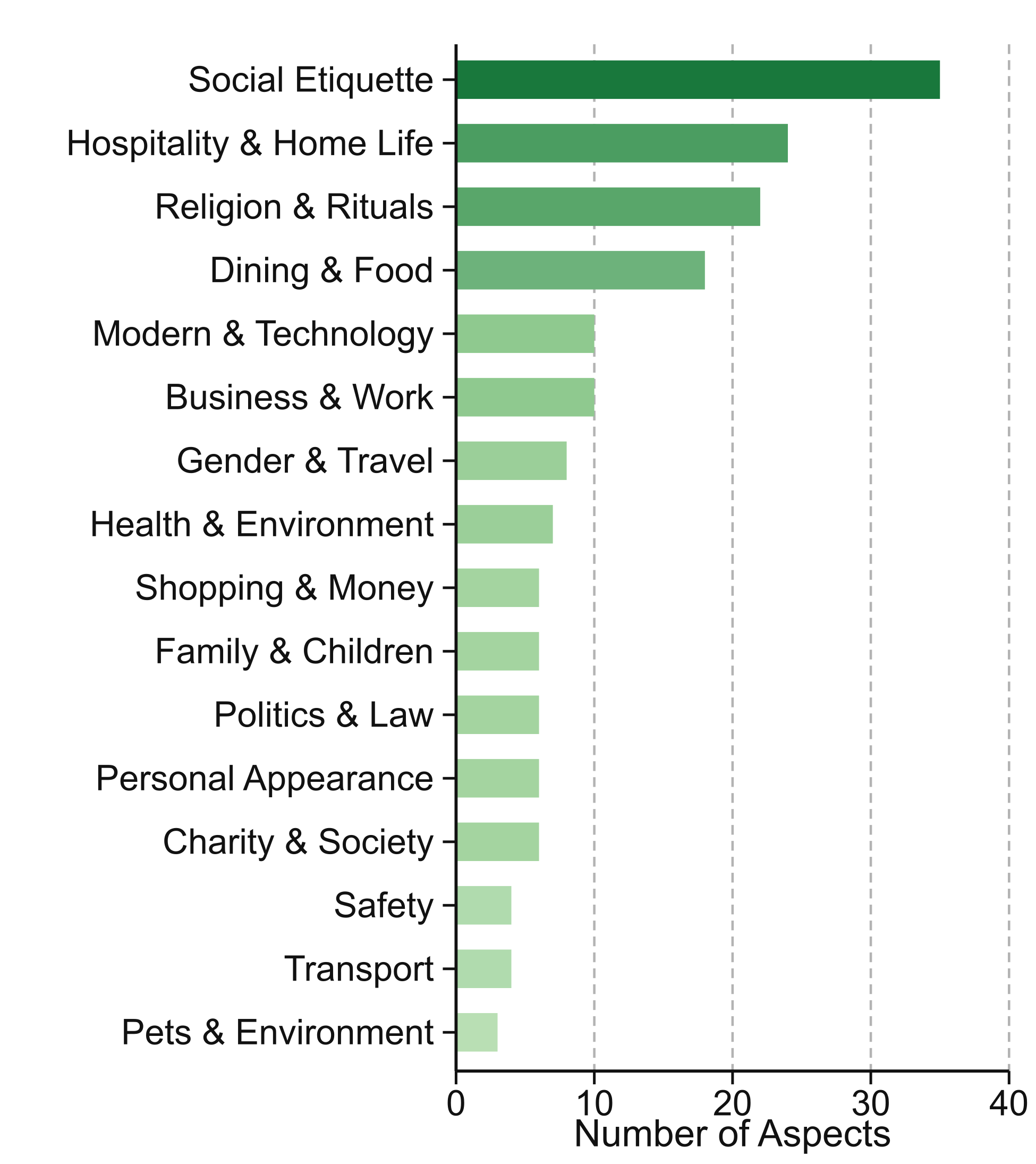 }
    \caption{Statistics of the number of aspects for each cultural category in \ds.}
    \label{fig:cultural_categories}
\end{figure}

Figure~\ref{fig:cultural_categories} provides summary statistics of the number of aspects for each category. The aspects for each category is listed in Table~\ref{tab:cultural_categories_aspects}. 

\begin{table}[!h]
\centering

\scalebox{0.8}{
\begin{tabular}{lrrr}
\toprule
\rowcolor{gray!30}
Country & Train & Test (Raw) & Test (Filtered) \\
\midrule
Egypt        & 1,219 & 329 & 329 \\
Morocco      & 1,114 & 237 & 226 \\
Saudi Arabia & 972  & 160 & 141 \\
Algeria      & 958  & 225 & 165 \\
Jordan       & 924  & 257 & 254 \\
Lebanon      & 907  & 207 & 199 \\
Tunisia      & 860  & 405 & 317 \\
Yemen        & 758  & 205 & 189 \\
Syria        & 757  & 172 & 128 \\
UAE          & 720  & 183 & 159 \\
Palestine    & 679  & 153 & 135 \\
Libya        & 628  & 119 & 114 \\
Iraq         & 598  & 144 & 135 \\
Oman         & 572  & 135 & 134 \\
Sudan        & 564  & 99  & 80  \\
Bahrain      & 545  & 148 & 147 \\
Qatar        & 517  & 109 & 104 \\
Somalia      & 488  & 144 & 32  \\
Kuwait       & 476  & 106 & 82  \\
Djibouti     & 436  & 137 & 51  \\
Comoros      & 421  & 109 & 82  \\
Mauritania   & 417  & 111 & 97  \\
\midrule
\textbf{Total} & \textbf{15,530} & \textbf{3,894} & \textbf{3,300} \\
\bottomrule
\end{tabular}
}
\caption{Statistics of the English preference dataset across 22 countries, detailing the sample counts for training, raw test, and human-filtered test splits.}
\label{tab:preference_dataset_stats_country}
\end{table}
\begin{table}[!h]
\centering

\scalebox{0.65}{
\begin{tabular}{lcccc}
\toprule
\rowcolor{gray!30}
Category & Train & Test (Raw) & Test (Filtered) & MCQ \\
\midrule
Social Etiquette & 3327 & 716 & 605 & 458  \\
Religion \& Rituals & 1647 & 440 & 336 & 172 \\
Dining \& Food & 1553 & 369 & 283 & 214 \\
Hospitality \& Home Life & 1497 & 386 & 325 & 246 \\
Gender \& Travel & 745 & 193 & 159 & 98 \\
Personal Appearance & 707 & 170 & 155 & 90 \\
Business \& Work & 691 & 200 & 145 & 98 \\
Shopping \& Money & 540 & 150 & 136 & 76 \\
Modern \& Technology & 553 & 135 & 119 & 94 \\
Politics \& Law & 360 & 104 & 100 & 56 \\
\bottomrule
\end{tabular}
}
\caption{Top 10 categories by sample count in the preference dataset across the training, raw test, and human-filtered test splits, along with the sample count in the MCQ dataset.}
\label{tab:top10_category_stats}
\end{table}

\begin{table*}[!h]
\centering
\small
\scalebox{0.8}{
\begin{tabular}{p{3cm}p{12cm}}
\toprule
\textbf{Category (Number of unique aspects)} & \textbf{Aspects} \\
\midrule
Social Etiquette (42) & Greetings, Names and Titles, Handshakes, Eye Contact, Introductions, Personal Space, Gestures \& Hand Gestures, Expressing Thanks, Expressing Refusal, Seating, Joking \& Humor, Public Affection, Conversation / Asking Questions, Language Use, Meeting Time / Punctuality, Public Behavior / Street Interactions, Queuing / Queue Etiquette, Communication Style, Using Arabic/French, Touching Others, Greeting with ``Salaam Alaikum'', Flexible Time / Punctuality, Urban/Rural Dialect Switching, Respect for Elders, Showing Soles of Feet, Right-Hand Use for Giving/Receiving, Avoiding Pointing with Index Finger, Silence and Pauses in Conversation, Indirect Communication / Saving Face, Honor and Shame Sensitivity, Compliments \& ``Mashallah'' Responses, Asking About Family (Rapport-Building), Taboo Topics (politics/sectarian issues), Public Displays of Anger, Loud Laughter in Public, Queue Courtesy for Families/Elders, Greeting Order When Entering a Room, Door Etiquette (who enters first), Avoid Passing Objects Over Someone's Head, Accommodating Left-Handed Guests, Respectful Address to Domestic Workers, Sports Events Etiquette (football matches) \\
\addlinespace

Hospitality \& Home Life (25) & Visiting Homes, Shoes in Homes, Invitations, Gifts, Hospitality, Thanking Host, Inviting to Eat, Leaving a Gathering, Hospitality to Strangers, Gift-Giving of Sweets, Home Hospitality Rituals, Role of Host in Welcoming, Visiting the Sick (Get-Well Etiquette), Condolences / Mourning Visits, Serving / Seating Elders First, Guest of Honor Seating, Polite Gift Refusal and Re-offer Ritual, Bringing Dates or Fruit to Hosts, Women's Gatherings (Hareem Majlis), Men's Majlis Etiquette, Incense (Bukhoor) for Guests, Gift Wrapping \& Color Preferences, Reciprocity / Return-Gift Norms, Hosting Unannounced Visitors, Sharing Food with Neighbors \\
\addlinespace

Religion \& Rituals (24) & Religious Sites, Religious Items, Ramadan, Prayer Times / Friday Observance / Friday Prayers, Public Worship Spaces, Festivals / Holidays, Wedding Etiquette, Friday (Jumu'ah) Observance, Henna Traditions, Saying ``Bismillah'' Before Eating, Eid al-Fitr Etiquette (Zakat al-Fitr, greetings), Eid al-Adha Etiquette (Qurbani sharing), Removing Shoes in Mosques, Adhan (Call to Prayer) Respect (pausing music), Using ``Inshallah / Alhamdulillah'' Appropriately, Visiting Shrines / Saints' Tombs, Ramadan: Non-Muslim Daytime Eating Etiquette, Prayer Rooms in Malls/Airports Etiquette, Wudu (Ablution) Facility Etiquette, Visiting Cemeteries Etiquette, Using Honorifics (Sheikh, Hajji), Addressing by Kunya (Abu/Umm), Name Order \& ``bin/bint'', Funeral Attire and Participation \\
\addlinespace

Dining \& Food (25) & Table Manners / Dining Etiquette, Bread Etiquette, Food (Dietary Norms), Finishing Food, Tea / Coffee Ritual, Refusing Food / Tea, Street Food, Serving \& Drinking Arabic Coffee (Qahwah), Communal Eating from a Shared Dish, Bread is Sacred, Suhoor \& Iftar Hosting, Offering Dates Before Meals, Eating with the Right Hand, Using Bread as an Utensil, Serving Elders First (Meals), Accepting Second Helpings, Avoiding Food Waste, Shisha / Hookah Etiquette, Street Coffee/Tea Kiosk Etiquette, Spice/Heat Consideration for Guests, Family Sections in Restaurants, Coffee vs Tea Preference by Region, Fruit Plate Ritual After Meals, Alcohol, Cafés \& Coffee Culture \\
\addlinespace

Business \& Work (14) & Business Etiquette / Business Cards / Business Attire, Dealing with Officials, Business Gift-Giving, Relationship-First Negotiation Style, Wasta (Networking \& Intermediation), Decision-Making Hierarchy, Scheduling Around Prayer/Friday, Polite Persistence in Follow-ups, Business Entertaining (non-alcoholic venues), Acceptable Small Talk Topics, Timing of Gifts in Business (after meeting), Business Card Handling (Arabic side up), Office Visiting Hours \& Drop-ins, Formal Greetings in Email/Messaging \\
\addlinespace

Modern \& Technology (9) & Mobile Phones, Texting/Messaging, Social Media Etiquette, Smoking Etiquette, Photography of Children, Saving Contact Names with Titles (phones), WhatsApp Voice Notes Etiquette, Livestreaming in Public Etiquette, Sharing Religious Content Online Etiquette \\
\addlinespace

Gender \& Travel (9) & Women Travelers, Men Travelers, LGBTQ+ / Queer Travelers, Gender Interaction, Gender-Separate Queues/Sections, Cross-Gender Friendship Boundaries, Chaperoned Socializing, Family vs Mixed Areas at Beaches, Cross-Gender Eye Contact Boundaries \\
\addlinespace

Health \& Environment (8) & Drinking Water, Littering / Environment, Sacredness of Water \& Fountains, Public Bathhouse Traditions, Water Conservation Norms, Fragrance Sensitivity in Public Spaces, Sauna/Steam Room Gender Separation, Picnic \& Park Etiquette \\
\addlinespace

Family \& Children (9) & Children / Traveling with Children, Birth Celebration (Aqiqah) Etiquette, Complimenting Children with ``Mashallah'', Evil Eye (Nazar) Beliefs \& Amulets, Elder Care Expectations, Parenting in Public (discipline norms), Visiting New Mothers Etiquette, Adoption \& Kinship Care Sensitivities, Birthday Celebration Norms \\
\addlinespace

Politics \& Law (5) & Monarchy, Respect for Monarchy \& Royal Imagery, National Day Celebrations, Flag Respect and National Symbols, Military Service Respect \\
\addlinespace

Shopping \& Money (6) & Carrying Cash, Tourist Pricing, Marketplace Haggling as Social Ritual, Money / Carrying Cash, Shopping / Marketplace Interaction / Bartering, Tipping (Baksheesh) \\
\addlinespace

Charity \& Society (7) & Beggars \& Street Kids, Charity / Street Giving, Charity (Zakat/Sadaqah) Giving Etiquette, Community Fundraising \& Mutual Aid, Recognizing Tribal/Clan Affiliations, Poetry \& Oral Storytelling at Gatherings, Camel \& Horse Traditions \\
\addlinespace

Personal Appearance (7) & Dress Code, Distinct Traditional Clothing, Mosque \& Shrine Dress Expectations, Swimwear \& Beach Modesty, Traditional Headwear Norms (keffiyeh/hijab), Tattoos \& Piercings Perceptions, Perfume/Oud Use in Social Settings \\
\addlinespace

Transport (4) & Taxi Etiquette, Public Transport / Transport Etiquette, Taxi: Front vs Back Seat Choice, Carpooling \& Splitting Costs \\
\addlinespace

Safety (3) & Personal Belongings, Street Harassment, Lost and Found \\
\addlinespace

Pets \& Environment (3) & Animals \& Pets, Pet Dogs in Public Spaces Acceptability, Compassion Toward Stray Animals \\
\midrule
\textbf{Total} & \textbf{16 categories and 200 unique aspects} \\
\bottomrule
\end{tabular}
}
\caption{Cultural categories and aspects of ArabPref. The taxonomy contains 16 categories and 200 unique aspects. Duplicate aspect entries in the mapping were counted once.}
\label{tab:cultural_categories_aspects}
\end{table*}

\subsection{Overall Data Statistics}
\label{apdx:data_statistics}

Table~\ref{tab:preference_dataset_stats_country} shows the number of preference samples across the training, raw test, and human-filtered test splits for all 22 Arab countries, with human annotators filtering prompt-chosen-reject triples. We ask annotators to check each triple and only retain those examples that reflect the local culture. 

Table~\ref{tab:top10_category_stats} presents the distribution of categories in the preference and MCQ datasets.

\subsection{Intra-Arab Cultural Variation Across Shared Aspects}
\label{apdx:intra_arab}


\begin{table*}[!h]
\centering
\scalebox{0.9}{
\begin{tabular}{llp{9cm}l}
\toprule
\rowcolor{gray!30}
Aspect & Country & Preferred Norm (paraphrased) & Stance \\
\midrule

\multirow{8}{*}{Alcohol}
 & Saudi Arabia & Strictly prohibited; can lead to severe penalties. & Prohibited \\
 & Kuwait       & Strictly prohibited; avoid bringing any alcohol.   & Prohibited \\
 & Libya        & Illegal; avoid bringing or offering alcohol.       & Prohibited \\
 & Algeria      & Safest to avoid possession entirely.               & Restricted \\
 & UAE          & Restricted to licensed venues; drink responsibly.  & Licensed only \\
 & Jordan       & Acceptable at licensed restaurants and hotels.     & Licensed only \\
 & Bahrain      & Can enjoy a beer at licensed restaurants/hotels.   & Permissible \\
 & Lebanon      & Liberal consumption; acceptable in most restaurants. & Liberal \\

\midrule

\multirow{7}{*}{Dress Code}
 & Yemen        & Men: thobe and turban; Women: jilbab/abaya.       & Traditional \\
 & Saudi Arabia & Women: abaya/hijab advised; conservative throughout. & Conservative \\
 & Iraq         & Cover arms and legs; chador in holy cities.        & Conservative \\
 & Egypt        & Cover shoulders and knees; avoid tight clothing.   & Modest Western \\
 & Algeria      & Modest professional attire; shoulders/knees covered. & Modest Western \\
 & Lebanon      & Smart casual or slightly formal; flexible.         & Smart casual \\
 & Tunisia      & Modest and respectful; avoid attire that attracts attention. & Smart casual \\

\midrule

\multirow{6}{*}{Affection}
 & Saudi Arabia & Refrain from hand-holding; express affection privately. & Prohibited \\
 & Egypt        & Avoid all public displays, including hand-holding.   & Prohibited \\
 & UAE          & Hand-holding acceptable; hugging discouraged.       & Restricted \\
 & Morocco      & Respect personal space; exercise discretion.        & Restricted \\
 & Tunisia      & Hand-holding acceptable for married couples in urban areas. & Conditional \\
 & Lebanon      & Hand-holding while walking is generally acceptable. & Permissible \\

\midrule

\multirow{6}{*}{Greetings}
 & Jordan       & Stand for elders; warm handshake and inquiry.      & Elder-focused \\
 & Kuwait       & Rise for elders; wait for elder to initiate contact. & Elder-focused \\
 & Sudan        & Men embrace; wait for women to initiate handshakes. & Gender-diff. \\
 & Tunisia      & Handshake for men; nod or wait for women's cue.     & Gender-diff. \\
 & Lebanon      & ``Marhaba''; warm handshake and informal welcome.   & Informal \\
 & Yemen        & General wave or nod acceptable in large gatherings. & Pragmatic \\

\bottomrule
\end{tabular}
}
\caption{Examples of divergent preferred norms across Arab nations for four cultural aspects in \texttt{ArabPref}. Preferred responses are condensed from dataset entries. The \textit{Stance} column summarizes the normative position along the observed spectrum.}
\label{tab:intra_arab}
\end{table*}

Table~\ref{tab:intra_arab} illustrates how preferred norms diverge systematically across Arab nations for four aspects shared by 15 or more countries in \ds. Despite a shared linguistic and religious tradition, countries exhibit distinct normative stances reflecting unique legal and social contexts.

\section{Implementation Details}
\subsection{LLM Inference.}
For open-weight LLMs, we use vLLM~\cite{10.1145/3600006.3613165} with temperature 0.7 and top-$p$ 1.0 for preference generation, and temperature 0 for MCQ solving. We set the context and generation limits to 2,048 (max\_model\_len=2048) and 512 tokens (max\_tokens=512), respectively. We evaluate MCQs using lm-evaluation-harness~\cite{eval-harness}. These settings standardize the generated outputs and facilitate consistent downstream evaluation. All experiments are conducted on two NVIDIA A100 GPUs, each with 80\,GB of memory. For proprietary model and evaluator, we use OpenAI's \texttt{gpt-5.4-2026-03-05} and \texttt{gpt-4o-2024-11-20}, DeepSeek's \texttt{deepseek-v4-flash}, and Alibaba Cloud's \texttt{qwen3.6-plus-2026-04-02}.

\subsection{Licenses}
\begin{itemize}
    \item \textbf{Apache 2.0:}
    Qwen2.5-7B-Instruct~\cite{qwen2025qwen25technicalreport},
    Qwen3-8B~\cite{yang2025qwen3technicalreport} and
    Jais-Adapted-7B-Chat~\cite{sengupta2023jais}

    \item \textbf{Llama Community License:}
    Llama-3.1-8B-Instruct~\cite{grattafiori2024llama3herdmodels}.

    \item \textbf{MIT:}
    DeepSeek-v4-Flash~\cite{deepseekai2026deepseekv4highlyefficientmilliontoken}.

    \item \textbf{OpenAI Services Agreement:}
    GPT-5.4~\cite{gpt5} and GPT-4o~\cite{openai2024gpt4technicalreport}.

    \item \textbf{Alibaba Cloud Model Studio Terms:}
    Qwen3.6-Plus~\cite{yang2025qwen3technicalreport}.

    \item \textbf{Anthropic Consumer Terms of Service:}
Claude-Sonnet-4~\cite{anthropic2025system}.

\end{itemize}

\subsection{Training Details}
We train all models using LoRA~\cite{hu2021lora} (\(\text{rank}=8\), \(\alpha=16\), applied to all linear modules) using the AdamW~\cite{loshchilov2019decoupledweightdecayregularization} optimizer (\(\beta_1=0.9\), \(\beta_2=0.999\), \(\epsilon=10^{-8}\)).
We use a per-device batch size of 1 and 8 gradient accumulation steps on 2 GPUs, resulting in an effective batch size of 16.
Supervised fine-tuning uses a learning rate of \(1 \times 10^{-4}\), while DPO and KTO use \(5 \times 10^{-6}\).
Each model is trained for 3 epochs with a cosine scheduler, a warmup ratio of 0.1, and a weight decay of 0.
For DPO and KTO, we set the value of \(\beta\) to 0.1. All experiments use BF16 precision and a maximum sequence length of 2,048 tokens.



\section{Evaluation Criteria}
\label{apdx:evaluation_criterion}

We use the following evaluation criteria:
\begin{itemize}
    \item \textbf{Cultural Appropriateness}: This evaluation criterion assesses the degree to which a response adheres to the specific norms, values, and social sensitivities of the target locale. A high value indicates that the content is culturally authentic and respectful, while lower scores reflect the presence of stereotypes, cultural misunderstandings, or insensitive references.

    \item \textbf{General Quality}: This evaluation criterion assesses response quality across three dimensions: factual accuracy, depth of explanation, and relevance. Unlike Cultural Appropriateness, which focuses on locale-specific norms, General Quality evaluates whether the response is correct, sufficiently detailed, and focused without extraneous information. A gold reference answer serves as the evaluation baseline.
    
    \item \textbf{Literacy}: This evaluation criterion assesses the linguistic fluency, grammatical correctness, and syntactic coherence of the output. It measures whether the model's generation is natural and readable, ensuring that the cultural message is not obscured by stylistic or grammatical errors.
\end{itemize}

In addition to the above three evaluation criteria, we further use additional ones for human annotation and prompts for LLM-as-a-judge evaluation, which can be found in Table~\ref{tab:evaluation_criterion}.

\begin{table*}[ht!]
\centering
\small
\renewcommand{\arraystretch}{1.3} 
\begin{tabular}{lp{4.2cm}p{4.2cm}p{4.2cm}}
\toprule
\textbf{Score} & \textbf{Cultural Appropriateness} & \textbf{General Quality} & \textbf{Literacy} \\
\midrule
\textbf{1} & Completely misaligned with the cultural context, containing offensive or highly inappropriate references. & Very incomplete, with critical information missing, making the response unusable. & Very poor fluency, riddled with errors, making the text nearly unreadable or incoherent. \\
\midrule
\textbf{2} & Major cultural misunderstandings or insensitive references that impact the overall understanding. & Significant omissions that make the instruction or question difficult to interpret or incomplete. & Multiple errors that hinder readability or cause significant confusion for the reader. \\
\midrule
\textbf{3} & Noticeable cultural inconsistencies or slight misinterpretations, but the general meaning remains clear. & Noticeable gaps in information or coverage that leave the instruction or question lacking. & Noticeable errors in grammar, spelling, or structure, but the text is still understandable. \\
\midrule
\textbf{4} & Minor cultural inaccuracies that do not significantly affect the overall understanding. & Slightly incomplete, with minor missing details that do not impact the overall understanding. & Minor grammatical or stylistic errors that do not significantly affect readability. \\
\midrule
\textbf{5} & Fully aligned with the cultural context, showing deep respect for cultural norms and values. & Fully complete, with no missing information, thoroughly covering the required context. & Perfect fluency, with no errors in grammar, spelling, or structure. The text reads naturally. \\
\bottomrule
\end{tabular}
\caption{Evaluation criterion for human annotation and LLM-as-a-judge.}
\label{tab:evaluation_criterion}
\end{table*}.


\newpage

\section{Alignment Methods}
\label{apdx:alignment_methods}
We use supervised fine-tuning (SFT), Direct Preference Optimization
(DPO), and Kahneman--Tversky Optimization (KTO) for the full-data
experiments. We additionally evaluate Demonstration ITerated Task
Optimization (DITTO) in the low-data sample-efficiency study only.

Let $x$ denote a prompt, $y$ a response, and $\pi_\theta$ the policy
model. We use $\log \pi_\theta(y \mid x)$ to denote the sequence-level
log-probability of $y$ conditioned on $x$.

\paragraph{SFT.}
Given a set of demonstrations
$\mathcal{D}_{\mathrm{SFT}}=\{(x,y)\}$, SFT minimizes the negative
log-likelihood of the target responses:
\begin{equation}
\mathcal{L}_{\mathrm{SFT}}(\theta)
=
-\mathbb{E}_{(x,y)\sim\mathcal{D}_{\mathrm{SFT}}}
\left[\log \pi_\theta(y \mid x)\right].
\end{equation}

\paragraph{DPO.}
For a preference pair $(x,y^{+},y^{-})$, where $y^{+}$ is preferred to
$y^{-}$, DPO compares the policy with a fixed reference policy
$\pi_{\mathrm{ref}}$. Its objective is
\begin{equation}
\Delta_\theta(x,y^{+},y^{-})
=
R_\theta(x,y^{+})-R_\theta(x,y^{-}).
\end{equation}

\begin{equation}
\begin{aligned}
\mathcal{L}_{\mathrm{DPO}}(\theta)
= - \mathbb{E}_{(x,y^{+},y^{-})}
\big[
\log \sigma\big(
\beta \Delta_\theta(x,y^{+},y^{-})
\big)
\big].
\end{aligned}
\end{equation}
where $\sigma$ is the logistic function and $\beta=0.1$ controls the
strength of the preference objective.

\paragraph{KTO.}
For each preference pair $(x,y^{+},y^{-})$, we construct two labeled
examples, $(x,y^{+},1)$ and $(x,y^{-},0)$, where $1$ and $0$ denote
desirable and undesirable responses, respectively. In our LoRA
implementation, $\pi_{\mathrm{ref}}$ is the base model with the adapter
disabled. We define
\begin{equation}
r_\theta(x,y)
=
\log \pi_\theta(y\mid x)
-
\log \pi_{\mathrm{ref}}(y\mid x).
\end{equation}
For a minibatch of size $B$, we estimate the KTO reference point as
\begin{equation}
\hat z_0
=
\operatorname{stopgrad}
\left[
\max\left(
0,
\frac{1}{B}\sum_{i=1}^{B}
r_\theta(x_i,\tilde y_i)
\right)
\right],
\end{equation}
where $\tilde y_i$ is a response cyclically shifted from another example
in the same minibatch. The loss for an example with label $l$ is
\begin{equation}
d_\theta(x,y)
=
\beta\bigl(r_\theta(x,y)-\hat z_0\bigr).
\end{equation}

\begin{equation}
\ell_{\mathrm{KTO}}(x,y,l)
=
\begin{cases}
\lambda_{+}\!\left[1-\sigma\bigl(d_\theta(x,y)\bigr)\right],
 l=1,\\
\lambda_{-}\!\left[1-\sigma\bigl(-d_\theta(x,y)\bigr)\right],
 l=0.
\end{cases}
\end{equation}
We minimize the minibatch mean of $\ell_{\mathrm{KTO}}$ and use
$\beta=0.1$ and $\lambda_{+}=\lambda_{-}=1.0$.

\paragraph{DITTO.}
We use DITTO only in the low-data study. Let
$\mathcal{D}_{E}=\{(x,y_E)\}$ denote the demonstration set. We first
obtain an initial policy $\pi_0$ by applying SFT to
$\mathcal{D}_{E}$ and keep $\pi_{\mathrm{ref}}=\pi_0$ fixed throughout
the subsequent iterations. At iteration $t$, the current policy
$\pi_t$ generates $M$ responses for each demonstration prompt, producing
an online dataset $\mathcal{D}_t$. We construct preference comparisons
according to
\begin{equation}
\mathcal{D}_{E}
\succ
\mathcal{D}_{t}
\succ
\mathcal{D}_{t-1}
\succ
\cdots
\succ
\mathcal{D}_{0},
\end{equation}
where $\succ$ denotes that the responses on the left are preferred to
those on the right. The resulting comparison set contains online pairs
between demonstrations and current outputs, replay pairs between
demonstrations and earlier outputs, and inter-model pairs between later
and earlier policy outputs.

We update $\pi_t$ with the DPO objective
\begin{equation}
\Delta_\theta^{t}(x,y^{w},y^{l})
=
R_\theta(x,y^{w})-R_\theta(x,y^{l}).
\end{equation}

\begin{equation}
\begin{aligned}
\mathcal{L}(\theta)
=
-\mathbb{E}_{(x,y^{w},y^{l})\sim\mathcal{C}_t}
\big[
\log \sigma\big(
\alpha \Delta_\theta^{t}(x,y^{w},y^{l})
\big)
\big].
\end{aligned}
\end{equation}
where $\mathcal{C}_t$ is the comparison set at iteration $t$, $y^{w}$
and $y^{l}$ are the preferred and dispreferred responses, respectively,
and $\alpha$ is the DPO temperature. We report DITTO only as a
sample-efficiency analysis because it is designed for adaptation from a
small number of demonstrations rather than full-data post-training.

\section{Sample Efficiency}
\label{apdx:sample_efficiency}

\begin{figure}[!h]
    \includegraphics[width=\linewidth]{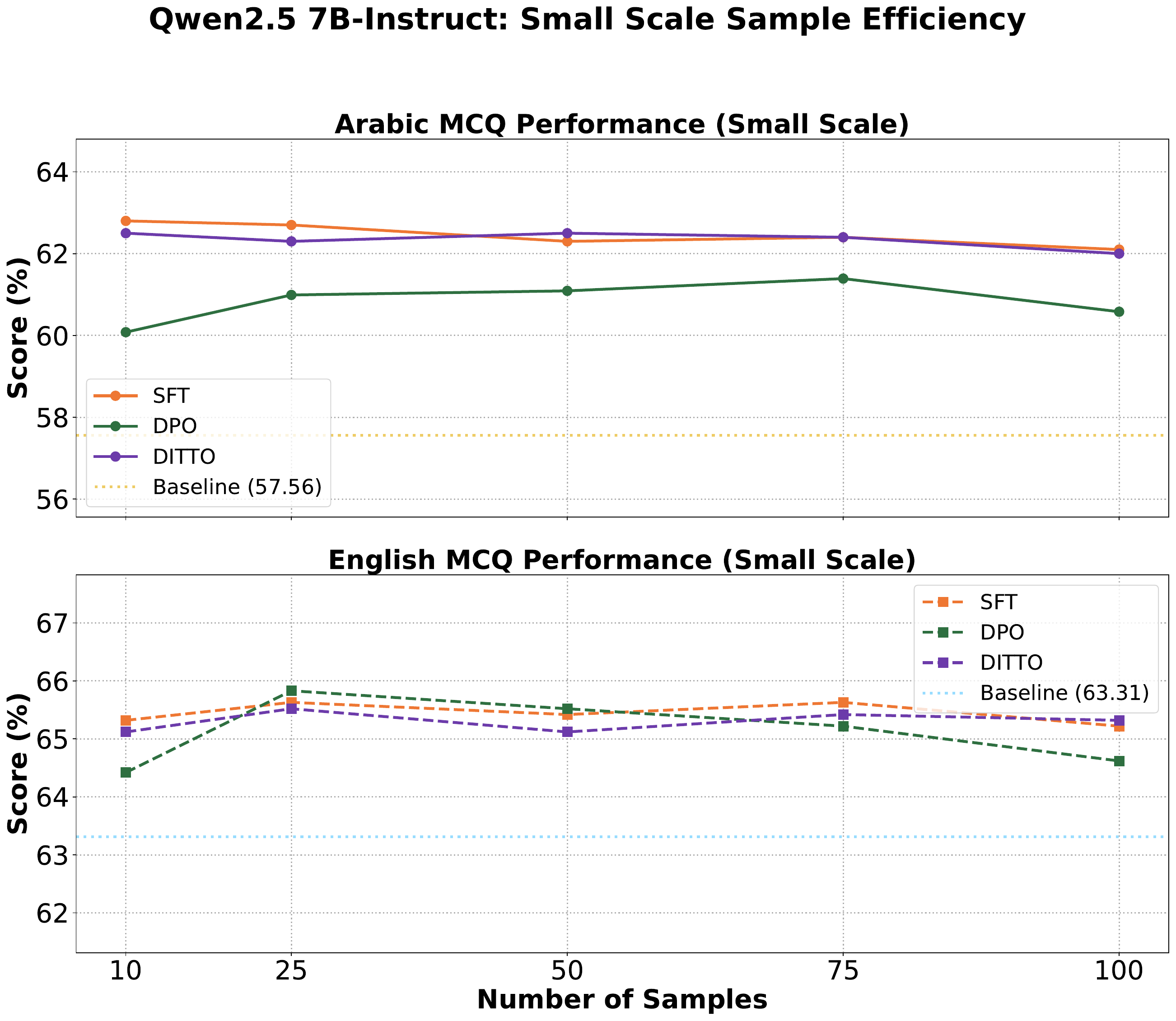}
    \caption{Sample efficiency of preference data alignment on Qwen2.5-7B-Instruct, evaluated on MCQs. Qwen2.5-7B-Instruct performance with SFT, DPO, and DITTO (10–100 samples). Dotted lines show zero-shot baselines.}
    \label{fig:sample_efficiancy_main}
\end{figure}

\begin{figure}[!h]
    \centering
    \includegraphics[width=\linewidth]{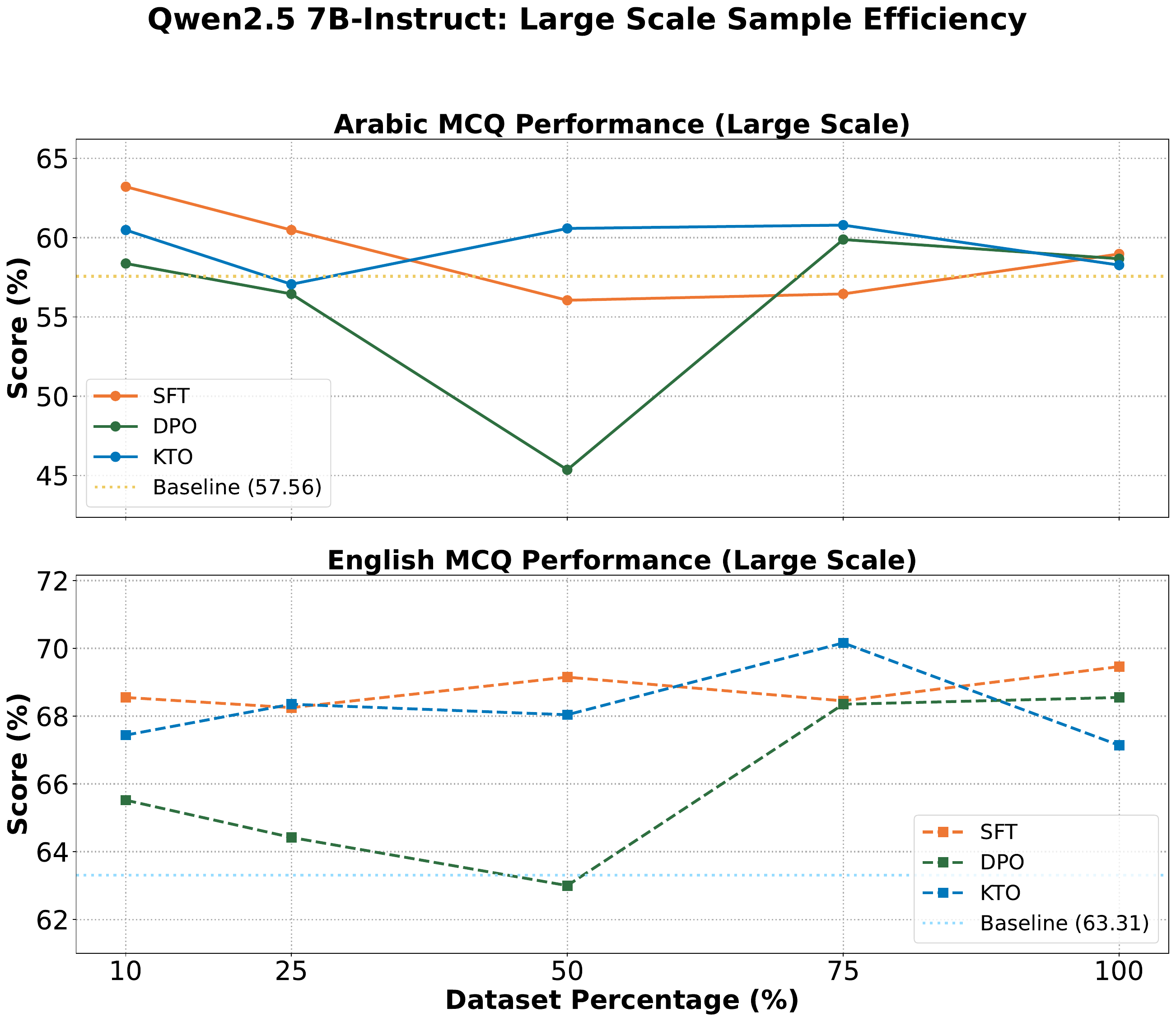}
    \caption{Sample efficiency of alignment methods on Arabic and English cultural preference MCQ evaluation. Qwen2.5-7B-Instruct performance across SFT, DPO, and KTO with 10–100\% of training data. Baselines (dotted) show zero-shot performance.}
    \label{fig:sample_efficiancy_2}
\end{figure}

\begin{figure}[!h]
    \centering
    \includegraphics[width=\linewidth]{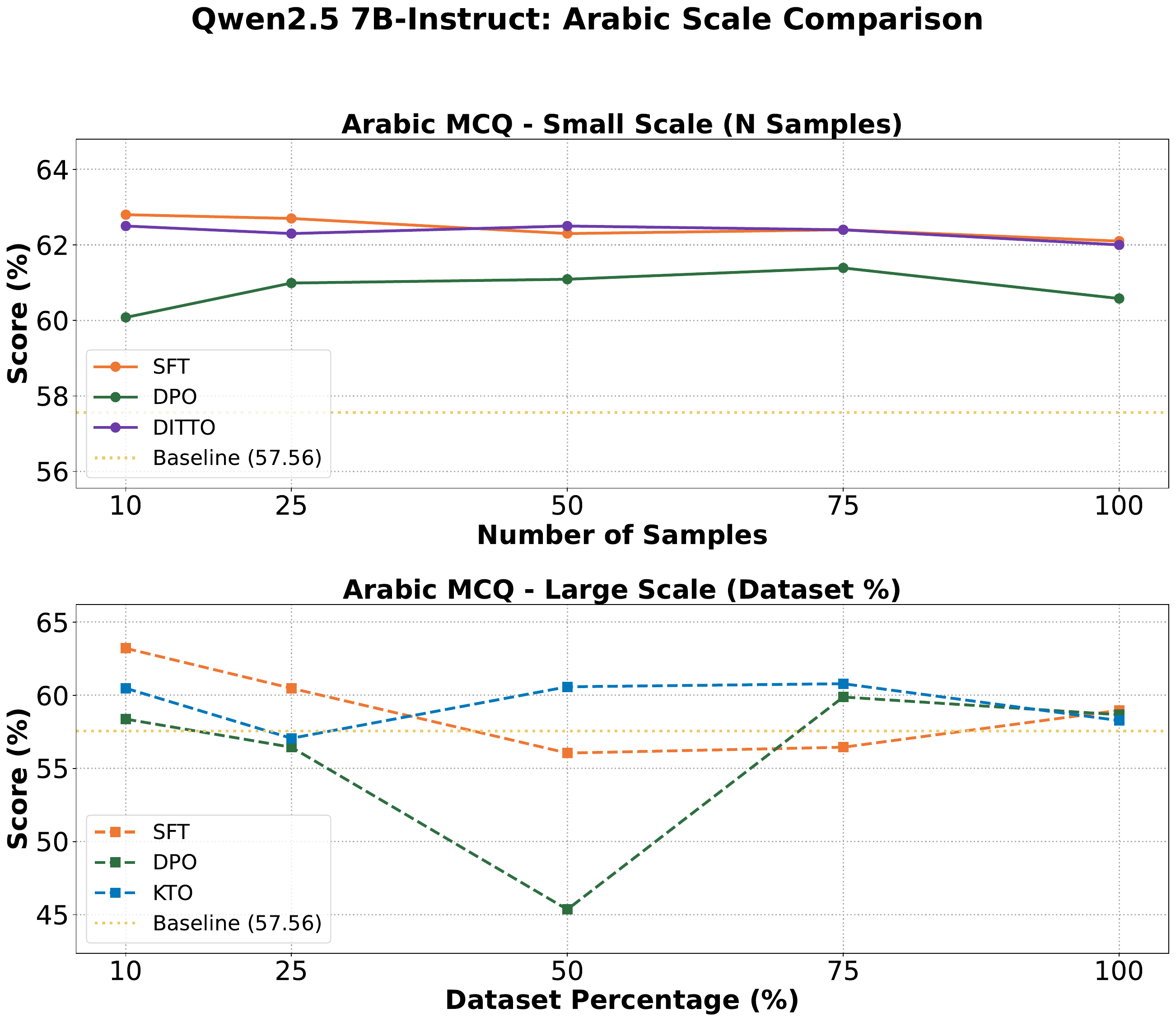}
    \caption{Scale comparison of alignment methods on Arabic cultural preference MCQ evaluation. Qwen2.5-7B-Instruct performance in small-scale (top; 10–100 samples with SFT, DPO, DITTO) and large-scale (bottom; 10–100\% data with SFT, DPO, KTO) settings. Baselines (dotted) show zero-shot performance.}
    \label{fig:sample_efficiancy_3}
\end{figure}

\begin{figure}[!h]
    \centering
    \includegraphics[width=\linewidth]{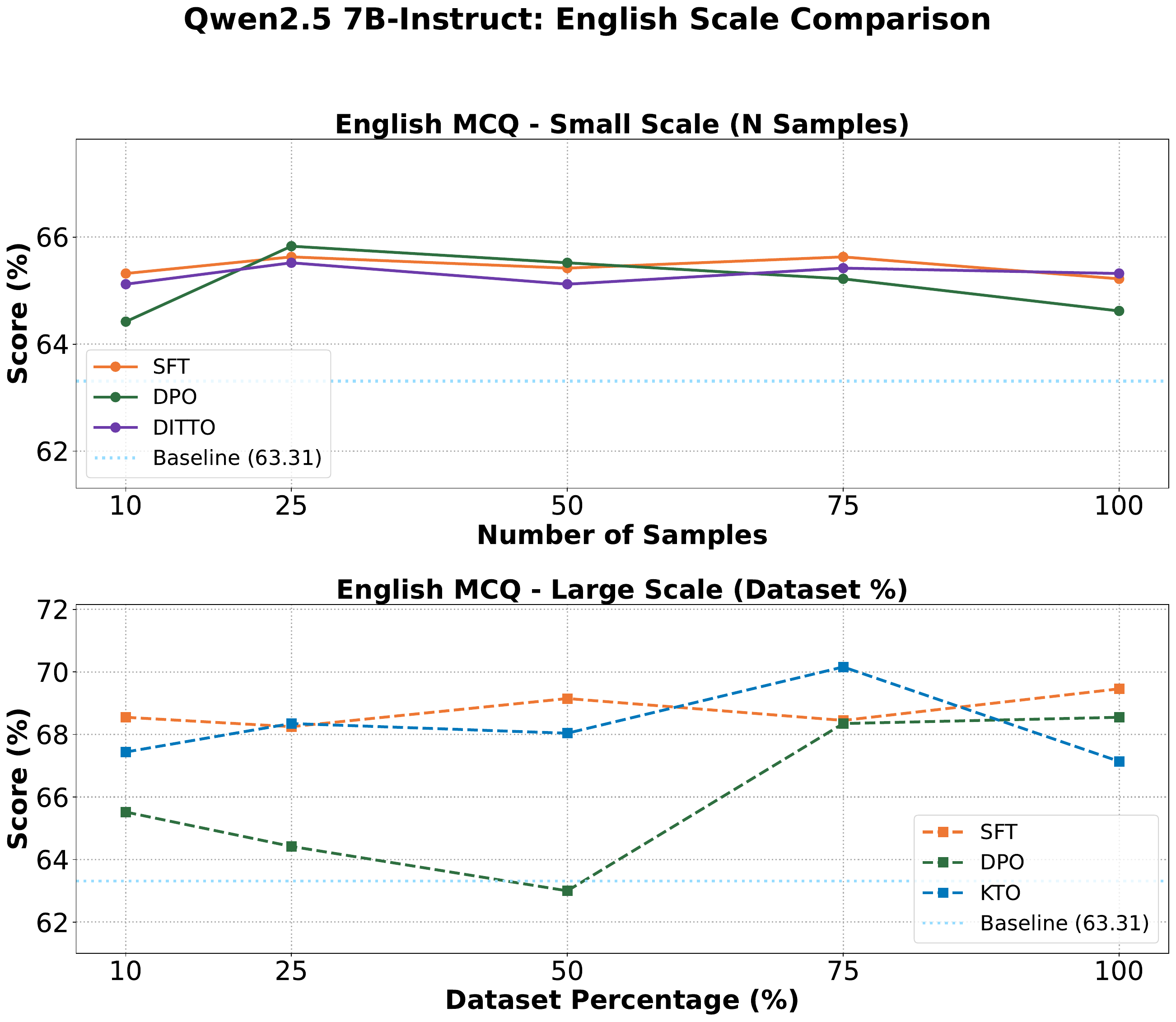}
    \caption{Scale comparison of alignment methods on English cultural preference MCQ evaluation. Qwen2.5-7B-Instruct performance in small-scale (top; 10–100 samples with SFT, DPO, DITTO) and large-scale (bottom; 10–100\% data with SFT, DPO, KTO) settings. Baselines (dotted) show zero-shot performance.}
    \label{fig:sample_efficiancy_4}
\end{figure}

Figure~\ref{fig:sample_efficiancy_main} illustrates the sample efficiency of various alignment methods on both Arabic and English, using 10–100 training samples from the \ds+\dataset{CARE} dataset for SFT and DPO, and 10–100 samples for DITTO. All methods outperform the baseline, with SFT and DITTO showing  better performance than DPO. While these methods offer efficient training with fewer samples, their performance still lags behind full-scale training for both SFT and DPO.

Figures~\ref{fig:sample_efficiancy_2} through \ref{fig:sample_efficiancy_4} provide a comprehensive analysis of how alignment performance scales across language and tasks. Figure~\ref{fig:sample_efficiancy_2} compares the sample efficiency of SFT, DPO, and KTO between Arabic and English contexts using 10--100\% of the training data,  Figures~\ref{fig:sample_efficiancy_3} and \ref{fig:sample_efficiancy_4} offer a bifurcated view of the scaling dynamics in Arabic and English, respectively.

\begin{table*}[!h]
\centering
\small
\setlength{\tabcolsep}{4.5pt}
\begin{tabular}{lcccccccc}
\toprule
\multirow{2}{*}{Method} 
& \multicolumn{4}{c}{English} 
& \multicolumn{4}{c}{Arabic} \\
\cmidrule(lr){2-5} \cmidrule(lr){6-9}
& Cultural & General & Literacy & Avg. 
& Cultural & General & Literacy & Avg. \\
\midrule
Vanilla 
& 3.41 & 3.58 & 4.07 & 3.69 
& 3.89 & 4.08 & 3.98 & 3.98 \\

\dataset{Palm} (SFT) 
& 4.02 & 4.28 & 4.57 & 4.29 
& 3.36 & 3.53 & 3.68 & 3.52 \\

\dataset{CARE} (DPO) 
& 4.28 & \textbf{4.55} & 4.06 & 4.30 
& \textbf{4.21} & \textbf{4.46} & 3.66 & \textbf{4.11} \\

\dataset{CARE} (KTO) 
& 4.11 & 4.37 & 4.00 & 4.16 
& 4.19 & 4.44 & 3.57 & 4.07 \\

\ds$_{\text{sampled}}$ (DPO) 
& 4.02 & 4.26 & 4.33 & 4.20 
& 3.87 & 4.14 & 4.09 & 4.03 \\

\ds$_{\text{sampled}}$ (KTO) 
& 3.87 & 4.31 & \textbf{4.65} & 4.28 
& 3.54 & 3.98 & 4.13 & 3.88 \\

\ds$_{\text{sampled}}$+\dataset{CARE} (DPO) 
& \textbf{4.50} & 4.32 & 4.37 & \textbf{4.40} 
& 4.01 & 3.87 & 4.01 & 3.96 \\

\ds$_{\text{sampled}}$+\dataset{CARE} (KTO) 
& 4.08 & 4.14 & 4.26 & 4.16 
& 3.67 & 3.96 & \textbf{4.14} & 3.92 \\

\ds$_{\text{sampled}}$+\dataset{CARE} (SFT) 
& 4.17 & 4.35 & 4.23 & 4.25 
& 3.76 & 4.23 & 4.05 & 4.01 \\
\bottomrule
\end{tabular}
\caption{
Scale-controlled sampled-data ablation on Qwen-2.5-7B-Instruct. 
We construct \ds$_{\text{sampled}}$ by sampling 30 preference pairs per country across diverse aspects, in order to reduce the data-volume advantage of combining \ds with \dataset{CARE}. 
Under this controlled setting, \ds$_{\text{sampled}}$+\dataset{CARE} with DPO achieves the best English Cultural score and English average score, outperforming the \dataset{CARE}-only DPO baseline. 
This indicates that \ds contributes complementary cultural preference signals beyond merely increasing the amount of training data.
}
\label{tab:sampled_ablation}
\end{table*}

To test whether combining \ds and \dataset{CARE} helps only through scale, we conduct a sampled-data ablation with Qwen-2.5-7B-Instruct. Specifically, we construct \ds$_{\text{sampled}}$ by sampling 30 preference pairs per country across diverse cultural aspects, yielding a substantially smaller and controlled subset of \ds. Subsequently, we evaluate models trained with \ds$_{\text{sampled}}$, \dataset{CARE}, and their combination.

Table~\ref{tab:sampled_ablation} shows that \ds$_{\text{sampled}}$+\dataset{CARE} with DPO achieves the highest English Cultural score and English average score, outperforming the \dataset{CARE}-only DPO baseline despite the controlled data setting. 

This scale-controlled data ablation strongly suggests that the observed performance improvement is not simply a byproduct of adding more training instances; rather, \ds introduces uniquely nuanced, complementary cultural preference signals that substantially improve cross-lingual cultural alignment, particularly during downstream English evaluation. 

At the same time, \dataset{CARE}-only DPO optimization remains the strongest baseline on the overall Arabic average score, indicating that these two distinct data resources contribute entirely different and complementary behavioral alignment signals across languages.

\section{Correlation Coefficiency}
\label{apdx:correlation_coefficiency}
\begin{table}[!t]
\centering
\begin{tabular}{lccc}
\toprule
$\rho$ & Cultural & General & Literacy \\
\midrule
En & 0.7140 & 0.8312 & 0.8950 \\
Ar & 0.6707 & 0.7801 & 0.7398 \\
\bottomrule
\end{tabular}
\caption{Spearman's $\rho$ values for different categories (Cultural, General, Literacy) for both English (En) and Arabic (Ar).}
\label{tab:rho_values}
\end{table}
\textbf{Correlation in data annotation}
We sample 100 preferences from English and Arabic, with 20 samples from each of the following countries: Egypt, Syria, Morocco, UAE, and Tunisia. Human annotators, all of whom are native Arabic speakers from these countries, are asked to score the samples based on cultural appropriateness, general quality, and literacy. 

The annotators are provided with the evaluation guidelines outlined in Appendix~\ref{apdx:evaluation_criterion} to ensure consistency. The current available results are presented in Table~\ref{tab:rho_values}. Both Arabic and English demonstrated high correlation between the machine scoring and human scoring based on the Spearman's $\rho$ values.

\section{Stability of LLM-as-a-Judge Evaluation}
\label{apdx:stability_of_llm_as_a_judge}

To assess the stability of our LLM-as-a-judge evaluation, we conducted a repeated-scoring audit using 100 country-stratified prompts, comprising 50 English and 50 Arabic prompts. We evaluated two model configurations per prompt, yielding 200 responses. 
Each response was scored on three dimensions, Cultural Appropriateness, General Quality, and Literacy, yielding 600 response, rubric units per repetition. We repeated the evaluation three times using the same judge configuration, resulting in 1,800 independent scoring calls. We measured agreement across the three repetitions using Fleiss' $\kappa$, treating the 1--5 Likert ratings as categorical judgments. The resulting Fleiss' $\kappa$ was $0.9503$, indicating near-perfect agreement across repeated evaluations. This high agreement suggests that the reported results are not substantially affected by stochastic variation in the judge outputs under our evaluation configuration. It also supports the reliability of the judge-based evaluation for comparing model performance across the three dimensions.









\tcbset{
  arabpref prompt/.style={
    colback=black!1,
    colframe=black!45,
    colbacktitle=black!6,
    coltitle=black,
    fonttitle=\bfseries\small,
    fontupper=\footnotesize,
    before upper=\raggedright,
    boxrule=0.45pt,
    arc=1mm,
    left=1.5mm,
    right=1.5mm,
    top=1.2mm,
    bottom=1.2mm,
    toptitle=1mm,
    bottomtitle=1mm,
    before skip=6pt,
    after skip=9pt,
  },
  arabpref generation/.style={
    colback=blue!2,
    colframe=blue!48!black,
    colbacktitle=blue!10,
  },
  arabpref task evaluation/.style={
    colback=green!2,
    colframe=green!38!black,
    colbacktitle=green!10,
  },
  arabpref judge/.style={
    colback=violet!2,
    colframe=violet!45!black,
    colbacktitle=violet!10,
  },
}
\newcounter{prompt}
\newcommand{\promptplaceholder}[1]{\texttt{\{#1\}}}

\section{Prompts for Generation and Evaluation}
\label{apdx:llm_as_a_judge_prompts}

In this section, we provide the exact prompts used for evaluation. Placeholders are typeset in monospace and replaced with the corresponding example fields at inference time. Whitespace, field names, output formats, and parser rules are versioned together with these prompts. For better reproducibility, we recommend using the temperature and sampling settings indicated earlier in the paper and the official provider APIs. 

To facilitate reproducibility, we strictly preserve the system and user message structure used in each experiment. All generated model outputs are parsed according to the specified target format without manual correction. The prompts include explicit task definitions, comprehensive cultural evaluation criteria, and output requirements to ensure consistent interpretation across all evaluated models. We also retain the original ordering of instructions and input fields, since subtle changes to prompt structure may significantly affect downstream model behavior and final evaluation outcomes.






\subsection{MCQ Rewriting}

Prompt~\ref{prompt:mcq-rewriting} rewrites the MCQ options in order to create plausible and culturally grounded distractors while preserving the correct answer. Moreover, it also standardizes the question language and returns the revised example the form of a well-formatted JSON. This prompt removes any explicit country references and subjective interpretations in order to ensure that each option describes behavior in neutral terms.

\vspace{8pt}
\refstepcounter{prompt}
\label{prompt:mcq-rewriting}
\begin{tcolorbox}[
  arabpref prompt,
  arabpref generation,
  title={Prompt \theprompt: MCQ Rewriting Prompt},
]
You are given a JSON object describing a cultural behavior question comparing
two or more locations. Revise the four options (\texttt{opa}, \texttt{opb},
\texttt{opc}, \texttt{opd}) to make them more distractive while keeping the
correct answer unchanged.

\medskip
\textbf{Rules:}
\begin{enumerate}[label=\arabic*), nosep, leftmargin=*]
    \item Correct any grammatical errors in the question.
    \item Unless necessary, keep the correct answer's option text unchanged.
    \item Among the three distractors:
    \begin{itemize}[nosep, leftmargin=*]
        \item One must describe the opposite behavior of the correct answer.
        \item The other two distractors should reflect realistic norms or
        values from the distractor country (or countries) listed in
        \texttt{distractor\_country}, given the aspect; if possible, make them
        different from the correct answer's action.
        \item You may modify the distractors, but do not change their
        positivity.
    \end{itemize}
    \item Ensure all options sound plausible and culturally grounded.
    \item Avoid explicitly mentioning any country names in the options; use
    anonymous descriptions.
    \item If the original text includes comment adjectives (e.g.,
    ``progressive,'' ``modern,'' or ``traditional''), rephrase them using
    neutral language.
    \item If the original text includes an interpretation of behavior (e.g.,
    ``regarded as disrespectful'' or ``it is acceptable''), remove the
    interpretation and describe only the behavior.
    \item Return \textbf{only} the final modified JSON in the same structure as
    the input.
\end{enumerate}
\end{tcolorbox}

\newpage
\subsection{MCQ Answering}

Prompt~\ref{prompt:mcq-answering} presents the model with a question and four
answer options. It instructs the model to select the most appropriate answer
and return only its corresponding letter. 

This restricted output format enables
consistent parsing and evaluation across models. By excluding explanations, the
prompt reduces variation in response structure and supports exact-match scoring.
The same instruction format is applied consistently to all four-choice examples
in the evaluation set.

\vspace{8pt}
\refstepcounter{prompt}
\label{prompt:mcq-answering}
\begin{tcolorbox}[
  arabpref prompt,
  arabpref task evaluation,
  title={Prompt \theprompt: MCQ Answering Prompt},
]
\textbf{Question:} \promptplaceholder{question}

\medskip
\textbf{Options:}

\smallskip
\textbf{A.} \promptplaceholder{option\_a}

\textbf{B.} \promptplaceholder{option\_b}

\textbf{C.} \promptplaceholder{option\_c}

\textbf{D.} \promptplaceholder{option\_d}

\medskip
Please choose the most appropriate answer by returning only the letter
(A, B, C, or D). Only return the letter.
\end{tcolorbox}
\vspace{4pt}

\subsection{Three-Choice MCQ Answering}

Prompt~\ref{prompt:mcq-answering-three} presents the model with a question and
three answer options. It requires the model to select the most appropriate
answer and return only its corresponding letter. This format is used for
examples containing three candidate answers while maintaining consistent
output parsing.

\vspace{8pt}
\refstepcounter{prompt}
\label{prompt:mcq-answering-three}
\begin{tcolorbox}[
  arabpref prompt,
  arabpref task evaluation,
  title={Prompt \theprompt: MCQ Answering Prompt (3-choices)},
]
\textbf{Question:} \promptplaceholder{question}

\medskip
\textbf{Options:}

\smallskip
\textbf{A.} \promptplaceholder{option\_a}

\textbf{B.} \promptplaceholder{option\_b}

\textbf{C.} \promptplaceholder{option\_c}

\medskip
Please choose the most appropriate answer by returning only the letter
(A, B or C). Only return the letter.
\end{tcolorbox}
\vspace{4pt}

\subsection{Pairwise Win-Rate Evaluation}

Prompt~\ref{prompt:pairwise-win-rate} compares two responses to the same
culturally situated question using a reference answer. It prioritizes cultural
appropriateness, followed by general quality and literacy, while discouraging
position and length biases. The restricted output identifies the preferred
response or a tie for consistent win-rate computation.

\vspace{8pt}
\refstepcounter{prompt}
\label{prompt:pairwise-win-rate}
\begin{tcolorbox}[
  arabpref prompt,
  arabpref task evaluation,
  title={Prompt \theprompt: Pairwise Win-Rate Evaluation Prompt},
]
You are an impartial evaluator comparing two responses to the same culturally
situated question.

\medskip
Evaluate in this priority order:
\begin{enumerate}[nosep, leftmargin=*]
    \item Cultural appropriateness for the specified country and scenario,
    using the reference answer as guidance.
    \item General quality: factual accuracy, relevance, and sufficient
    coverage.
    \item Literacy: fluency, grammar, and clarity in the language of the
    question.
\end{enumerate}

Do not prefer a response because it is listed first or because it is longer.
Return a tie only when neither response has a clear overall advantage under
the criteria above.

\medskip
\textbf{Country:} \promptplaceholder{country}

\textbf{Question:} \promptplaceholder{question}

\textbf{Reference Answer:} \promptplaceholder{reference\_answer}

\medskip
\textbf{Response 1:} \promptplaceholder{response\_1}

\textbf{Response 2:} \promptplaceholder{response\_2}

\medskip
Return exactly one character: 1 if Response 1 is better, 2 if Response 2 is
better, or 0 for a tie.
\end{tcolorbox}
\vspace{4pt}

\subsection{Cultural Appropriateness Evaluation}

Prompt~\ref{prompt:cultural-appropriateness} evaluates whether a response
aligns with the target country's cultural norms and values. It considers
cultural sensitivity, appropriateness, and the authenticity of cultural nuance
relative to the reference answer. The judge assigns a rating from 1 to 5 using
the defined rubric. The country and question context are provided to ground the
assessment in the intended cultural setting rather than general conventions.

\vspace{8pt}
\refstepcounter{prompt}
\label{prompt:cultural-appropriateness}
\begin{tcolorbox}[
  arabpref prompt,
  arabpref judge,
  title={Prompt \theprompt: Criterion: Cultural Appropriateness},
]
Please serve as an unbiased evaluator and assess the assistant response only
for \textbf{Cultural Appropriateness}: alignment with the target country's
cultural norms and values, sensitivity to offensive or inappropriate
references, and authenticity of cultural nuance.

\medskip
Use this scale:
\begin{itemize}[nosep, leftmargin=*]
    \item \textbf{1} --- Completely misaligned or highly inappropriate.
    \item \textbf{2} --- Major cultural misunderstandings or insensitive
    references.
    \item \textbf{3} --- Noticeable cultural inconsistencies, but the main
    meaning is clear.
    \item \textbf{4} --- Mostly aligned, with only minor cultural inaccuracies.
    \item \textbf{5} --- Fully aligned with the supplied cultural reference.
\end{itemize}

\medskip
\textbf{Country:} \promptplaceholder{country}

\textbf{Question:} \promptplaceholder{question}

\textbf{Reference Answer:} \promptplaceholder{reference\_answer}

\textbf{Assistant's Response:} \promptplaceholder{response}

\medskip
Return exactly: \texttt{Rating: [[number]]}
\end{tcolorbox}
\vspace{4pt}

\subsection{Literacy Evaluation}

Prompt~\ref{prompt:literacy} evaluates the linguistic quality of a response in
the language of the question. It considers fluency, grammar, spelling, clarity,
and natural flow without assessing response content. The judge assigns a rating
from 1 to 5 using the defined rubric.

\vspace{8pt}
\refstepcounter{prompt}
\label{prompt:literacy}
\begin{tcolorbox}[
  arabpref prompt,
  arabpref judge,
  title={Prompt \theprompt: Criterion: Literacy},
]
Please serve as an unbiased evaluator and assess the assistant response only
for \textbf{Literacy} in the language of the question: fluency, grammar and
spelling, clarity, and natural linguistic flow.

\medskip
Use this scale:
\begin{itemize}[nosep, leftmargin=*]
    \item \textbf{1} --- Very poor fluency; errors make the response unreadable
    or incoherent.
    \item \textbf{2} --- Multiple errors substantially hinder readability.
    \item \textbf{3} --- Noticeable errors, but the response remains
    understandable.
    \item \textbf{4} --- Fluent and clear, with only minor errors.
    \item \textbf{5} --- Fully fluent, natural, and free of material language
    errors.
\end{itemize}

\medskip
\textbf{Question language:} \promptplaceholder{language}

\textbf{Question:} \promptplaceholder{question}

\textbf{Assistant's Response:} \promptplaceholder{response}

\medskip
Return exactly: \texttt{Rating: [[number]]}
\end{tcolorbox}
\vspace{4pt}

\subsection{General Quality Evaluation}

Prompt~\ref{prompt:general-quality} evaluates the overall quality of a response
based on factual accuracy, depth, completeness, and relevance. It uses the
reference answer as guidance while allowing alternative correct formulations.
The judge assigns a rating from 1 to 5 using the defined rubric.

\vspace{8pt}
\refstepcounter{prompt}
\label{prompt:general-quality}
\begin{tcolorbox}[
  arabpref prompt,
  arabpref judge,
  title={Prompt \theprompt: Criterion: General Quality},
]
Please serve as an unbiased evaluator and assess the assistant response only
for \textbf{General Quality}: factual accuracy, depth and completeness, and
relevance to the question. Use the reference answer as guidance, while allowing
other correct formulations.

\medskip
Use this scale:
\begin{itemize}[nosep, leftmargin=*]
    \item \textbf{1} --- Highly inaccurate, unusable, or irrelevant.
    \item \textbf{2} --- Partially accurate but missing major information or
    containing major errors.
    \item \textbf{3} --- Mostly addresses the question with adequate detail
    and some omissions.
    \item \textbf{4} --- Accurate and relevant, with only minor omissions.
    \item \textbf{5} --- Fully accurate, relevant, and sufficiently complete.
\end{itemize}

\medskip
\textbf{Country:} \promptplaceholder{country}

\textbf{Question:} \promptplaceholder{question}

\textbf{Reference Answer:} \promptplaceholder{reference\_answer}

\textbf{Assistant's Response:} \promptplaceholder{response}

\medskip
Return exactly: \texttt{Rating: [[number]]}
\end{tcolorbox}
\vspace{8pt}

\section{Annotation Guidelines}
\label{apdx:annotation_guidelines}

\begin{figure*}
    \centering
    \includegraphics[width=\linewidth]{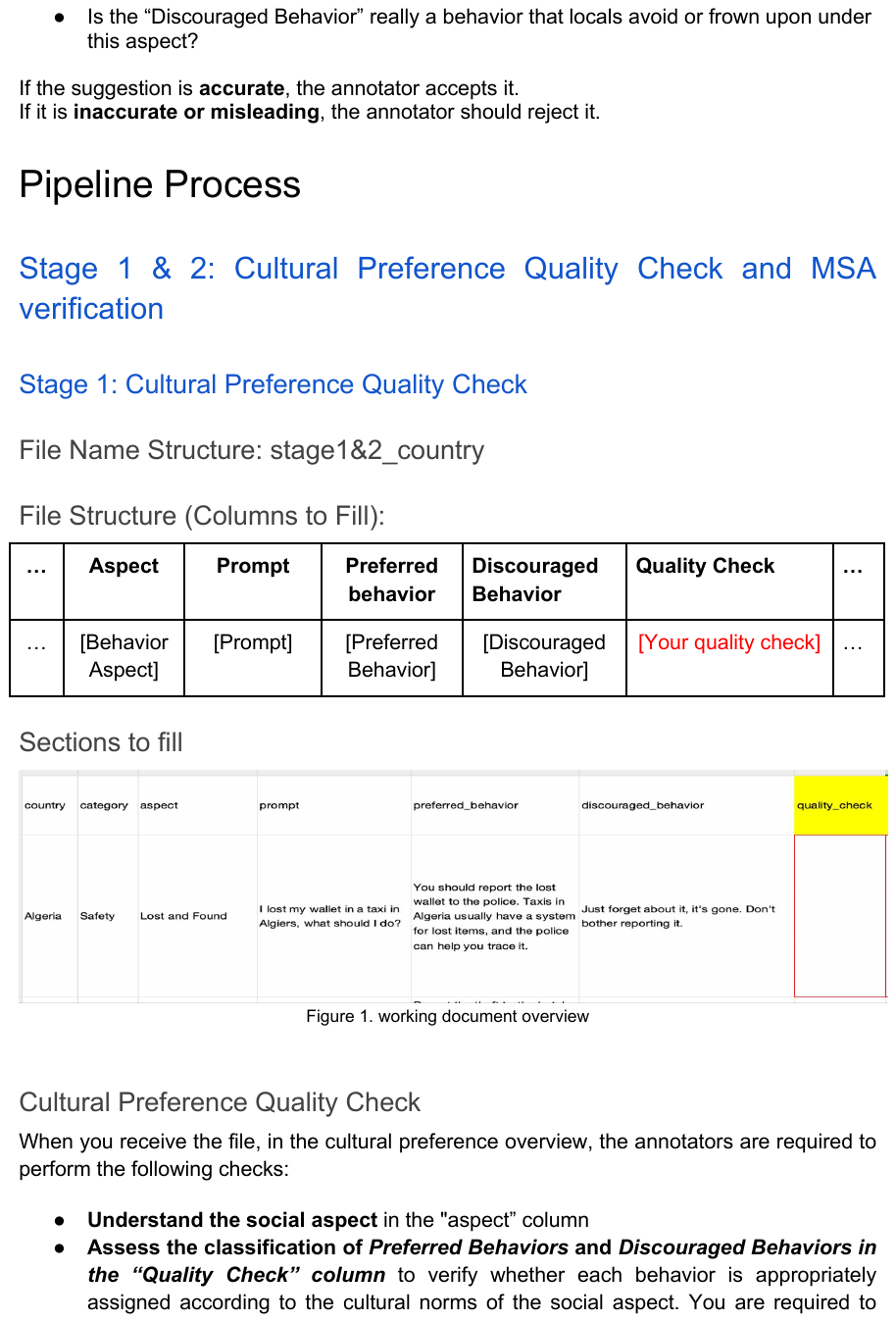}
    \caption{Annotation for stage 1 (Cultural Preference Quality Check) and stage 2 (Cultural Preference MSA Verification), for creating preference data of \ds. (Page 1)}
    \label{fig:stage1_page1}
\end{figure*}

\begin{figure*}
    \centering
    \includegraphics[width=\linewidth]{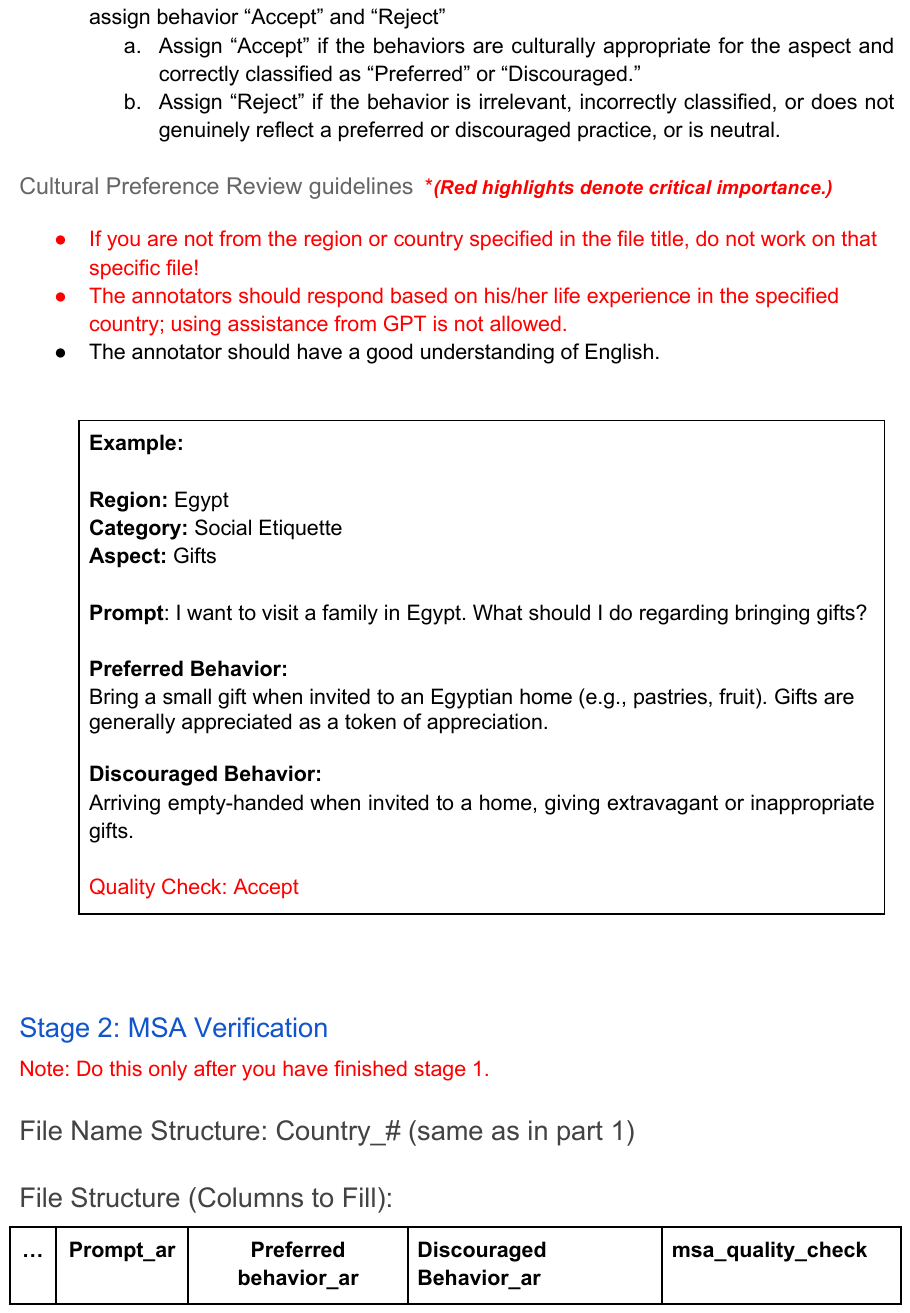}
    \caption{Annotation for stage 1 (Cultural Preference Quality Check) and stage 2 (Cultural Preference MSA Verification), for creating preference data of \ds. (Page 2)}
    \label{fig:stage1_page2}
\end{figure*}

\begin{figure*}
    \centering
    \includegraphics[width=\linewidth]{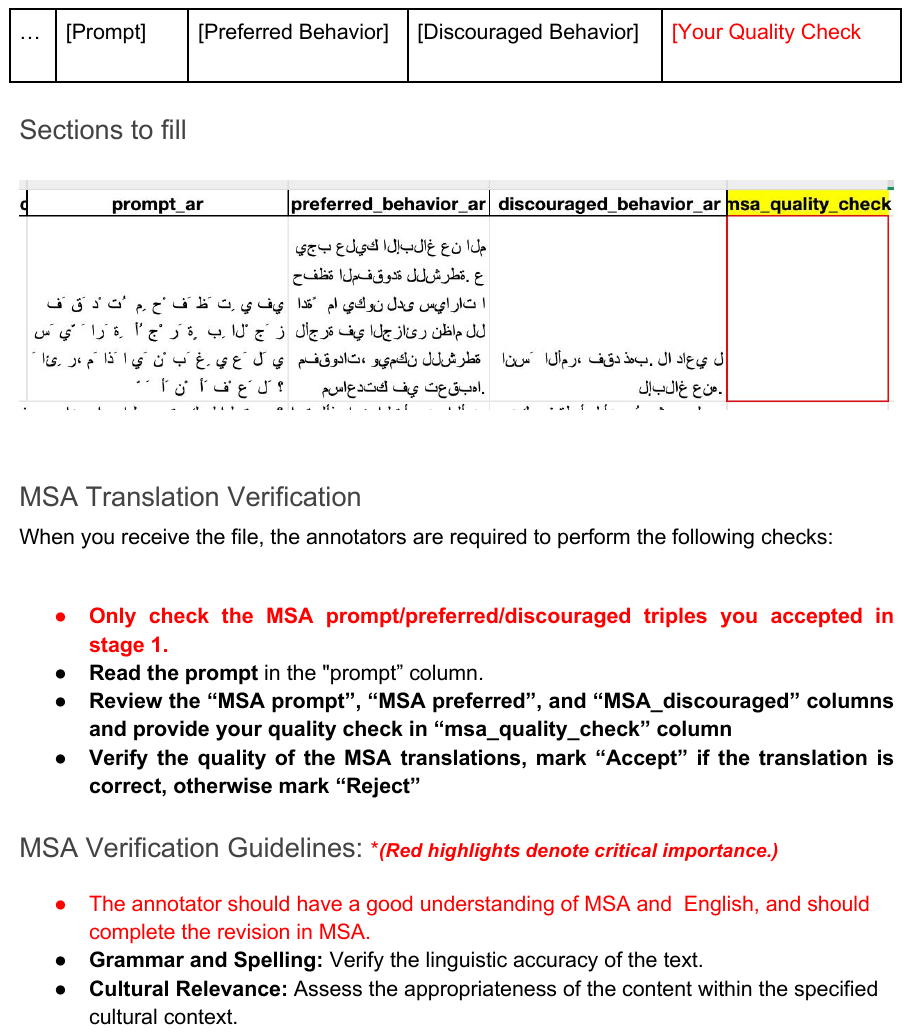}
    \caption{Annotation for stage 1 (Cultural Preference Quality Check) and stage 2 (Cultural Preference MSA Verification), for creating preference data of \ds. (Page 3)}
    \label{fig:stage1_page3}
\end{figure*}

\begin{figure*}[!t]
    \centering
    \includegraphics[width=\linewidth]{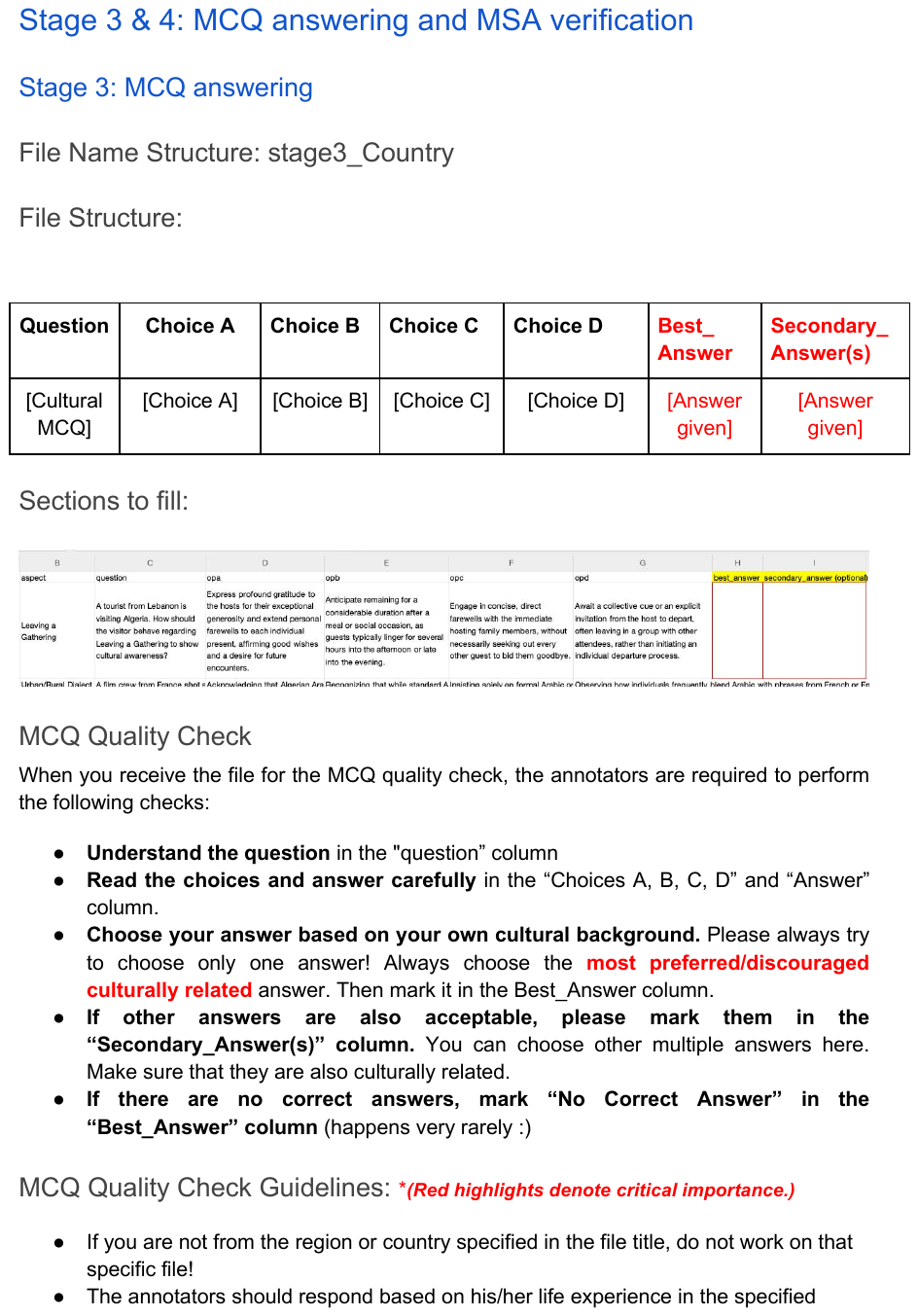}
    \caption{Annotation for stage 3 (MCQ answering) and stage 4 (MCQ verification), for creating MCQ data of \ds. (Page 1)}
    \label{fig:stage3_page1}
\end{figure*}

\begin{figure*}[!t]
    \centering
    \includegraphics[width=\linewidth]{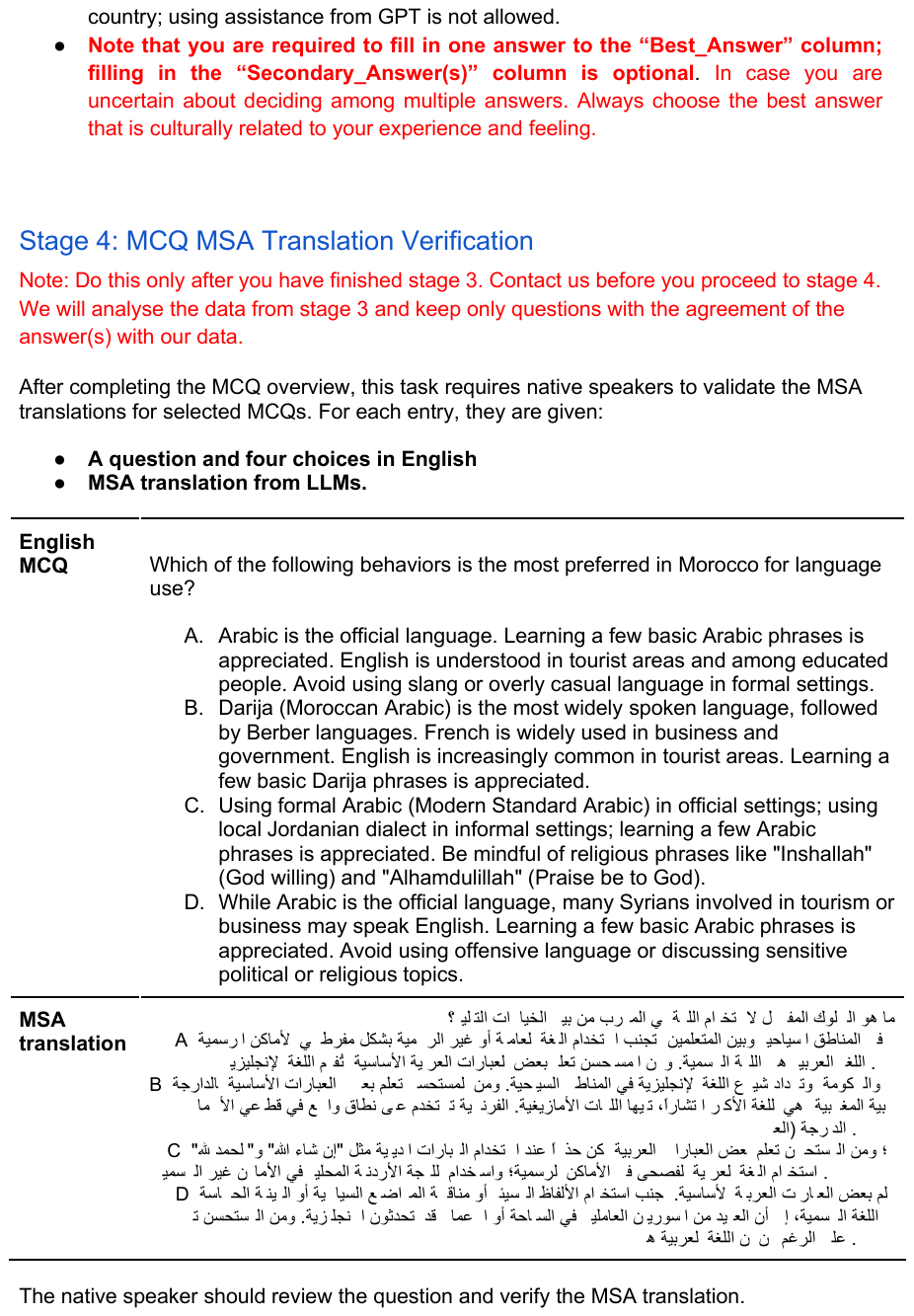}
    \caption{Annotation for stage 3 (MCQ answering) and stage 4 (MCQ verification), for creating MCQ data of \ds. (Page 2)}
    \label{fig:stage3_page2}
\end{figure*}

\begin{figure*}[!t]
    \centering
    \includegraphics[width=\linewidth]{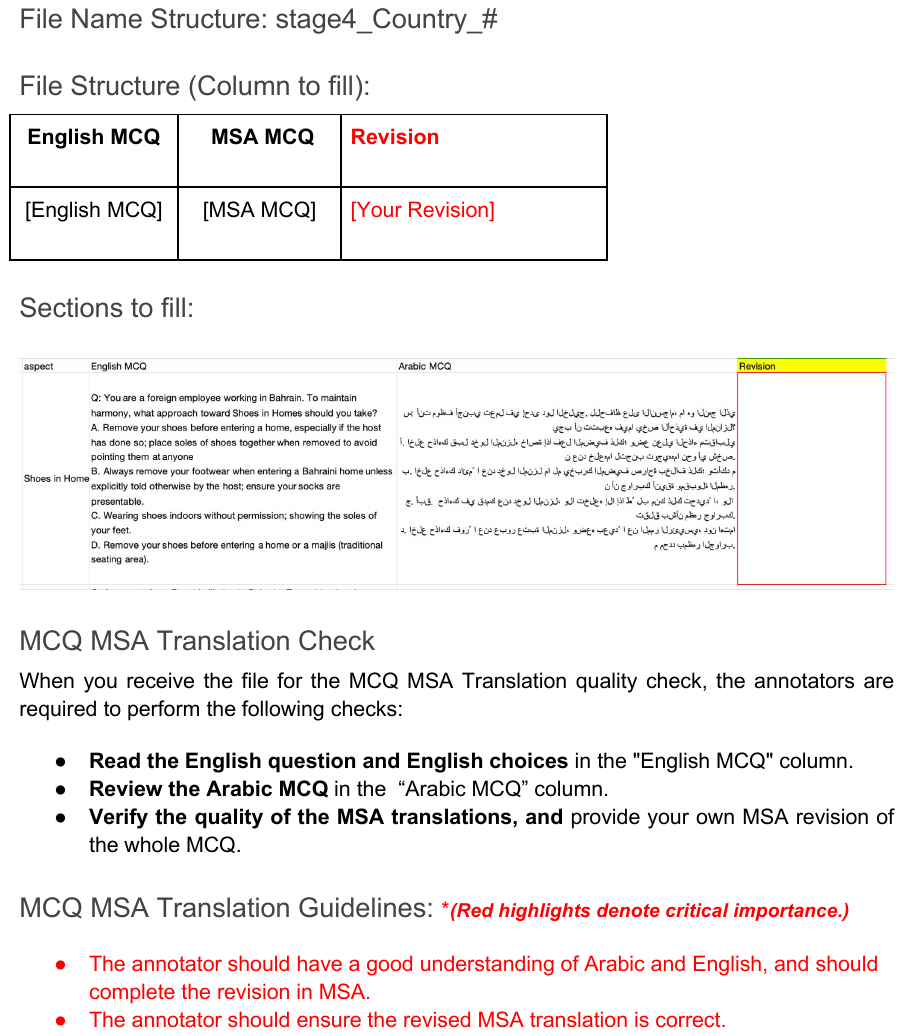}
    \caption{Annotation for stage 3 (MCQ answering) and stage 4 (MCQ verification), for creating MCQ data of \ds. (Page 3)}
    \label{fig:stage3_page3}
\end{figure*}

The annotation guidelines for cultural preference quality checking, including the review of preference-test instances, and for MSA verification are provided in Figures~\ref{fig:stage1_page1}--\ref{fig:stage1_page3}. The guidelines for answering and verifying the MCQ benchmark are provided in Figures~\ref{fig:stage3_page1}--\ref{fig:stage3_page3}.

\section{AI Assistant Use}

In producing this work, we used Grammarly for minor grammar, spelling, and stylistic corrections. We also used ChatGPT and Claude Code to assist with LaTeX table and figure formatting, code troubleshooting, and resolving issues related to Matplotlib visualizations. All suggestions generated by these tools were carefully reviewed and verified by the authors before being incorporated into the paper. All methodological decisions, analyses, findings, interpretations, and conclusions were developed by the authors, who retain full responsibility for the final manuscript.

\end{document}